\documentclass{now}

\usepackage{epstopdf}
\usepackage{amsmath}
\usepackage{amssymb}
\usepackage{algorithmic}
\usepackage{algorithm}
\usepackage{natbib}
\usepackage{epsfig}
\usepackage{color}
\usepackage{rotating}

\def\citealt{\citep}
\def\citet{\citep}

 \usepackage{algolyx}
\makeatother

\newcounter{assumption}
\renewcommand{\theassumption}{A\arabic{assumption}}

\newenvironment{ass}[1][]{\begin{trivlist}\item[] \refstepcounter{assumption}%
 {\bf Assumption\ \theassumption\ #1} }{
 \ifvmode\smallskip\fi\end{trivlist}}

\newcounter{model}
\renewcommand{\themodel}{\arabic{model}}

\newenvironment{model}[1][]{\begin{trivlist}\item[] \refstepcounter{model}%
 {\bf Model\ \themodel\ #1} }{
 \ifvmode\smallskip\fi\end{trivlist}}

\def\Re{\mathbb{R}}
\def\Nat{\mathbb{N}}
\def\argmax{\mathop{\rm arg\,max}}

\def\log{\mathop{\rm log}}

\def\Var{{\bf Var}}
\def\Cov{{\bf Cov}}
\def\exptE{{\bf E}}
\def\diag{\mathop{\rm diag}}
\def\regret{{\bf Regret}}
\def\KL{{\bf KL}}
\def\Ent{{\bf H}}

\def\A{{\mathcal{A}}}

\def\D{{\mathcal{D}}}

\def\F{{\mathcal{F}}}

\def\I{{\mathcal{I}}}

\def\M{{\mathcal{M}}}
\def\N{{\mathcal{N}}}
\def\O{{\mathcal{O}}}
\def\P{{\mathcal{P}}}

\def\Scal{{\mathcal{S}}}

\def\X{{\mathcal{X}}}
\def\Y{{\mathcal{Y}}}
\def\Z{{\mathcal{Z}}}

\def\vecalpha{{\boldsymbol{\alpha}}}
\def\vecphi{{\boldsymbol{\phi}}}

\def\vecsigma{{\boldsymbol{\sigma}}}
\def\vectheta{{\boldsymbol{\theta}}}

\def\vec0{{\boldsymbol{0}}}

\def\vecb{{\boldsymbol{b}}}
\def\vecf{{\boldsymbol{f}}}
\def\bvecf{{\boldsymbol{\bar{f}}}}
\def\veck{{\boldsymbol{k}}}

\def\vecr{{\boldsymbol{r}}}
\def\vecu{{\boldsymbol{u}}}

\def\vecw{{\mathbf{w}}}

\def\vecy{{\boldsymbol{y}}}

\def\vecF{{\boldsymbol{F}}}
\def\vecN{{\boldsymbol{N}}}
\def\vecR{{\boldsymbol{R}}}
\def\vecV{{\boldsymbol{V}}}
\def\vecW{{\boldsymbol{W}}}
\def\vecY{{\boldsymbol{Y}}}

\def\matSigma{{\boldsymbol{\Sigma}}}

\def\matPhi{{\boldsymbol{\Phi}}}

\def\matB{{\boldsymbol{B}}}
\def\matC{{\boldsymbol{C}}}

\def\matG{{\boldsymbol{G}}}
\def\matH{{\boldsymbol{H}}}
\def\matI{{\boldsymbol{I}}}
\def\matK{{\boldsymbol{K}}}

\def\matS{{\boldsymbol{S}}}
\def\matU{{\boldsymbol{U}}}

\def\matY{{\boldsymbol{Y}}}

\title{Bayesian Reinforcement Learning: A Survey}

\author{\begin{tabular}{c}
Mohammad Ghavamzadeh \tabularnewline
Adobe Research \& INRIA \tabularnewline
mohammad.ghavamzadeh@inria.fr \tabularnewline
\tabularnewline
Shie Mannor \tabularnewline
Technion \tabularnewline
shie@ee.technion.ac.il\tabularnewline
\tabularnewline
Joelle Pineau \tabularnewline
McGill University \tabularnewline
jpineau@cs.mcgill.ca \tabularnewline
\tabularnewline
Aviv Tamar\tabularnewline
University of California, Berkeley\tabularnewline
avivt@berkeley.edu\tabularnewline
\end{tabular}
}

\begin{document}

\copyrightowner{M.~Ghavamzadeh, S.~Mannor, J.~Pineau, and A.~Tamar}
\volume{8}
\issue{5-6}
\pubyear{2015}
\copyrightyear{2015}
\isbn{978-1-68083-088-0}
\doi{10.1561/2200000049}
\firstpage{359}
\lastpage{492}

\frontmatter  

\maketitle

\tableofcontents

\mainmatter


\begin{abstract}

Bayesian methods for machine learning have been widely investigated, yielding principled methods for incorporating prior information into inference algorithms. In this survey, we provide an in-depth review of the role of Bayesian methods for the reinforcement learning (RL) paradigm. The major incentives for incorporating Bayesian reasoning in RL are: {\bf 1)} it provides an elegant approach to action-selection (exploration/exploitation) as a function of the uncertainty in learning; and {\bf 2)} it provides a machinery to incorporate  prior knowledge into the algorithms. We first discuss models and methods for Bayesian inference in the simple single-step Bandit model. We then review the extensive recent literature on Bayesian methods for model-based RL, where prior information can be expressed on the parameters of the Markov model. We also present Bayesian methods for model-free RL, where priors are expressed over the value function or policy class. The objective of the paper is to provide a comprehensive survey on Bayesian RL algorithms and their theoretical and empirical properties.
\end{abstract}



\chapter{Introduction}
\label{s:intro}

A large number of problems in science and engineering, from robotics to game playing, tutoring systems, resource management, financial portfolio management, medical treatment design and beyond, can be characterized as sequential decision-making under uncertainty. Many interesting sequential decision-making tasks can be formulated as reinforcement learning (RL) problems~\citep{Bertsekas96NP,Sutton98IR}. In an RL problem, an agent interacts with a dynamic, stochastic, and incompletely known environment, with the goal of finding an action-selection strategy, or {\em policy}, that optimizes some long-term performance  measure.


One of the key features of RL is the focus on learning a control policy to optimize the choice of actions over several time steps.  This is usually learned from sequences of data. In contrast to supervised learning methods that deal with independently and identically distributed (i.i.d.)~samples from the domain, the RL agent learns from the samples that are collected from the trajectories generated by its sequential interaction with the system. Another important aspect is the effect of the agent's policy on the data collection; different policies naturally yield different distributions of sampled trajectories, and thus, impacting what can be learned from the data.

Traditionally, RL algorithms have been categorized as being either \textit{model-based} or \textit{model-free}. In the former category, the agent uses the collected data to first build a model of the domain's dynamics and then uses this model to optimize its policy. In the latter case, the agent directly learns an optimal (or good) action-selection strategy from the collected data. There is some evidence that the first method provides better results with less data~\citep{Atkeson97}, and the second method may be more efficient in cases where the solution space (e.g.,~policy space) exhibits more regularity than the underlying dynamics, though there is some disagreement about this,.

A major challenge in RL is in identifying good data collection strategies, that effectively balance between the need to explore the space of all possible policies, and the desire to focus data collection towards trajectories that yield better outcome (e.g.,~greater chance of reaching a goal, or minimizing a cost function). This is known as the \textit{exploration-exploitation} tradeoff. This challenge arises in both model-based and model-free RL algorithms.

Bayesian reinforcement learning (BRL) is an approach to RL that leverages methods from Bayesian inference to incorporate information into the learning process.  It assumes that the designer of the system can express prior information about the problem in a probabilistic distribution, and that new information can be incorporated using standard rules of Bayesian inference. The information can be encoded and updated using a parametric representation of the system dynamics, in the case of model-based RL, or of the solution space, in the case of model-free RL.

A major advantage of the BRL approach is that it provides a principled way to tackle the exploration-exploitation problem. Indeed, the Bayesian posterior naturally captures the full state of knowledge, subject to the chosen parametric representation, and thus, the agent can select actions that maximize the expected gain with respect to this information state.

Another major advantage of BRL is that it implicitly facilitates regularization. By assuming a prior on the value function, the parameters defining a policy, or the model parameters, we avoid the trap of letting a few data points steer us away from
the true parameters. On the other hand, having a prior precludes overly rapid convergence. The role of the prior is therefore to soften the effect of sampling a finite dataset, effectively leading to regularization. We note that regularization in RL has been addressed for the value function \citep{ewrlFarahmandGSM08} and for policies \citep{FarahmandGSM08}. A major issue with these regularization schemes is that it is not clear how to select the regularization coefficient. Moreover, it is not clear why an optimal value function (or a policy) should belong to some pre-defined set.

Yet another advantage of adopting a Bayesian view in RL is the principled Bayesian approach for handling parameter uncertainty. Current frequentist approaches for dealing with modelling errors in sequential decision making are either very conservative, or computationally infeasible \citep{nilim2005robust}. By explicitly modelling the distribution over unknown system parameters, Bayesian methods offer a promising approach for solving this difficult problem.

Of course, several challenges arise in applying Bayesian methods to the RL paradigm.  First, there is the challenge of selecting the correct  representation for expressing prior information in any given domain.  Second, defining the decision-making process over the information state is typically computationally more demanding than directly considering the natural state representation.  Nonetheless, a large array of models and algorithms have been proposed for the BRL framework, leveraging a variety of structural assumptions and approximations to provide feasible solutions.

\begin{figure}
\centering
\includegraphics[scale=0.6]{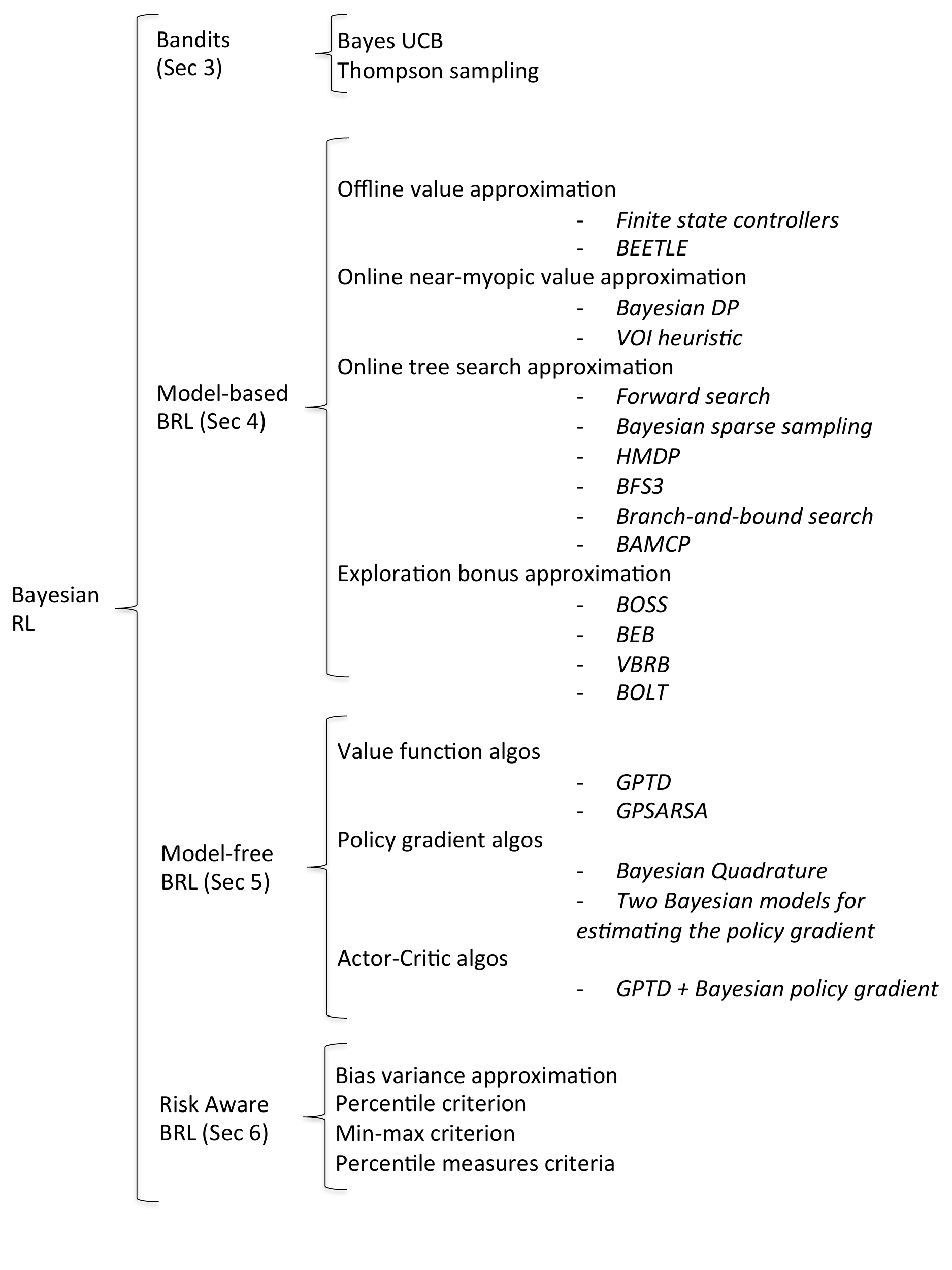}
\caption{Overview of the Bayesian RL approaches covered in this survey.}
\label{fig:taxonomy}
\end{figure}

The main objective of this paper is to provide a comprehensive survey on BRL algorithms and their theoretical and empirical properties.  In Chapter~\ref{s:background}, we provide a review of the main mathematical concepts and techniques used throughout this paper.  Chapter~\ref{s:Bayesian-bandits} surveys the Bayesian learning methods for the case of single-step decision-making, using the {\em bandit} framework. This section serves both as an exposition of the potential of BRL in a simpler setting that is well understood, but is also of independent interest, as bandits have widespread applications.
The main results presented here are of a theoretical nature, outlining known performance bounds for the regret minimization criteria.
Chapter~\ref{s:model-based-brl} reviews existing methods for \emph{model-based} BRL, where the posterior is expressed over parameters of the system dynamics model.  Chapter~\ref{s:model-free-brl} focuses on BRL methods that do not explicitly learn a model of the system, but rather the posterior is expressed over the solution space.  Chapter~\ref{s:risk} focuses on a particular advantage of BRL in dealing with risk due to parameter-uncertainty, and surveys several approaches for incorporating such risk into the decision-making process. Finally, Chapter~\ref{s:misc} discusses various extensions of BRL for special classes of problems (PAC-Bayes model selection, inverse RL, multi-agent RL, and multi-task RL). Figure~\ref{fig:taxonomy} outlines the various BRL approaches covered throughout the paper.

\subsection*{An Example Domain}
\label{ss:brl}

We present an illustrative domain suitable to be solved using the BRL techniques surveyed in this paper. This running example will be used throughout the paper to elucidate the difference between the various BRL approaches and to clarify various BRL concepts.

\begin{example}[The Online Shop]
\label{exm:online_shop}

In the online shop domain, a retailer aims to maximize profit by sequentially suggesting products to online shopping customers. Formally, the domain is characterized by the following model:

\begin{itemize}
\item A set of possible customer states, $\X$. States can represent intrinsic features of the customer such as gender and age, but also dynamic quantities such as the items in his shopping cart, or his willingness to shop;
\item A set of possible product suggestions and advertisements, $\A$;
\item A probability kernel, $P$, defined below.
\end{itemize}

An {\bf\em episode} in the online shop domain begins at time $t=0$, when a customer with features $x_0 \in \X$ enters the online shop.
Then, a sequential interaction between the customer and the online shop begins, where at each step $t=0,1,2,\dots$, an advertisement $a_t\in \A$ is shown to the customer, and following that the customer makes a decision to either (i) add a product to his shopping cart; (ii) not buy the product, but continue to shop; (iii) stop shopping and check out. Following the customers decision, his state changes to $x_{t+1}$ (reflecting the change in the shopping cart, willingness to continue shopping, etc.). We assume that this change is captured by a probability kernel $P(\left. x_{t+1} \right|x_t,a_t)$.

When the customer decides to check out, the episode ends, and a profit is obtained according to the items he had added to his cart. The goal is to find a product suggestion policy, $x \rightarrow a\in\A$, that maximizes the expected total profit.
\end{example}

When the probabilities of customer responses $P$ are known in advance, calculating an optimal policy for the online shop domain is basically a \emph{planning} problem, which may be solved using traditional methods for resource allocation~\citep{powell2007approximate}. A more challenging, but realistic,  scenario is when $P$ is not completely known beforehand, but has to be \emph{learned} while interacting with customers. The BRL framework employs \emph{Bayesian}  methods for learning $P$, and for learning an optimal product suggestion policy.

There are several advantages for choosing a Bayesian approach for the online shop domain. First, it is likely that some prior knowledge about $P$ is available. For example, once a customer adds a product of a particular brand to his cart, it is likely that he prefers additional products of the same brand over those of a different one. Taking into account such knowledge is natural in the Bayesian method, by virtue of the \emph{prior} distribution over $P$. As we shall see, the Bayesian approach also naturally extends to more general forms of \emph{structure} in the problem.

A second advantage concerns what is known as the \emph{exploitation--exploration} dilemma: should the decision-maker display only the most profitable product suggestions according to his current knowledge about $P$, or rather take exploratory actions that may turn out to be less profitable, but provide useful information for future decisions? The Bayesian method offers a principled approach to dealing with this difficult problem by explicitly quantifying the value of exploration, made possible by maintaining a \emph{distribution} over $P$.

The various parameter configurations in the online shop domain lead to the different learning problems surveyed in this paper. In particular:
\begin{itemize}
\item For a single-step interaction, i.e.,~when the episode terminates after a single product suggestion, the problem is captured by the multi-armed bandit model of Chapter \ref{s:Bayesian-bandits}.
\item For small-scale problems, i.e.,~a small number of products and customer types, $P$ may be learnt explicitly. This is the model-based approach of Chapter \ref{s:model-based-brl}.
\item For large problems, a near-optimal policy may be obtained without representing $P$ explicitly. This is the model-free approach of Chapter \ref{s:model-free-brl}.
\item When the customer state is not fully observed by the decision-maker, we require models that incorporate partial observability; see  \S\ref{ss:pomdp} and \S\ref{ss:ext-pomdp}.
\end{itemize}

Throughout the paper, we revisit the online shop domain, and specify explicit configurations that are relevant to the surveyed methods.


\chapter{Technical Background}
\label{s:background}

In this section we provide a brief overview of the main concepts and introduce the notations used throughout the paper. We begin with a quick review of the primary mathematical tools and algorithms leveraged in the latter sections to define BRL models and algorithms.



\section{Multi-Armed Bandits}
\label{s:MAB}

As was previously mentioned, a key challenge in sequential decision-making under uncertainty is the exploration/exploitation dilemma: the tradeoff between either taking actions that are most rewarding according to the current state of knowledge, or taking exploratory actions, which may be less immediately rewarding, but may lead to better informed decisions in the future.

The \emph{stochastic multi-armed bandit} (MAB) problem is perhaps the simplest model of the exploration/exploitation tradeoff. Originally formulated as the problem of a gambler choosing between different slot machines in a casino (`one armed bandits'), the stochastic MAB model features a decision-maker that sequentially chooses actions $a_t \in \A$, and observes random outcomes $Y_t(a_t) \in \Y$ at discrete time steps $t=1,2,3,\ldots$. A known function $r : \Y \to \Re$ associates the random outcome to a reward, which the agent seeks to maximize. The outcomes $\left\{ Y_t(a) \right\}$ are drawn i.i.d.~over time from an unknown probability distribution $P(\cdot|a)\in \P(\Y)$, where $\P(\Y)$ denotes the set of probability distributions on (Borel) subsets of $\Y$ (see \cite{Bubeck12RA} for reference).
\begin{model}[(Stochastic $K$-Armed Bandit)]
Define a $K$-MAB to be a tuple $\langle\A,\Y,P,r\rangle$ where\\
$\bullet$ $\A$ is the set of actions (arms), and $|\A|=K$,\\
$\bullet$ $\Y$ is the set of possible outcomes,\\
$\bullet$ $P(\cdot|a)\in \P(\Y)$ is the outcome probability, conditioned on action $a\in\A$ being taken,\\
$\bullet$ $r(Y) \in \Re$ represents the reward obtained when outcome $Y\in \Y$ is observed.
\end{model}
A rule that prescribes to the agent which actions to select, or {\em policy}, is defined as a mapping from past observations to a distribution over the set of actions.
Since the probability distributions are initially unknown, the decision-maker faces a tradeoff between exploitation, i.e.,~selecting the arm she believes is optimal, or making exploratory actions to gather more information about the true probabilities. Formally, this tradeoff is captured by the notion of \emph{regret}:
\begin{definition}[Regret]
Let $a^* \in \argmax_{a \in \A} \exptE_{y \sim P(\cdot|a)} \big[ r(y)\big]$ denote the optimal arm. The $T$-period regret of the sequence of actions $a_1,\dots,a_T$ is the random variable
\begin{equation*}
\label{eq:regret_definition}
\regret (T) = \sum_{t=1}^{T} \Big[r\big(Y_t(a^*)\big) - r\big(Y_t(a_t)\big)\Big].
\end{equation*}
\end{definition}

\noindent
Typically, MAB algorithms focus on the \emph{expected regret}, $\exptE\big[\regret (T)\big]$, and provide policies that are guaranteed to keep it small in some sense.

As an illustration of the MAB model, let us revisit the online shop example.
\begin{example}[Online shop -- bandit setting]\label{exm:online_shop_bandit}
Recall the online shop domain of Example \ref{exm:online_shop}. In the MAB setting, there is no state information about the customers, i.e.,~$\X=\emptyset$. In addition, each interaction lasts a single time step, after which the customer checks out, and a profit is obtained according to his purchasing decision. The regret minimization problem corresponds to determining which sequence of advertisements $a_1,\dots,a_T$ to show to a stream of $T$ incoming customers, such that the total revenue is close to that which would have been obtained with the optimal advertisement stream.
\end{example}

In some cases, the decision-maker may have access to some additional information that is important for making decisions, but is not captured in the MAB model. For example, in the online shop domain of Example \ref{exm:online_shop_bandit}, it is likely that some information about the customer, such as his age or origin, is available. The \emph{contextual bandit} model (a.k.a. associative bandits, or bandits with side information) is an extension of the MAB model that takes such information into account.

The decision-making process in the contextual bandit model is similar to the MAB case. However, at each time step $t$, the decision maker first observes a context $s_t\in\Scal$, drawn i.i.d.~over time from a probability distribution $P_{S}(\cdot)\in \P(\Scal)$, where $\P(\Scal)$ denotes the set of probability distributions on (Borel) subsets of $\Scal$. The decision-maker then chooses an action $a_t \in \A$ and observes a random outcome $Y_t(a_t,s_t) \in \Y$, which now depends both on the context and the action. The outcomes $\left\{ Y_t(a) \right\}$ are drawn i.i.d.~over time from an unknown probability distribution $P(\cdot|a,s)\in \P(\Y)$, where $\P(\Y)$ denotes the set of probability distributions on (Borel) subsets of $\Y$.
\begin{model}[(Contextual Bandit)]
Define a contextual bandit to be a tuple $\langle \Scal,\A,\Y,P_{S},P,r\rangle$ where\\
$\bullet$ $\Scal$ is the set of contexts,\\
$\bullet$ $\A$ is the set of actions (arms),\\
$\bullet$ $\Y$ is the set of possible outcomes,\\
$\bullet$ $P_{S}(\cdot)\in \P(\Scal)$ is the context probability,\\
$\bullet$ $P(\cdot|a,s)\in \P(\Y)$ is the outcome probability, conditioned on action $a\in\A$ being taken when the context is $s\in\Scal$,\\
$\bullet$ $r(Y) \in \Re$ represents the reward obtained when outcome $Y\in \Y$ is observed.
\end{model}
\newpage
The Markov decision process model, as presented in the following section, may be seen as an extension of the contextual bandit model to a \emph{sequential} decision-making model, in which the context is no longer i.i.d., but may change over time according to the selected actions.

\section{Markov Decision Processes}
\label{ss:mdp}

The Markov Decision Process (MDP) is a framework for sequential decision-making in Markovian dynamical systems~\citep{bellman57,Puterman94MD}. It can be seen as an extension of the MAB framework by adding the notion of a system \emph{state}, that may dynamically change according to the performed actions and affects the outcomes of the system.

\begin{model}[(Markov Decision Process)]
Define an MDP $\M$ to be a tuple $\langle\Scal,\A,P,P_0,q\rangle$ where\\
$\bullet$ $\Scal$ is the set of states,\\
$\bullet$ $\A$ is the set of actions,\\
$\bullet$ $P(\cdot | s,a)\in\P(\Scal)$ is the probability distribution over next states, conditioned on action $a$ being taken in state $s$,\\
$\bullet$ $P_0\in\P(\Scal)$ is the probability distribution according to which the initial state is selected,\\
$\bullet$ $R(s,a) \sim q(\cdot | s,a) \in \P(\Re)$ is a random variable representing the reward obtained when action $a$ is taken in state $s$.
\end{model}

Let $\P(\Scal)$, $\P(\A)$, and $\P(\Re)$ be the set of probability distributions on (Borel) subsets of $\Scal$, $\A$, and $\Re$ respectively.\footnote{$\Re$ is the set of real numbers.}
We assume that $P$, $P_0$ and $q$ are stationary.
Throughout the paper, we use upper-case and lower-case letters to refer to random variables and the values taken by random variables, respectively. For example, $R(s,a)$ is the random variable of the immediate reward, and $r(s,a)$ is one possible realization of this random variable. We denote the expected value of $R(s,a)$, as $\bar{r}(s,a)=\int r\;q(dr|s,a)$.
\begin{ass}[(MDP Regularity)]
\label{ass:regularity}
We assume that the random immediate rewards are bounded by $R_{\max}$ and the expected immediate rewards are bounded by $\bar{R}_{\max}$. Note that $\bar{R}_{\max}\leq R_{\max}$.
\end{ass}
A rule according to which the agent selects its actions at each possible state, or {\em policy}, is defined as a mapping from past observations to a distribution over the set of actions. A policy is called {\em Markov} if the distribution depends only on the last state of the observation sequence. A policy is called {\em stationary} if it does not change over time. A stationary Markov policy $\mu(\cdot|s)\in\P(\A)$ is a probability distribution over the set of actions given a state $s\in\Scal$. A policy is {\em deterministic} if the probability distribution concentrates on a single action for all histories. A deterministic policy is identified by a mapping from the set of states to the set of actions, i.e.,~$\mu:\Scal\rightarrow\A$. In the rest of the paper, we use the term policy to refer to stationary Markov policies.

The MDP controlled by a policy $\mu$ induces a Markov chain $\M^\mu$ with reward distribution $q^\mu(\cdot|s)=q\big(\cdot|s,\mu(s)\big)$ such that $R^\mu(s)=R\big(s,\mu(s)\big)\in q^\mu(\cdot|s)$, transition kernel $P^\mu(\cdot|s)=P\big(\cdot|s,\mu(s)\big)$, and stationary distribution over states $\pi^\mu$ (if it admits one). In a Markov chain $\M^\mu$, for state-action pairs $z=(s,a)\in\Z=\Scal\times\A$, we define the transition density and the initial (state-action) density as $P^{\mu}(z'|z)=P(s'|s,a)\mu(a'|s')$ and $P^\mu_0(z_0)=P_0(s_0)\mu(a_0|s_0)$, respectively. We generically use $\xi=\{z_0,z_1,\ldots,z_T\}\in\Xi$, $T\in\{0,1,\ldots,\infty\}$, to denote a path (or trajectory) generated by this Markov chain. The probability (density) of such a path is given by
\begin{equation*}
\Pr(\xi|\mu)=P^\mu_0(z_0)\prod_{t=1}^TP^\mu(z_t|z_{t-1}).
\end{equation*}
We define the (possibly discounted, $\gamma\in[0,1]$) return of a path as a function $\rho:\Xi\rightarrow\Re$, $\rho(\xi)=\sum_{t=0}^T\gamma^tR(z_t)$. For each path $\xi$, its (discounted) return, $\rho(\xi)$, is a random variable with the expected value $\bar{\rho}(\xi)=\sum_{t=0}^T\gamma^t\bar{r}(z_t)$.\footnote{When there is no randomness in the rewards, i.e.,~$r(s,a)=\bar{r}(s,a)$, then $\rho(\xi)=\bar{\rho}(\xi)$.} Here, $\gamma\in[0,1]$ is a discount factor that determines the exponential devaluation rate of delayed rewards.\footnote{When $\gamma=1$, the policy must be proper, i.e., guaranteed to terminate, see~\citep{Puterman94MD} or~\citep{Bertsekas96NP} for more details.} The expected return of a policy $\mu$ is defined by
\begin{equation}
\label{eq:pol-expt-ret1}
\eta(\mu)=\exptE\big[\rho(\xi)\big]=\int_{\Xi}\bar{\rho}(\xi)\Pr(\xi|\mu)d\xi.
\end{equation}
The expectation is over all possible trajectories generated by policy $\mu$ and all possible rewards collected in them.

Similarly, for a given policy $\mu$, we can define the (discounted) return of a state $s$, $D^\mu(s)$, as the sum of (discounted) rewards that the agent encounters when it starts in state $s$ and follows policy $\mu$ afterwards
\begin{equation}\label{eq:D-def}
D^{\mu}(s)=\sum_{t=0}^\infty\gamma^tR(Z_t)\mid Z_0=\big(s,\mu(\cdot|s)\big),\hspace{0.15in}\text{with}\hspace{0.075in}S_{t+1}\sim P^\mu(\cdot|S_t).
\end{equation}
The expected value of $D^\mu$ is called the value function of policy $\mu$
\begin{equation}\label{eq:val-func}
V^{\mu}(s)=\exptE\big[D^\mu(s)\big]=\exptE\left[\left.\sum_{t=0}^\infty\gamma^tR(Z_t)\;\right| Z_0=\big(s,\mu(\cdot|s)\big)\right].
\end{equation}
Closely related to value function is the action-value function of a policy, the total expected (discounted) reward observed by the agent when it starts in state $s$, takes action $a$, and then executes policy $\mu$
\begin{equation*}
Q^{\mu}(z)=\exptE\big[D^\mu(z)\big]=\exptE\left[\left.\sum_{t=0}^\infty\gamma^tR(Z_t)\;\right| Z_0=z\right],
\end{equation*}
where similarly to $D^\mu(s)$, $D^\mu(z)$ is the sum of (discounted) rewards that the agent encounters when it starts in state $s$, takes action $a$, and follows policy $\mu$ afterwards. It is easy to see that for any policy $\mu$, the functions $V^\mu$ and $Q^\mu$ are bounded by $\bar{R}_{\max}/(1-\gamma)$. 
We may write the value of a state $s$ under a policy $\mu$ in terms of its immediate reward and the values of its successor states under $\mu$ as
\begin{equation}\label{eq:Bellman-eq}
V^\mu(s)=R^\mu(s)+\gamma\int_\Scal P^\mu(s'|s)V^\mu(s')ds',
\end{equation}
which is called the {\em Bellman equation} for $V^\mu$.

Given an MDP, the goal is to find a policy that attains the best possible values, $V^*(s)=\sup_\mu V^\mu(s)$, for all states $s\in\Scal$. The function $V^*$ is called the {\em optimal} value function. A policy is optimal (denoted by $\mu^*$) if it attains the optimal values at all the states, i.e., $V^{\mu^*}(s)=V^*(s)$ for all $s\in\Scal$. In order to characterize optimal policies it is useful to define the optimal action-value
function
\begin{equation}\label{eq:q-optimal}
Q^*(z)=\sup_\mu Q^\mu(z).
\end{equation}
Further, we say that a deterministic policy $\mu$ is {\em greedy} with respect to an action-value function $Q$, if for all $s\in\Scal$ and $a\in\A$, $\mu(s)\in\argmax_{a\in\A}Q(s,a)$. Greedy policies are important because any greedy policy with respect to $Q^*$ is optimal. 
Similar to the value function of a policy (Eq.~\ref{eq:Bellman-eq}), the optimal value function of a state $s$ may be written in terms of the optimal values of its successor states as
\begin{equation}\label{eq:opt-Bellman-eq}
V^*(s)=\max_{a\in\A}\left[R(s,a)+\gamma\int_\Scal P(s'|s,a)V^*(s')ds'\right],
\end{equation}
which is called the {\em Bellman optimality equation}.
%
%
Note that, similar to Eqs.~\ref{eq:Bellman-eq} and~\ref{eq:opt-Bellman-eq}, we may define the Bellman equation and the Bellman optimality equation for action-value function.


It is important to note that almost all methods for finding the optimal solution of an MDP are based on two {\em dynamic programming} (DP) algorithms: {\em value iteration} and {\em policy iteration}. Value iteration (VI) begins with a value function $V_0$ and at each iteration $i$, generates a new value function by applying Eq.~\ref{eq:opt-Bellman-eq} to the current value function, i.e.,
\begin{equation*}
V_i^*(s)=\max_{a\in\A}\left[R(s,a)+\gamma\int_\Scal P(s'|s,a)V_{i-1}^*(s')ds'\right],\quad\quad\quad\forall s\in\Scal.
\end{equation*}
Policy iteration (PI) starts with an initial policy. At each iteration, it evaluates the value function of the current policy, this process is referred to as {\em policy evaluation} (PE), and then performs a {\em policy improvement} step, in which a new policy is generated as a greedy policy with respect to the value of the current policy. Iterating the policy evaluation -- policy improvement process is known to produce a strictly monotonically improving sequence of policies. If the improved policy is the same as the policy improved upon, then we are assured that the optimal policy has been found.


\section{Partially Observable Markov Decision Processes}
\label{ss:pomdp}

The Partially Observable Markov Decision Process (POMDP) is an extension of MDP to the case where the state of the system is not necessarily observable~\citep{astrom65,smallwood73,kaelbling98}.

\begin{model}[(Partially Observable Markov Decision Process)]
Define a POMDP $\M$ to be a tuple $\langle\Scal,\A,\O,P,\Omega,P_0,q\rangle$ where\\
$\bullet$ $\Scal$ is the set of states,\\
$\bullet$ $\A$ is the set of actions,\\
$\bullet$ $\O$ is the set of observations,\\
$\bullet$ $P(\cdot | s,a)\in\P(\Scal)$ is the probability distribution over next states, conditioned on action $a$ being taken in state $s$,\\
$\bullet$ $\Omega(\cdot | s,a)\in\P(\O)$ is the probability distribution over possible observations, conditioned on action $a$ being taken to reach state $s$ where the observation is perceived,\\
$\bullet$ $P_0\in\P(\Scal)$ is the probability distribution according to which the initial state is selected,\\
$\bullet$ $R(s,a) \sim q(\cdot | s,a) \in \P(\Re)$ is a random variable representing the reward obtained when action $a$ is taken in state $s$.
\end{model}

\noindent
All assumptions are similar to MDPs, with the addition that $\P(\O)$ is the set of probability distributions on (Borel) subsets of $\O$, and $\Omega$ is stationary. As a motivation for the POMDP model, we revisit again the online shop domain.
\begin{example}[Online shop -- hidden state setting]
Recall the online shop domain of Example \ref{exm:online_shop}. In a realistic scenario, some features of the customer, such as its gender or age, might not be visible to the decision maker due to privacy, or other reasons. In such a case, the MDP model, which requires the full state information for making decisions is not suitable. In the POMDP model, only observable quantities, such as the items in the shopping cart, are used for making decisions.
\end{example}

Since the state is not directly observed, the agent must rely on the recent history of actions and observations, $\{o_{t+1}, a_{t}, o_{t},\ldots, o_1, a_0\}$ to infer a distribution over states. This \emph{belief} (also called \emph{information state})  is defined over the state probability simplex, $b_t \in \Delta$, and can be calculated recursively as:
\begin{equation}
\label{eq:belief}
b_{t+1} (s') = \frac{\Omega(o_{t+1} | s',a_t) \int_\Scal P(s'|s,a_t)b_t(s) ds}{\int_\Scal \Omega(o_{t+1} | s'',a_t) \int_\Scal P(s''|s,a_t)b_t(s)ds ds''} \; ,
\end{equation}
where the sequence is initialized at $b_0:=P_0$ and the denominator can be seen as a simple normalization function.  For convenience, we sometimes denote the belief update (Eq.~\ref{eq:belief}) as $b_{t+1}=\tau(b_{t},a,o)$.

In the POMDP framework, the action-selection policy is defined as $\mu : \Delta(s) \rightarrow \A$. Thus, solving a POMDP involves finding the optimal policy, $\mu^*$, that maximizes the expected discounted sum of rewards for all belief states. This can be defined using a variant of the Bellman equation
\begin{equation}\label{eq:pomdp-bellman}
V^*(b_t)= \max_{a \in \A} \left[ \int_\Scal R(s,a)b_t(s)ds + \gamma\int_\O \Pr(o|b_{t},a)V^*\big(\tau(b_t,a,o)\big)do\right]\;.
\end{equation}
The optimal value function for a finite-horizon POMDP can be shown to be piecewise-linear and convex~\citep{smallwood73,porta05}.  In that case, the value function $V_t$ at any finite planning horizon $t$ can be represented by a finite set of linear segments $\Gamma_t=\{\alpha_0, \alpha_1,\ldots, \alpha_m\}$, often called $\alpha$-vectors (when $\Scal$ is discrete) or $\alpha$-functions (when $\Scal$ is continuous). Each defines a linear function over the belief state space associated with some action $a \in \A$. The value of a given $\alpha_i$ at a belief $b_t$ can be evaluated by linear interpolation
\begin{equation}
\alpha_i(b_t)=\int_\Scal \alpha_i(s)b_t(s)ds.
\end{equation}
The value of a belief state is the maximum value returned by one of the $\alpha$-functions for this belief state
\begin{equation}\label{eq:pomdp-v*max}
V^*_t(b_t)=\max_{\alpha \in \Gamma_t} \int_\Scal \alpha(s) b_t(s) \;.
\end{equation}
In that case, the best action, $\mu^*(b_t)$, is the one associated with the $\alpha$-vector that returns the best value.

The belief as defined here can be interpreted as a state in a particular kind of MDP, often called the \emph{belief-MDP}.  The main intuition is that for any partially observable system with known parameters, the belief is actually fully observable and computable (for small enough state spaces and is approximable for others). Thus, the planning problem is in fact one of planning in an MDP, where the state space corresponds to the belief simplex.  This does not lead to any computational advantage, but conceptually, suggests that known results and methods for MDPs can also be applied to the belief-MDP.


\section{Reinforcement Learning}
\label{ss:rl}

Reinforcement learning (RL)~\citep{Bertsekas96NP,Sutton98IR} is a class of learning {\em problems} in which an agent (or controller) interacts with a dynamic, stochastic, and incompletely known environment, with the goal of finding an action-selection strategy, or {\em policy}, to optimize some measure of its long-term performance. The interaction is conventionally modeled as an MDP, or if the environment state is not always completely observable, as a POMDP~\citep{Puterman94MD}.

Now that we have defined the MDP model, the next step is to solve it, i.e., to find an optimal policy.\footnote{It is not possible to find an optimal policy in many practical problems. In such cases, the goal would be to find a reasonably good policy, i.e.,~a policy with large enough value function.} In some cases, MDPs can be solved analytically, and in many cases they can be solved iteratively by {\em dynamic} or {\em linear programming} (e.g.,~see~\citep{Bertsekas96NP}). However, in other
cases these methods cannot be applied because either the state space is too large, a system model is available only as a simulator, or no system model is available at all. It is in these cases that RL techniques and algorithms may be helpful.

Reinforcement learning solutions can be viewed as a broad class of sample-based methods for solving MDPs. In place of a model, these methods use sample trajectories of the system and the agent interacting, such as could be obtained from a simulation. It is not unusual in practical applications for such a simulator to be available when an explicit transition-probability model of the sort suitable for use by dynamic or linear programming is not~\citep{Tesauro94TS,Crites98EG}. Reinforcement learning methods can also be used with no model at all, by obtaining sample trajectories from direct interaction with the system~\citep{Baxter98KC,Kohl04PG,Ng04AH}.

Reinforcement learning solutions can be categorized in different ways. Below we describe two common categorizations that are relevant to the structure of this paper.\\

{\bf Model-based vs.~Model-free Methods:} Reinforcement learning algorithms that explicitly learn a system model and use it to solve an MDP are called {\em model-based} methods. Some examples of these methods are the Dyna architecture~\citep{Sutton91DI} and prioritized sweeping~\citep{Moore93PS}. {\em Model-free} methods are those that do not explicitly learn a system model and only use sample trajectories obtained by direct interaction with the system. These methods include popular algorithms such as  Q-learning~\citep{Watkins89LD}, SARSA~\citep{Rummery94OQ}, and LSPI~\citep{Lagoudakis03LS}. \\

{\bf Value Function vs.~Policy Search Methods:} An important class of RL algorithms are those that first find the optimal value function, and then extract an optimal policy from it. This class of RL algorithms contains {\em value function} methods, of which some well-studied examples include value iteration (e.g.,~\citealt{Bertsekas96NP,Sutton98IR}), policy iteration (e.g.,~\citealt{Bertsekas96NP,Sutton98IR}), Q-learning~\citep{Watkins89LD}, SARSA~\citep{Rummery94OQ}, LSPI~\citep{Lagoudakis03LS}, and fitted Q-iteration~\citep{Ernst05TB}. An alternative approach for solving an MDP is to directly search in the space of policies. These RL algorithms are called {\em policy search} methods. Since the number of policies is exponential in the size of the state space, one has to resort to meta-heuristic search \citep{CrossEntropyPolicySearch} or to local greedy methods.
An important class of policy search methods is that of {\em policy gradient} algorithms. In these algorithms, the policy is taken to be an arbitrary differentiable function of a parameter vector, and the search in the policy space is directed by the gradient of a performance function with respect to the policy parameters (e.g.,~\citealt{Williams92SS,Marbach98SM,Baxter01IP}). There is a third class of RL methods that use policy gradient to search in the policy space, and at the same time estimate a value function. These algorithms are called {\em actor-critic} (e.g.,~\citealt{Konda00AA,Sutton00PG,Bhatnagar07IN,Peters08NA,Bhatnagar09NA}). They can be thought of as RL analogs of dynamic programming's policy iteration method. Actor-critic methods are based on the simultaneous online estimation of the parameters of two structures, called the {\em actor} and the {\em critic}. The actor and the critic correspond to conventional action-selection policy and value function, respectively. These problems are separable, but are solved simultaneously to find an optimal policy.


\section{Bayesian Learning}
\label{ss:bl}

In Bayesian learning, we make inference about a random variable $X$ by producing a probability distribution for $X$. Inferences, such as point and interval estimates, may then be extracted from this distribution. Let us assume that the random variable $X$ is hidden and we can only observe a related random variable $Y$. Our goal is to infer $X$ from the samples of $Y$. A simple example is when $X$ is a physical quantity and $Y$ is its noisy measurement. Bayesian inference is usually carried out in the following way:
%
\begin{enumerate}
\item We choose a probability density $P(X)$, called the {\em prior distribution}, that expresses our beliefs about the random variable $X$ before we observe any data.
\item We select a statistical model $P(Y|X)$ that reflects our belief about $Y$ given $X$. This model represents the statistical dependence between $X$ and $Y$.
\item We observe data $Y=y$.
\item We update our belief about $X$ by calculating its posterior distribution using Bayes rule
\begin{equation*}
P(X|Y=y)=\frac{P(y|X)P(X)}{\int P(y|X')P(X')dX'}\;.
\end{equation*}
%
\end{enumerate}
\noindent

Assume now that $P(X)$ is parameterized by an unknown vector of parameters $\vectheta$ in some parameter space $\Theta$; we denote this as $P_{\vectheta}(X)$.  Let $X_1,\ldots,X_n$ be a random i.i.d.~sample drawn from $P_{\vectheta}(X)$. In general, updating the posterior $P_{\vectheta}(X|Y=y)$ is difficult, due to the need to compute the normalizing constant $\int_\Theta P(y|X')P_{\vectheta}(X')d\vectheta$. However, for the case of \emph{conjugate} family distributions, we can update the posterior in closed-form by simply updating the parameters of the distribution. In the next two sections, we consider three classes of conjugate distributions: Beta and Dirichlet distributions and Gaussian Processes (GPs).


\subsection{Beta and Dirichlet Distributions}
\label{sss:Dir}

A simple example of a conjugate family is the Beta distribution, which is conjugate to the Binomial distribution. Let $Beta(\alpha,\beta)$ be defined by the density function $f(p|\alpha,\beta) \propto p^{\alpha-1}(1-p)^{\beta-1}$ for $p \in [0,1]$, and parameters $\alpha,\beta \geq 0$. Let $Binomial(n,p)$ be defined by the density function $f(k|n,p) \propto p^{k}(1-p)^{n-k}$ for $k \in \{0,1,\dots,n\}$, and parameters $p \in [0,1]$ and $n \in \mathbb{N}$. Consider $X \sim Binomial(n,p)$ with unknown probability parameter $p$, and consider a prior $Beta(\alpha,\beta)$ over the unknown value of $p$. Then following an observation $X=x$, the posterior over $p$ is also Beta distributed and is defined by $Beta(\alpha+x,\beta+n-x)$.

Now let us consider the multivariate extension of this conjugate family. In this case, we have the Multinomial distribution, whose conjugate is the Dirichlet distribution. Let $X \sim Multinomial_k(p,N)$ be a random variable with unknown probability distribution $p=(p_1, \ldots, p_k)$.  The Dirichlet distribution is parameterized by a count vector $\phi = (\phi_1, \ldots, \phi_k)$, where $\phi_i \geq 0$, such that the density of probability distribution $p = (p_1, \ldots, p_k)$ is defined as $f(p|\phi) \propto \prod_{i = 1}^k p_i^{\phi_i - 1}$.  The distribution $Dirichlet(\phi_1, \ldots, \phi_k)$ can also be interpreted as a prior over the unknown Multinomial distribution $p$, such that after observing $X=\textbf{n}$, the posterior over $p$ is also Dirichlet and is defined by $Dirichlet(\phi_1 + n_1, \ldots, \phi_k + n_k)$.


\subsection{Gaussian Processes and Gaussian Process Regression}
\label{sss:GP}

A Gaussian process (GP) is an indexed set of jointly Gaussian random variables, i.e., $F(x),\;x\in\X$ is a Gaussian process if and only if for every finite set of indices $\{x_1,\ldots,x_T\}$ in the index set $\X$, $\big(F(x_1),\ldots,F(x_T)\big)$ is a vector-valued Gaussian random variable. The GP $F$ may be thought of as a random vector if $\X$ is finite, as a random series if $\X$ is countably infinite, and as a random function if $\X$ is uncountably infinite. In the last case, each instantiation of $F$ is a function $f:\X\rightarrow\Re$. For a given $x$, $F(x)$ is a random variable, normally distributed jointly with the other components of $F$. A GP $F$ can be fully specified by its mean $\bar{f}:\X\rightarrow\Re,\;\bar{f}(x)=\exptE[F(x)]$ and covariance $k:\X\times\X\rightarrow\Re,\;k(x,x')=\Cov[F(x),F(x')]$, i.e., $F(\cdot)\sim\N\big(\bar f(\cdot),k(\cdot,\cdot)\big)$. The kernel function $k(\cdot,\cdot)$ encodes our prior knowledge concerning the correlations between the components of $F$ at different points. It may be thought of as inducing a measure of proximity between the members of $\X$. It can also be shown that $k$ defines the function space within which the search for a solution takes place (see~\citealt{Scholkopf02LK,ShaweTaylor04KM} for more details).

\begin{figure}[t]
\centering
\includegraphics[scale=0.45]{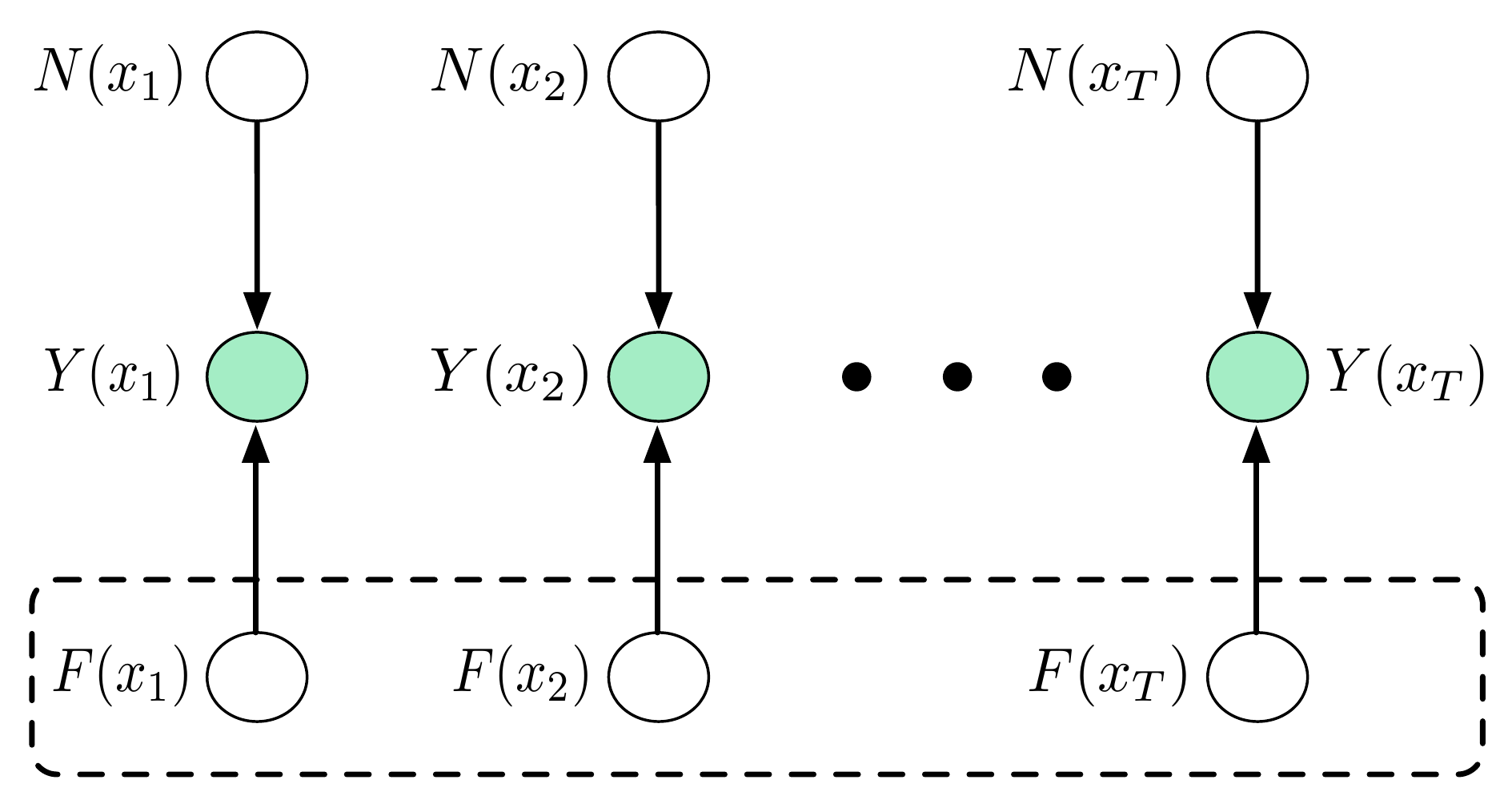}
\caption{A directed graph illustrating the conditional independencies between the latent $F(x_t)$ variables (bottom row), the noise variables $N(x_t)$ (top row), and the observable $Y(x_t)$ variables (middle row), in GP regression (when $\matH=\matI$). All of the $F(x_t)$ variables should be interconnected by arrows (forming a clique), due to the dependencies introduced by the prior. To avoid cluttering the diagram, this was marked by the dashed frame surrounding them.}
\label{fig:GP-reg}
\end{figure}

Now let us consider the following generative model:
\begin{equation}\label{eq:lin-stat-model1}
Y(x)=\matH F(x)+N(x),
\end{equation}
where $\matH$ is a general linear transformation, $F$ is an unknown function, $x$ is the input, $N$ is a Gaussian noise and independent of $F$, and $Y$ is the observable process, modeled as a noisy version of $\matH F$. The objective here is to infer $F$ from samples of $Y$. Bayesian learning can be applied to this problem in the following way.
%
\begin{enumerate}
\item We choose a prior distribution for $F$ in the form of a GP, $F(\cdot)\sim\N\big(\bar f(\cdot),k(\cdot,\cdot)\big)$. When $F$ and $N$ are Gaussian and independent of each other, the generative model of Eq.~\ref{eq:lin-stat-model1} is known as the {\em linear statistical model}~\citep{Scharf91SS}.
\item The statistical dependence between $F$ and $Y$ is defined by the model-equation~\eqref{eq:lin-stat-model1}.
\item We observe a sample in the form of $\D_T=\{(x_t,y_t)\}_{t=1}^T$.
\item We calculate the posterior distribution of $F$ conditioned on the sample $\D_T$ using the Bayes rule. Below is the process to calculate this posterior distribution.
\end{enumerate}
%
Figure~\ref{fig:GP-reg} illustrates the GP regression setting (when $\matH=\matI$) as a graphical model in which arrows mark the conditional dependency relations between the nodes corresponding to the latent $F(x_t)$ and the observed $Y(x_t)$ variables.
The model-equation~\eqref{eq:lin-stat-model1} evaluated at the training samples may be written as
\begin{equation}\label{eq:lin-stat-model2}
\vecY_T=\matH \vecF_T+\vecN_T,
\end{equation}
where $\vecF_T=\big(F(x_1),\ldots,F(x_T)\big)^\top$, $\vecY_T=(y_1,\ldots,y_T)^\top$, and $\vecN_T\sim\N(\vec0,\matSigma)$. Here $[\matSigma]_{i,j}$ is the measurement noise covariance between the $i$th and $j$th samples. In the linear statistical model, we then have $\vecF_T\sim\N(\bar{\vecf},\matK)$, where $\bar\vecf=\big(\bar{f}(x_1),\ldots,\bar{f}(x_T)\big)^\top$ and $[\matK]_{i,j}=k(x_i,x_j)$ is a $T\times T$ kernel matrix. Since both $\vecF_T$ and $\vecN_T$ are Gaussian and independent from each other, we have $\vecY_T\sim\N(\matH\bar\vecf,\matH\matK\matH^\top+\matSigma)$. Consider a query point $x$, we then have
\begin{equation*}
\begin{pmatrix}
F(x) \\
\vecF_T \\
\vecN_T
\end{pmatrix}
=\N\left\{
\begin{pmatrix}
\bar f(x) \\
\bar\vecf \\
\vec0
\end{pmatrix}
,
\begin{bmatrix}
k(x,x) & \veck(x)^\top & \vec0 \\
\veck(x) & \matK & \vec0 \\
\vec0 & \vec0 & \matSigma
\end{bmatrix}
\right\},
\end{equation*}
where $\veck(x)=\big(k(x_1,x),\ldots,k(x_T,x)\big)^\top$. Using Eq.~\ref{eq:lin-stat-model2}, we have the following transformation:
\begin{equation}\label{eq:lin-stat-model3}
\begin{pmatrix}
F(x) \\
\vecF_T \\
\vecY_T
\end{pmatrix}
=
\begin{bmatrix}
1 & 0 & 0 \\
0 & \matI & \vec0 \\
0 & \matH & \matI
\end{bmatrix}
\begin{pmatrix}
F(x) \\
\vecF_T \\
\vecN_T
\end{pmatrix},
\end{equation}
where $\matI$ is the identity matrix. From Eq.~\ref{eq:lin-stat-model3}, we have
\begin{equation*}
\begin{pmatrix}
F(x) \\
\vecF_T \\
\vecY_T
\end{pmatrix}
=\N\left\{
\begin{pmatrix}
\bar f(x) \\
\bar\vecf \\
\matH\bar\vecf
\end{pmatrix}
,
\begin{bmatrix}
k(x,x) & \veck(x)^\top & \veck(x)^\top\matH^\top \\
\veck(x) & \matK & \matK\matH^\top \\
\matH\veck(x) & \matH\matK & \matH\matK\matH^\top+\matSigma
\end{bmatrix}
\right\}.
\end{equation*}
Using the Gauss-Markov theorem~\citep{Scharf91SS}, we know that $F(x)|\vecY_T$ (or equivalently $F(x)|\D_T$) is Gaussian, and obtain the following expressions for the posterior mean and covariance of $F(x)$ conditioned on the sample $\D_T$:
\begin{align}\label{eq:lin-stat-mod4}
\exptE\big[F(x)|\D_T\big]&=\bar f(x)+\veck(x)^\top\matH^\top(\matH\matK\matH^\top+\matSigma)^{-1}(\vecy_T-\matH\bar\vecf), \nonumber \\
\Cov\big[F(x),F(x')|\D_T\big]&=k(x,x')-\veck(x)^\top\matH^\top(\matH\matK\matH^\top+\matSigma)^{-1}\matH\veck(x'),
\end{align}
where $\vecy_T=(y_1,\ldots,y_T)^\top$ is one realization of the random vector $\vecY_T$. It is possible to decompose the expressions in Eq.~\ref{eq:lin-stat-mod4} into input dependent terms (which depend on $x$ and $x'$) and terms that only depend on the training samples, as follows:
\begin{align}\label{eq:lin-stat-mod5}
\exptE\big[F(x)|\D_T\big]&=\bar f(x)+\veck(x)^\top\vecalpha, \nonumber \\
\Cov\big[F(x),F(x')|\D_T\big]&=k(x,x')-\veck(x)^\top\matC\veck(x'),
\end{align}
where
\begin{equation}\label{eq:lin-stat-mod6}
\begin{split}
\vecalpha&=\matH^\top(\matH\matK\matH^\top+\matSigma)^{-1}(\vecy_T-\matH\bar\vecf),\\
\matC&=\matH^\top(\matH\matK\matH^\top+\matSigma)^{-1}\matH.
\end{split}
\end{equation}
From Eqs.~\ref{eq:lin-stat-mod5} and~\ref{eq:lin-stat-mod6}, we can conclude that $\vecalpha$ and $\matC$ are {\em sufficient statistics} for the posterior moments.

If we set the transformation $\matH$ to be the identity and assume that the noise terms corrupting each sample are i.i.d.~Gaussian, i.e.,~$\vecN_T\sim\N(\vec0,\sigma^2\matI)$, where $\sigma^2$ is the variance of each noise term, the linear statistical model is reduced to the {\em standard linear regression model}. In this case, the posterior moments of $F(x)$ can be written as
\begin{equation}\label{eq:nonparam-GPreg1}
\begin{split}
\exptE\big[F(x)|\D_T\big]&=\bar f(x)+\veck(x)^\top(\matK+\sigma^2\matI)^{-1}(\vecy_T-\bar\vecf)\\
&=\bar f(x)+\veck(x)^\top\vecalpha, \\
\Cov\big[F(x),F(x')|\D_T\big]&=k(x,x')-\veck(x)^\top(\matK+\sigma^2\matI)^{-1}\veck(x')\\
&=k(x,x')-\veck(x)^\top\matC\veck(x'),
\end{split}
\end{equation}
with
\begin{equation}\label{eq:nonparam-GPreg2}
\vecalpha=(\matK+\sigma^2\matI)^{-1}(\vecy_T-\bar\vecf),\hspace{0.5in}\matC=(\matK+\sigma^2\matI)^{-1}.
\end{equation}

The GP regression described above is kernel-based and non-parametric. It is also possible to employ a parametric representation under very similar assumptions. In the parametric setting, the GP $F$ is assumed to consist of a linear combination of a finite number $n$ of basis functions $\varphi_i:\X\rightarrow\Re,\;i=1,\ldots,n$. In this case, $F$ can be written as $F(\cdot)=\sum_{i=1}^n\varphi_i(\cdot)W_i=\vecphi(\cdot)^\top \vecW$, where $\vecphi(\cdot)=\big(\varphi_1(\cdot),\ldots,\varphi_n(\cdot)\big)^\top$ is the feature vector and $\vecW=(W_1,\ldots,W_n)^\top$ is the weight vector. The model-equation~\eqref{eq:lin-stat-model1} now becomes
\begin{equation*}
Y(x)=\matH\vecphi(x)^\top\vecW+N(x).
\end{equation*}
In the parametric GP regression, the randomness in $F$ is due to $\vecW$ being a random vector. Here we consider a Gaussian prior over $\vecW$, distributed as $\vecW\sim\N(\bar\vecw,\matS_\vecw)$. By applying the Bayes rule, the posterior (Gaussian) distribution of $\vecW$ conditioned on the observed data $\D_T$ can be computed as
\begin{align}\label{eq:param-LSM1}
\exptE\big[\vecW|\D_T\big]&=\bar\vecw+\matS_\vecw\matPhi\matH^\top(\matH\matPhi^\top\matS_\vecw\matPhi\matH^\top+\matSigma)^{-1}(\vecy_T-\matH\matPhi^\top\bar\vecw), \nonumber \\
\Cov\big[\vecW|\D_T\big]&=\matS_\vecw-\matS_\vecw\matPhi\matH^\top(\matH\matPhi^\top\matS_\vecw\matPhi\matH^\top+\matSigma)^{-1}\matH\matPhi^\top\matS_\vecw,
\end{align}
where $\matPhi=\big[\vecphi(x_1),\ldots,\vecphi(x_T)\big]$ is a $n\times T$ feature matrix. Finally, since $F(x)=\vecphi(x)^\top\vecW$, the posterior mean and covariance of $F$ can be easily computed as

\begin{small}
\begin{equation}\label{eq:param-LSM2}
\begin{split}
\exptE\big[F(x)|\D_T\big]&=\vecphi(x)^\top\bar\vecw\\
&+\vecphi(x)^\top\matS_\vecw\matPhi\matH^\top\!\!(\matH\matPhi^\top\matS_\vecw\matPhi\matH^\top\!\!\!\!+\!\!\matSigma)^{-1}(\vecy_T-\matH\matPhi^\top\bar\vecw), \\
\Cov\big[F(x),F(x')|\D_T\big]&=\vecphi(x)^\top\matS_\vecw\vecphi(x')\\
&-\vecphi(x)^\top\matS_\vecw\matPhi\matH^\top\!\!(\matH\matPhi^\top\matS_\vecw\matPhi\matH^\top\!\!\!\!+\!\!\matSigma)^{-1}\matH\matPhi^\top\matS_\vecw\vecphi(x').
\end{split}
\end{equation}
\end{small}

\noindent
Similar to the non-parametric setting, for the standard linear regression model with the prior $\vecW\sim\N(\vec0,\matI)$, Eq.~\ref{eq:param-LSM1} may be written as
\begin{equation}\label{eq:param-GPreg1}
\begin{split}
\exptE\big[\vecW|\D_T\big]&=\matPhi(\matPhi^\top\matPhi+\sigma^2\matI)^{-1}\vecy_T,\\
\Cov\big[\vecW|\D_T\big]&=\matI-\matPhi(\matPhi^\top\matPhi+\sigma^2\matI)^{-1}\matPhi^\top.
\end{split}
\end{equation}
To have a smaller matrix inversion when $T>n$, Eq.~\ref{eq:param-GPreg1} may be written as
\begin{equation}\label{eq:param-GPreg2}
\begin{split}
\exptE\big[\vecW|\D_T\big]&=(\matPhi\matPhi^\top+\sigma^2\matI)^{-1}\matPhi\vecy_T,\\
\Cov\big[\vecW|\D_T\big]&=\sigma^2(\matPhi\matPhi^\top+\sigma^2\matI)^{-1}.
\end{split}
\end{equation}
For more details on GPs and GP regression, see~\citet{Rasmussen06GP}.




\chapter{Bayesian Bandits}
\label{s:Bayesian-bandits}
\vspace{-10pt}
This section focuses on Bayesian learning methods for regret minimization in the multi-armed bandits (MAB) model. We review classic performance bounds for this problem and state-of-the-art results for several Bayesian approaches.

In the MAB model (\S\ref{s:MAB}), the only unknown quantities are the outcome probabilities $P(\cdot|a)$. The idea behind Bayesian approaches to MAB is to use Bayesian inference (\S\ref{ss:bl}) for learning $P(\cdot|a)$ from the outcomes observed during sequential interaction with the MAB. Following the framework of \S\ref{ss:bl}, the outcome probabilities are parameterized by an unknown vector $\vectheta$, and are henceforth denoted by $P_\vectheta(\cdot|a)$. The parameter vector $\vectheta$ is assumed to be drawn from a prior distribution $P_{\text{prior}}$.
As a concrete example, consider a MAB with Bernoulli arms:
\begin{example}[Bernoulli $K$-MAB with a Beta prior]\label{exm:Bernoulli_MAB}
Consider a $K$-MAB where the set of outcomes is binary $\Y = \{0,1\}$, and the reward is the identity $r(Y) = Y$. Let the outcome probabilities be parameterized by $\vectheta \in \Re^K$, such that $Y(a) \sim \textrm{Bernoulli}[\vectheta_{a}]$. The prior for each $\vectheta_{a}$ is $Beta(\alpha_a,\beta_a)$. Following an observation outcome $Y(a)=y$, the posterior for $\vectheta_{a}$ is updated to $Beta(\alpha_a+y,\beta_a+y-1)$. Also note that the (posterior) expected reward for arm $a$ is now $\exptE \left[ r(Y(a)) \right] = \exptE \left[ \exptE \left[ \left. Y(a)\right| \vectheta_{a}\right]\right] = \exptE \left[ \vectheta_{a} \right] = \frac{\alpha_a+y}{\alpha_a+y + \beta_a+y-1}$.
\end{example}
The principal challenge of Bayesian MAB algorithms, however, is to utilize the posterior $\vectheta $ for selecting an adequate policy that achieves low regret.

Conceptually, there are several reasons to adopt a Bayesian approach for MABs. First, from a modelling perspective, the prior for $\vectheta$ is a natural means for embedding prior knowledge, or structure of the problem. Second, the posterior for $\vectheta$ encodes the uncertainty about the outcome probabilities at each step of the algorithm, and may be used for guiding the exploration to more relevant areas.

A Bayesian point of view may also be taken towards the performance measure of the algorithm.
Recall that the performance of a MAB algorithm is measured by its expected regret. This expectation, however, may be defined in two different ways, depending on how the parameters $\vectheta$ are treated. In the frequentist approach, henceforth termed \emph{`frequentist regret'} and denoted $\exptE_\vectheta\big[\regret (T)\big]$, the parameter vector $\vectheta$ is fixed, and treated as a constant in the expectation. The expectation is then computed with respect to the stochastic outcomes and the action selection policy. On the other hand, in \emph{`Bayesian regret'}, a.k.a.~\emph{Bayes risk}, $\vectheta$ is assumed to be drawn from the prior, and the expectation $\exptE\big[\regret (T)\big]$ is over the stochastic outcomes, the action selection rule, and the prior distribution of $\vectheta$. Note that the optimal action $a^*$ depends on $\vectheta$. Therefore, in the expectation it is also considered a random variable.

We emphasize the separation of Bayesian MAB algorithms and Bayesian regret analysis. In particular, the performance of a Bayesian MAB algorithm (i.e., an algorithm that uses Bayesian learning techniques) may be measured with respect to a frequentist regret, or a Bayesian one.

\section{Classical Results}

In their seminal work, Lai and Robbins~\citet{Lai85AE} proved asymptotically tight bounds on the frequentist regret in terms of the Kullback-Leibler (KL) divergence between the distributions of the rewards of the different arms. These bounds grow logarithmically with the number of steps $T$,  such that regret is $\O\left( \ln T \right)$. Mannor and Tsitsiklis~\citet{Mannor04SC} later showed non-asymptotic lower bounds with a similar logarithmic dependence on $T$. For the Bayesian regret, the lower bound on the regret is $\O\left( \sqrt{KT}\right)$ (see, e.g.,~Theorem 3.5 of Bubeck and Cesa-Bianchi~\citealt{Bubeck12RA}).

In the Bayesian setting, and for models that admit sufficient statistics, Gittins~\citet{Gittins79BP} showed that an optimal strategy may be found by solving a specific MDP planning problem.
The key observation here is that the dynamics of the posterior for each arm may be represented by a special MDP termed a \emph{bandit process} \citep{Gittins79BP}.

\begin{definition}[Bandit Process]
A bandit process is an MDP with two actions $\A = \{0,1\}$. The control $0$ freezes the process in the sense that $P(s'=s|s,0)=1$, and $R(s,0)=0$. Control $1$ continues the process, and induces a standard MDP transition to a new state with probability $P(\cdot|s,1)$, and reward $R(s,1)$.
\end{definition}
For example, consider the case of Bernoulli bandits with a Beta prior as described in Example \ref{exm:Bernoulli_MAB}. We identify the state $s_a$ of the bandit process for arm $a$ as the posterior parameters $s_a = (\alpha_a,\beta_a)$. Whenever arm $a$ is pulled, the continuation control is applied as follows: we draw some $\vectheta_{a}$ from the posterior, and then draw an outcome $Y \sim \textrm{Bernoulli}[\vectheta_{a}]$. The state subsequently transitions to an updated posterior (cf. Example \ref{exm:Bernoulli_MAB}) $s_a' = (\alpha_a+Y,\beta_a+Y-1)$, and a reward of $\exptE \left[ r(Y) \right]$ is obtained, where the expectation is taken over the posterior.

The $K$-MAB problem, thus, may be seen as a particular instance of a general model termed \emph{simple family of alternative bandit processes} (SFABP) \citealt{Gittins79BP}.
\begin{model}[(Simple Family of Alternative Bandit Processes)]
A simple family of alternative bandit processes is a set of $K$ bandit processes. For each bandit process $i\in 1,\dots,K$ we denote by $s_i$, $P_i(\cdot|s_i,1)$, and $R_i(s_i,1)$ the corresponding state, transition probabilities, and reward, respectively. At each time $t = 1,2,\dots$, a single bandit $i_t\in 1,\dots,K$ that is in state $s_{i_t}(t)$ is activated, by applying to it the continuation control, and all other bandits are frozen. The objective is to find a bandit selection policy that maximizes the expected total $\gamma$-discounted reward
\begin{equation*}\label{eq:Gittins_discounted_sum}
    \exptE \left[ \sum_{t=0}^{\infty} \gamma^t R_{i_t}(s_{i_t}(t),1)\right].
\end{equation*}
\end{model}

An important observation is that for the $K$-MAB problem, as defined above, the SFABP performance measure captures an expectation of the total discounted reward with respect to the \emph{prior} distribution of the parameters $\vectheta$. In this sense, this approach is a \emph{full} Bayesian $K$-MAB solution (cf. the Bayesian regret vs. frequentist regret discussion above). In particular, note that the SFABP performance measure implicitly balances exploration and exploitation. To see this, recall the Bernoulli bandits example, and note that the reward of each bandit process is the \emph{posterior} expected reward of its corresponding arm. Since the posterior distribution is encoded in the process state, future rewards in the SFABP performance measure depend on the future state of knowledge, thereby implicitly quantifying the value of additional observations for each arm.

The SFABP may be seen as a single MDP, with a state $(s_1,s_2,\dots,s_K)$ that is the conjunction of the $K$ individual bandit process states. Thus, naively, an optimal policy may be computed by solving this MDP. Unfortunately, since the size of the state space of this MDP is exponential in $K$, such a naive approach is intractable. The virtue of the celebrated Gittins index theorem~\citep{Gittins79BP}, is to show that this problem nevertheless has a tractable solution.

\begin{theorem} \citep{Gittins79BP}\label{thm:Gittins_Idx}
The objective of SFABP is maximized by following a policy that always chooses the bandit with the largest Gittins index
\begin{equation*}
    G_i(s_i) = \sup_{\tau \geq 1} \frac{\exptE \left[ \left. \sum_{t=0}^{\tau - 1} \gamma^t R_{i}(s_{i}(t),1)\right| s_{i}(0) = s_i\right]}{\exptE \left[ \left. \sum_{t=0}^{\tau - 1} \gamma^t \right| s_{i}(0) = s_i\right]},
\end{equation*}
where $\tau$ is any (past measurable) stopping time.
\end{theorem}

The crucial advantage of Theorem \ref{thm:Gittins_Idx}, is that calculating $G_i$ may be done \emph{separately} for each arm, thereby avoiding the exponential complexity in $K$. The explicit calculation, however, is technically involved, and beyond the scope of this survey. For reference see \citet{Gittins79BP}, and also \citet{tsitsiklis1994short} for a simpler derivation, and \citet{Nino2011CC} for a finite horizon setting.

Due to the technical complexity of calculating optimal Gittin's index policies, recent approaches concentrate on much simpler algorithms, that nonetheless admit optimal upper bounds (i.e., match the order of their respective lower bounds up to constant factors) on the expected regret.

\section{Bayes-UCB}

The upper confidence bound (UCB) algorithm~\citep{Auer2002FA} is a popular frequentist approach for MABs that employs an `optimistic' policy to reduce the chance of overlooking the best arm. The algorithm starts by playing each arm once. Then, at each time step $t$, UCB plays the arm $a$ that maximizes $<r_a>+\sqrt{\frac{2 \ln t}{t_a}}$, where $<r_a>$ is the average reward obtained from arm $a$, and $t_a$ is the number of times arm $a$ has been playing so far. The optimistic upper confidence term $\sqrt{\frac{2 \ln t}{t_a}}$ guarantees that the empirical average does not underestimate the best arm due to `unlucky' reward realizations.

The Bayes-UCB algorithm of Kaufmann et al.~\citet{Kaufmann12BU} extends the UCB approach to the Bayesian setting. A posterior distribution over the expected reward of each arm is maintained, and at each step, the algorithm chooses the arm with the maximal posterior $(1-\beta_t)$-quantile, where $\beta_t$ is of order $1/t$. Intuitively, using an upper quantile instead of the posterior mean serves the role of `optimism', in the spirit of the original UCB approach. For the case of Bernoulli distributed outcomes and a uniform prior on the reward means, Kaufmann et al.~\citet{Kaufmann12BU} prove a \emph{frequentist} upper bound on the expected regret that matches the lower bound of Lai and Robbins~\citet{Lai85AE}.



\section{Thompson Sampling}
\label{ss:thompson}

The Thompson Sampling algorithm (TS) suggests a natural Bayesian approach to the MAB problem using randomized probability matching~\citep{Thompson33TS}. Let $P_{\text{post}}$ denote the posterior distribution of $\vectheta$ given the observations up to time $t$. In TS, at each time step $t$, we sample a parameter $\hat{\vectheta}$ from the posterior and select the optimal action with respect to the model defined by $\hat{\vectheta}$. Thus, effectively, we match the action selection probability to the posterior probability of each action being optimal.
The outcome observation is then used to update the posterior $P_{\text{post}}$.

\begin{algorithm}[ht]
\caption{\label{alg:ts} Thompson Sampling}
\begin{algor}[1]
\item [{*}]{\bf TS}$(P_{\text{prior}})$ \\
$\bullet$ $P_{\text{prior}}$ prior distribution over $\vectheta$\\
\item [{*}] $P_{\text{post}}:=P_{\text{prior}}$
\item [for] $t=1,2,\dots$
\item [{*}] Sample $\hat{\vectheta}$ from $P_{\text{post}}$
\item [{*}] Play arm $a_t = \argmax_{a \in \A} \exptE_{y \sim P_{\hat{\vectheta}}(\cdot|a)} \left[ r(y)\right]$
\item [{*}] Observe outcome $Y_t$ and update $P_{\text{post}}$
\item [endfor]
\end{algor}
\end{algorithm}

Recently, the TS algorithm has drawn considerable attention due to its state-of-the-art empirical performance~\citep{Scott10MB,Chapelle11EE}, which also led to its use in several industrial applications~\citep{Graepel10WB,Tang13AA}. We survey several theoretical studies that confirm TS is indeed a sound MAB method with state-of-the-art performance guarantees.

We mention that the TS idea is not limited to MAB problems, but can be seen as a general sampling technique for Bayesian learning, and has been applied also to contextual bandit and RL problems, among others \citep{strens00,osband13,abbasi15}. In this section, we present results for the MAB and contextual bandit settings, while the RL related results are given in Chapter \ref{s:model-based-brl}.

Agrawal and Goyal~\citet{Agrawal11AT} presented \emph{frequentist} regret bounds for TS. Their analysis is specific to Bernoulli arms with a uniform prior, but they show that by a clever modification of the algorithm it may also be applied to general arm distributions. Let $\Delta_a = \exptE_\vectheta\big[ r\big(Y(a^*)\big) - r\big(Y(a)\big) \big]$ denote the difference in expected reward between the optimal arm and arm $a$, when the parameter is $\vectheta$.

\newpage
\begin{theorem} \citep{Agrawal11AT}\label{thm:TS_Agarwal}
For the K-armed stochastic bandit problem, the TS algorithm has (frequentist) expected regret
\begin{equation*}
    \exptE_\vectheta\big[\regret (T)\big] \leq \O\Bigg( \bigg(\sum_{a \neq a^*}\frac{1}{\Delta_a^2}\bigg)^2 \ln T\Bigg).
\end{equation*}
\end{theorem}

Improved upper bounds were later presented \citet{Kaufmann12TS} for the specific case of Bernoulli arms with a uniform prior. These bounds are (order) optimal, in the sense that they match the lower bound of~\citet{Lai85AE}. Let $\KL_\vectheta(a,a^*)$ denote the Kullback-Leibler divergence between Bernoulli distributions with parameters $\exptE_\vectheta\big[ r(Y(a)) \big]$ and $\exptE_\vectheta\big[ r(Y(a^*))\big]$.

\begin{theorem} \citep{Kaufmann12TS}\label{thm:TS_Kaufmann}
For any $\epsilon > 0$, there exists a problem-dependent constant $C(\epsilon, \vectheta)$ such that the regret of TS satisfies
\begin{equation*}
    \exptE_\vectheta\big[\regret (T)\big] \leq \left(1+\epsilon\right) \sum_{a \neq a^*}\frac{\Delta_a\big(\ln(T)+\ln\ln(T)\big)}{\KL_\vectheta\left(a,a*\right)} + C(\epsilon, \vectheta).
\end{equation*}
\end{theorem}

More recently, Agrawal and Goyal~\citet{agrawal2013further} showed a \emph{problem independent} frequentist regret bound of order $\O\left( \sqrt{KT \ln T}\right)$, for the case of Bernoulli arms with a Beta prior. This bound holds \emph{for all} $\vectheta$, and implies that a Bayesian regret bound with a similar order holds; up to the logarithmic factor, this bound is order-optimal.

Liu and Li~\citet{Liu2015prior} investigated the importance of having an \emph{informative} prior in TS. Consider the special case of two-arms and two-models, i.e.,~$K=2$, and $\vectheta\in\{\theta_{\text{true}},\theta_{\text{false}}\}$, and assume that $\theta_{\text{true}}$ is the true model parameter. When $P_{\text{prior}}\left(\theta_{\text{true}}\right)$ is small, the \emph{frequentist} regret of TS is upper-bounded by $\O\left( \sqrt{\frac{T}{P_{\text{prior}}\left(\theta_{\text{true}}\right)}}\right)$, and when $P_{\text{prior}}\left(\theta_{\text{true}}\right)$ is sufficiently large, the regret is upper-bounded by $\O\left( \sqrt{(1-P_{\text{prior}}\left(\theta_{\text{true}}\right))T}\right)$ \citep{Liu2015prior}. For the special case of two-arms and two-models, regret lower-bounds of matching orders are also provided in \citep{Liu2015prior}. These bounds show that an informative prior, i.e.,~a large $P_{\text{prior}}\left(\theta_{\text{true}}\right)$, significantly impacts the regret.

One appealing property of TS is its natural extension to cases with structure or dependence between the arms. For example, consider the following extension of the online shop domain:

\begin{example}[Online Shop with Multiple Product Suggestions]
Consider the bandit setting of the online shop domain, as in Example \ref{exm:online_shop_bandit}. Instead of presenting to the customer a single product suggestion, the decision-maker may now suggest $M$ different product suggestions $a_1,\dots,a_M$ from a pool of suggestions $\I$. The customer, in turn, decides whether or not to buy each product, and the reward is the sum of profits from all items bought.

Naively, this problem may be formalized as a MAB with the action set, $\A$, being the set of all possible combinations of $M$ elements from $\I$.
In such a formulation, it is clear that the outcomes for actions with overlapping product suggestions are correlated.
\end{example}

In TS, such dependencies between the actions may be incorporated by simply updating the posterior. Recent advances in Bayesian inference such as particle filters and Markov Chain Monte-Carlo (MCMC) provide efficient numerical procedures for complex posterior updates. Gopalan et al.~\citet{Gopalan14TS} presented frequentist high-probability regret bounds for TS with general reward distributions and priors, and correlated actions. Let $N_T (a)$ denote the play count of arm $a$ until time $T$ and let $D_{\hat{\vectheta}} \in \Re^{|\A|}$ denote a vector of Kullback-Leibler divergences $D_{\hat{\vectheta}}\left(a\right) = \KL\left( P_\vectheta(\cdot|a) \| P_{\hat{\Theta}}(\cdot|a) \right)$, where $\vectheta$ is the true model in a frequentist setting. For each action $a\in\A$, we define $S_a$ to be the set of model parameters $\vectheta$ in which the optimal action is $a$. Within $S_a$, let $S'_a$ be the set of models $\hat{\vectheta}$ that exactly match the true model $\vectheta$ in the sense of the marginal outcome distribution for action $a$: $\KL\left( P_\vectheta(\cdot|a) \| P_{\hat{\vectheta}}(\cdot|a) \right)=0$. Moreover, let $e^{(j)}$ denote the $j$-th unit vector in a finite-dimensional Euclidean space. The following result holds under the assumption of finite action and observation spaces, and that the prior has a finite support with a positive probability for the true model $\vectheta$ (`grain of truth' assumption).

\begin{theorem} \citep{Gopalan14TS}\label{thm:TS_Gopalan}
For $\delta,\epsilon \in (0,1)$, there exists $T^*>0$ such that for all $T>T^*$, with probability at least $1-\delta$, $$\sum_{a\neq a^*} N_T (a) \leq B + C(\log T),$$ where $B\equiv B(\delta,\epsilon,\A,\Y,\vectheta,P_{\text{prior}})$ is a problem-dependent constant that does not depend on $T$ and
\begin{equation*}
\begin{split}
  C(\log T) := \max &\quad \sum_{k=1}^{K-1} z_k(a_k)\\
    s.t. &\quad z_k \in \Nat_+^{K-1} \times \{0\}, \quad a_k \in \A \setminus \{a^*\},\quad k<K,\\
    & \quad z_i \succeq z_k, \quad z_i(a_k) = z_k(a_k), \quad i\geq k, \\
    & \quad \forall j\geq 1, k\leq K-1: \\
    & \quad \quad \quad \quad \quad \quad \min_{\hat{\vectheta}\in S'_{a_k}} \left< z_k, D_{\hat{\vectheta}} \right> \geq \frac{1+\epsilon}{1-\epsilon} \log T, \\
    & \quad \quad \quad \quad \quad \quad \min_{\hat{\vectheta}\in S'_{a_k}} \left< z_k - e^{(j)}, D_{\hat{\vectheta}} \right> \geq \frac{1+\epsilon}{1-\epsilon} \log T. \\
\end{split}
\end{equation*}
\end{theorem}

As Gopalan et al.~\citet{Gopalan14TS} explain, the bound in Theorem~\ref{thm:TS_Gopalan} accounts for the dependence between the arms, and thus, provides tighter guarantees when there is information to be gained from this dependence. For example, in the case of selecting subsets of $M$ arms described earlier, calculating the term $C(\log T)$ in Theorem~\ref{thm:TS_Gopalan} gives a bound of $\O \big( (K-M) \log T\big)$, even though the total number of actions is $\binom{K}{M}$.

We proceed with an analysis of the Bayesian regret of TS. Using information theoretic tools, Russo and Van Roy~\citet{Russo14IT} elegantly bound the Bayesian regret, and investigate the benefits of having an informative prior. Let $p_t$ denote the distribution of the selected arm at time $t$. Note that by the definition of the TS algorithm, $p_t$ encodes the posterior distribution that the arm is optimal. Let $p_{t,a}(\cdot)$ denote the \emph{posterior} outcome distribution at time $t$, when selecting arm $a$, and let $p_{t,a}(\cdot|a^*)$ denote the posterior outcome distribution at time $t$, when selecting arm $a$, conditioned on the event that $a^*$ is the optimal action. Both $p_{t,a}(\cdot)$ and $p_{t,a}(\cdot|a^*)$ are random and depend on the prior and on the history of actions and observations up to time $t$. A key quantity in the analysis of~\citet{Russo14IT} is the \emph{information ratio} defined by
\begin{equation}
\label{eq:inf_ratio}
    \Gamma_t = \frac{\exptE_{a\sim p_t} \big[ \exptE_{y\sim p_{t,a}(\cdot|a)} r(y)- \exptE_{y\sim p_{t,a}(\cdot)} r(y)\big]}{\sqrt{\exptE_{a\sim p_t,a^*\sim p_t} \big[\KL\big(p_{t,a}(\cdot|a^*)\|p_{t,a}(\cdot)\big) \big]}}.
\end{equation}
Note that $\Gamma_t$ is also random and depends on the prior and on the history of the algorithm up to time $t$. As Russo and Van Roy~\citet{Russo14IT} explain, the numerator in Eq.~\ref{eq:inf_ratio} roughly captures how much knowing that the selected action is optimal influences the expected reward observed, while the denominator measures how much on average, knowing which action is optimal changes the observations at the selected action. Intuitively, the information ratio tends to be small when knowing which action is optimal significantly influences the anticipated observations at many other actions.

The following theorem relates a bound on $\Gamma_t$ to a bound on the Bayesian regret.
\begin{theorem} \citep{Russo14IT}\label{thm:TS_Russo}
For any $T\in \Nat$, if $\Gamma_t \leq \bar{\Gamma}$ almost surely for each $t\in \{1,\dots,T\}$, the TS algorithm satisfies
\begin{equation*}
\exptE\big[\regret (T)\big] \leq \bar{\Gamma} \sqrt{\Ent(p_1)T},
\end{equation*}
where $\Ent(\cdot)$ denotes the Shannon entropy.
\end{theorem}

Russo and Van Roy~\citet{Russo14IT} further showed that in general, $\Gamma_t \leq \sqrt{K / 2}$, giving an order-optimal upper bound of $\O\left(\sqrt{\frac{1}{2}\Ent(p_1)KT}\right)$. However, structure between the arms may be exploited to further bound the information ratio more tightly. For example, consider the case of linear optimization under bandit feedback where we have $\A \subset \Re^d$, and the reward satisfies $\exptE_{y \sim P_{\vectheta}(\cdot|a)}\left[ r(y)\right] = a^\top \vectheta$. In this case, an order-optimal bound of $\O\left(\sqrt{\frac{1}{2}\Ent(p_1)dT}\right)$ holds~\citet{Russo14IT}. It is important to note that the term $\Ent(p_1)$ is bounded by $\log (K)$, but in fact may be much smaller when there is an \emph{informative} prior available.

An analysis of TS with a slightly different flavour  was given by Guha and Munagala~\citet{guha2014stochastic}, who studied the \emph{stochastic regret} of TS, defined as the expected number of times a sub-optimal arm is chosen, where the expectation is Bayesian, i.e.,~taken with respect to $P_{\text{prior}}(\vectheta)$. For some horizon $T$, and prior $P_{\text{prior}}$, let $OPT(T,P_{\text{prior}})$ denote the stochastic regret of an optimal policy. Such a policy exists, and in principle may be calculated using dynamic programming (cf. the Gittins index discussion above). For the cases of two-armed bandits, and K-MABs with Bernoulli point priors, Guha and Munagala~\citet{guha2014stochastic} show that the stochastic regret of TS, labeled $TS(T,P_{\text{prior}})$ is a $2$-approximation of $OPT(T,P_{\text{prior}})$, namely $TS(T,P_{\text{prior}})\leq 2OPT(T,P_{\text{prior}})$. Interestingly, and in contrast to the asymptotic regret results discussed above, this result holds \emph{for all} $T$.

We conclude by noting that contextual bandits may be approached using Bayesian techniques in a very similar manner to the MAB algorithms described above. The only difference is that the unknown vector $\vectheta$ should now parameterize the distribution over actions and context, $P_\vectheta(\cdot|a,s)$. Empirically, the efficiency of TS was demonstrated in an online-advertising application of contextual bandits \citet{Chapelle11EE}. On the theoretical side, Agrawal and Goyal~\citet{agrawal2013thompson} study contextual bandits with rewards that linearly depend on the context, and show a frequentist regret bound of $\tilde{\O}\left( \frac{d^2}{\epsilon}\sqrt{T^{1+\epsilon}}\right)$, where $d$ is the dimension of the context vector, and $\epsilon$ is an algorithm-parameter that can be chosen in $(0,1)$. For the same problem, Russo and Van Roy~\citet{russo2014learning} derive a Bayesian regret bound of order $\O \left( d \sqrt{T} \ln T\right)$, which, up to logarithmic terms, matches the order of the $\O\left( d \sqrt{T}\right)$ lower bound for this problem \citep{rusmevichientong2010linearly}.

\chapter{Model-based Bayesian Reinforcement Learning}
\label{s:model-based-brl}

This section focuses on Bayesian learning methods that explicitly maintain a posterior over the model parameters and use this posterior to select actions.  Actions can be selected both for \emph{exploration} (i.e.,~achieving a better posterior), or \emph{exploitation} (i.e.,~achieving maximal return). We review basic representations and algorithms for model-based approaches, both in MDPs and POMDPs. We also present approaches based on sampling of the posterior, some of which provide finite sample guarantees on the learning process.


\section{Models and Representations}
\label{ss:belief-mdp}

Initial work on model-based Bayesian RL appeared in the control literature, under the topic of Dual Control~\citep{feldbaum61,filatov00}. The goal here is to explicitly represent uncertainty over the model parameters, $P, q$, as defined in \S\ref{ss:mdp}. One way to think about this problem is to see the parameters as unobservable states of the system, and to cast the problem of planning in an MDP with unknown parameters as planning under uncertainty using the POMDP formulation.  In this case, the belief tracking operation of Eq.~\ref{eq:belief} will keep a joint posterior distribution over model parameters and the true physical state of the system, and a policy will be derived to select optimal actions with respect to this posterior.

Let $\theta_{s,a,s'}$ be the (unknown) probability of transitioning from state $s$ to state $s'$ when taking action $a$, $\theta_{s,a,r}$ the (unknown) probability of obtaining reward $r$ when taking action $a$ at state $s$, and $\theta\in\Theta$ the set of all such parameters.   The belief $P_0(\theta)$ expresses our initial knowledge about the model parameters.  We can then compute $b_t(\theta)$, the belief after a $t$-step trajectory $\{s_{t+1}, r_{t}, a_{t}, s_{t}, \dots, s_1, r_0, a_0, s_0\}$. Considering a single observation $(s,a,r,s')$, we have
\begin{equation}
b_{t+1} (\theta') = \eta \Pr(s', r | s,a,\theta') \int_{\Scal,\Theta} \Pr(s',\theta'|s,a,\theta)b_{t}(\theta) dsd\theta \; ,
\label{eq:hyper-belief}
\end{equation}
where $\eta$ is the normalizing factor.  It is common to assume that the uncertainty between the parameters is independent, and thus compute $b_{t}(\theta)=\prod_{s,a}b_{t}(\theta_{s,a,s'})b_{t}(\theta_{s,a,r})$.

Recalling that a POMDP can be represented as a belief-MDP (see \S\ref{ss:pomdp}), a convenient way to interpret Eq.~\ref{eq:hyper-belief} is as an MDP where the state is defined to be a belief over the unknown parameters.  Theoretically, optimal planning in this representation can be achieved using MDP methods for continuous states, the goal being to express a policy over the belief space, and thus, over any posterior over model parameters. From a computational perspective, this is not necessarily a useful objective. First, it is computationally expensive, unless there are only a few unknown model parameters. Second, it is unnecessarily hard: indeed, the goal is to \emph{learn} (via the acquisition of data) an optimal policy that is robust to parameter uncertainty, not necessarily to pre-compute a policy for any possible parameterization. An alternative approach is to make specific assumptions about the structural form of the uncertainty over the model parameters, thereby allowing us to consider an alternate representation of the belief-MDP, one that is mathematically equivalent (under the structural assumptions), but more amenable to computational analysis.

An instance of this type of approach is the Bayes-Adaptive MDP (BAMDP) model~\citep{duff02}.  Here we restrict our attention to MDPs with discrete state and action sets. In this case, transition probabilities consist of Multinomial distributions. The posterior over the transition function can therefore be represented using a Dirichlet distribution, $P(\cdot|s,a) \sim \phi_{s,a}$.  For now, we assume that the reward function is known; extensions to unknown rewards are presented in \S\ref{ss:ext-rewards}.

\begin{model}[(Bayes-Adaptive MDP)]
Define a Bayes-Adaptive MDP $\M$ to be a tuple $\langle\Scal',\A,P',P_0',R'\rangle$ where\\
$\bullet$ $\Scal'$ is the set of hyper-states, $\Scal \times \Phi$,\\
$\bullet$ $\A$ is the set of actions,\\
$\bullet$ $P'(\cdot|s,\phi,a)$ is the transition function between hyper-states, conditioned on action $a$ being taken in hyper-state $(s,\phi)$,\\
$\bullet$ $P'_0\in\P(\Scal \times \Phi)$ combines the initial distribution over physical states, with the prior over transition functions $\phi_0$,\\
$\bullet$ $R'(s,\phi,a)=R(s,a)$ represents the reward obtained when action $a$ is taken in state $s$.
\end{model}

The BAMDP is defined as an extension of the conventional MDP model.  The state space of the BAMDP combines jointly the initial (physical) set of states $\Scal$, with the posterior parameters on the transition function $\Phi$. We call this joint space the \emph{hyper-state}. Assuming the posterior on the transition is captured by a Dirichlet distribution, we have  $\Phi = \{\phi_{s,a} \in \mathbb{N}^{|\Scal|}, \forall s,a \in\Scal \times A\}$.  The action set is the same as in the original MDP.  The reward function is assumed to be known for now, so it depends only on the physical (real) states and actions.

The transition model of the BAMDP captures transitions between hyper-states. Due to the nature of the state space, this transition function has a particular structure.   By the chain rule, $\Pr(s',\phi'|s,a,\phi) = \Pr(s'|s,a,\phi) \Pr(\phi'|s,a,s',\phi)$. The first term can be estimated by taking the expectation over all possible transition functions to yield $\frac{\phi_{s,a,s'}}{\sum_{s'' \in \Scal}\phi_{s,a,s''}}$. For the second term, since the update of the posterior $\phi$ to $\phi'$ is deterministic, $\Pr(\phi'|s,a,s',\phi)$ is $1$ if $\phi'_{s,a,s'} = \phi_{s,a,s'}+1$, and $0$, otherwise. Because these terms depend only on the previous hyper-state $(s,\phi_{s,a})$ and action $a$, transitions between hyper-states preserve the Markov property.  To summarize, we have (note that $\mathbb{I}(\cdot)$ is the {\em indicator} function):
\begin{equation}
P'(s',\phi'|s,a,\phi) = \frac{\phi_{s,a,s'}}{\sum_{s'' \in \Scal}\phi_{s,a,s''}} \mathbb{I}(\phi'_{s,a,s'}=\phi_{s,a,s'}+1).
\label{eq:bamdp-belief}
\end{equation}

It is worth considering the number of states in the BAMDP.  Initially (at $t=0$), there are only $|\Scal|$, one per real MDP, state (we assume a single prior $\phi_0$ is specified). Assuming a fully connected state space in the underlying MDP (i.e.,~$P(s'|s,a)>0, \forall s,a$), then at $t=1$ there are already $|\Scal| \times |\Scal|$ states, since $\phi \rightarrow \phi'$ can increment the count of any one of its $|\Scal|$ components. So at horizon $t$, there are $|\Scal|^t$ reachable states in the BAMDP. There are clear computational challenges in computing an optimal policy over all such beliefs.

Computational concerns aside, the value function of the BAMDP can be expressed using the Bellman equation
\begin{equation} \label{eq:ValFuncBAMDP}
\begin{split}
V^*(s,\phi) &= \max_{a \in \A} \left[ R'(s,\phi,a) + \gamma \sum_{(s',\phi') \in \Scal'} P'(s',\phi'|s,\phi,a) V^*(s',\phi') \right]\\
&= \max_{a \in \A} \left[ R(s,a) + \gamma \sum_{s' \in \Scal} \frac{\phi^a_{s,s'}}{\sum_{s'' \in \Scal}\phi^a_{s,s''}} V^*(s',\phi') \right].
\end{split}
\end{equation}

Let us now consider a simple example, the Chain problem, which is used extensively for empirical demonstrations throughout the literature on model-based BRL.

\begin{example}[The Chain problem]

The 5-state Chain problem~\citep{strens00}, shown in Figure~\ref{fig:chain}, requires the MDP agent to select between two abstract actions $\{1, 2\}$.  Action $1$ causes the agent to move to the right with probability $0.8$ (effect ``a'' in Figure~\ref{fig:chain}) and causes the agent to reset to the initial state with probability $0.2$ (effect ``b'' in Figure~\ref{fig:chain}). Action $2$ causes the agent to reset with probability $0.8$ (effect ``b'') and causes the agent to move to the right with probability $0.2$ (effect ``a''). The action $b$ has constant reward of $+2$. Rewards vary based on the state and effect (``a'' and ``b''), as shown in Figure~\ref{fig:chain}. The optimal policy is to always choose action $1$, causing the agent to potentially receive $+10$ several times until slipping back (randomly) to the initial state. Of course if the transition probabilities and rewards are not known, the agent has to trade-off exploration and exploitation to learn this optimal policy.

\begin{figure}[!htb]
\centering
\includegraphics[scale=0.6]{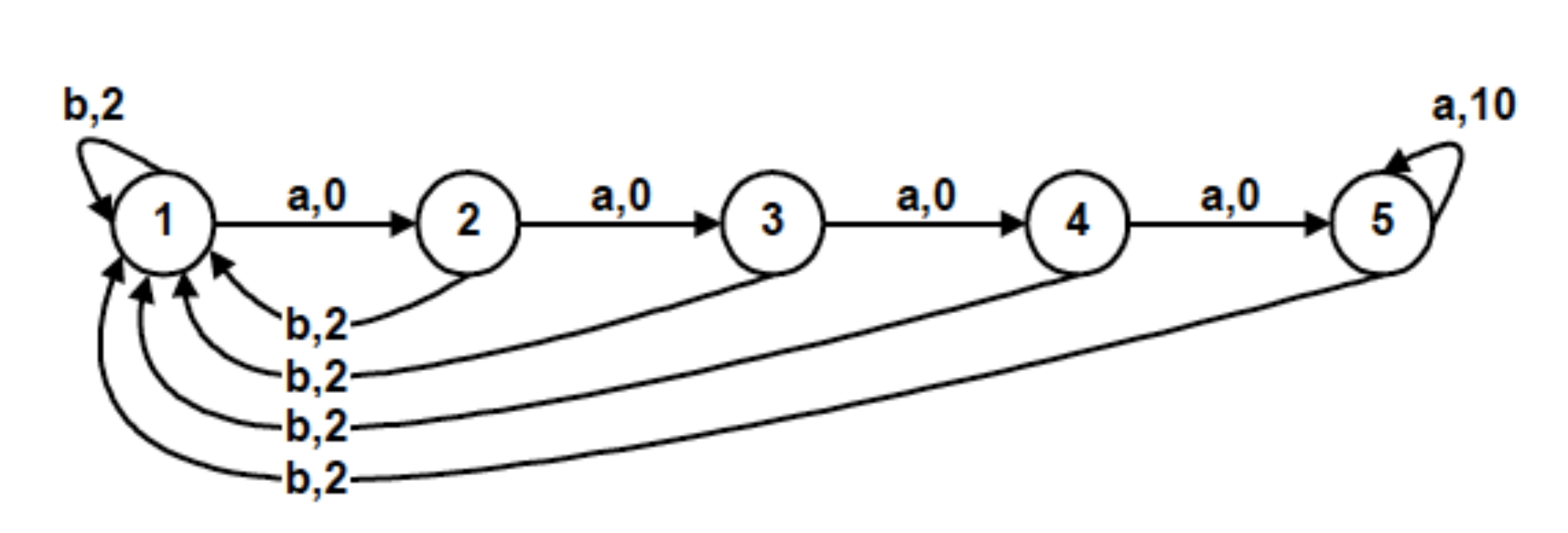}
\caption{The Chain problem (figure reproduced from~\citealt{strens00})}
\label{fig:chain}
\end{figure}

\end{example}


\section{Exploration/Exploitation Dilemma}
\label{ss:explore-exploit}

A key aspect of reinforcement learning is the issue of \emph{exploration}. This corresponds to the question of determining how the agent should choose actions while learning about the task. This is in contrast to the phase called \emph{exploitation}, through which actions are selected so as to maximize expected reward with respect to the current value function estimate.

In the Bayesian RL framework, exploration and exploitation are naturally balanced in a coherent mathematical framework. Policies are expressed over the full information state (or belief), including over model uncertainty. In that framework, the optimal Bayesian policy will be to select actions based on how much reward they yield, but also based on how much information they provide about the parameters of the domain, information which can then be leveraged to acquire even more reward.

\begin{definition}[Bayes optimality]
Let
\begin{equation*}
V^*_t(s,\phi)=\max_{a\in\A} \left[ R(s,a) + \gamma \int_{\Scal,\Phi} P(s'|b,s,a)V^*_{t-1}(s',\phi') ds' d\theta' \right]
\end{equation*}
be the optimal value function for a $t$-step planning horizon in a BAMDP.
\end{definition}

Any policy that maximizes this expression is called a \emph{Bayes-optimal} policy. In general, a Bayes-optimal policy is mathematically lower than the optimal policy (Eq.~\ref{eq:opt-Bellman-eq}), because it may require additional actions to acquire information about the model parameters. In the next section, we present planning algorithms that seek the Bayes-optimal policy; most are based on heuristics or approximations due to the computation complexity of the problem. In the following section, we also review algorithms that focus on a slightly different criteria, namely efficient (polynomial) learning of the optimal value function.



\section{Offline Value Approximation}
\label{ss:offline-alg}

We now review various classes of approximate algorithms for estimating the value function in the BAMDP.  We begin with \emph{offline} algorithms that compute the policy \emph{a priori}, for any possible state and posterior.   The goal is to compute an action-selection strategy that optimizes the expected return over the hyper-state of the BAMDP.  Given the size of the state space in the BAMDP, this is clearly intractable for most domains.  The interesting problem then is to devise approximate algorithms that leverage structural constraints to achieve computationally feasible solutions.


\subsection*{Finite-state controllers}

Duff \citet{duff01bamdpfsc} suggests using Finite-State Controllers (FSC) to compactly represent the optimal policy $\mu^*$ of a BAMDP, and then finding the best FSC in the space of FSCs of some bounded size.

\begin{definition}[Finite-State Controller]
A finite state controller for a BAMDP is defined as a graph, where nodes represent memory states and edges represent observations in the form of $(s,a,s')$ triplets.  Each node is also associated with an action (or alternately a distribution over actions), which represents the policy itself.
\end{definition}

In this representation, memory states correspond to finite trajectories, rather than full hyper-states. Tractability is achieved by limiting the number of memory states that are included in the graph.  Given a specific finite-state controller (i.e., a compact policy), its expected value can be calculated in closed form using a system of Bellman equations, where the number of equations/variables is equal to $|\Scal| \times |Q|$, where $|\Scal|$ is the number of states in the original MDP, and $|Q|$ is the number of memory states in the FSC~\citep{duff01bamdpfsc}. The remaining challenge is to optimize the policy. This is achieved using a gradient descent algorithm.  A Monte-Carlo gradient estimation is proposed to make it more tractable. This approach presupposes the existence of a good FSC representation for the policy. In general, while conceptually and mathematically straight-forward, this method is computationally feasible only for small domains with few memory states. For many real-world domains, the number of memory states needed to achieve good performance is far too large.


\subsection*{Bayesian Exploration Exploitation Tradeoff in LEarning (BEETLE)}

An alternate approximate offline algorithm to solve the BAMDP is called BEETLE~\citep{poupart06beetle}. This is an extension of the Perseus algorithm~\citep{Spaan05Perseus} that was originally designed for conventional POMDP planning, to the BAMDP model. Essentially, at the beginning, hyper-states $(s,\phi)$ are sampled from random interactions with the BAMDP model. An equivalent continuous POMDP (over the space of states and transition functions) is solved instead of the BAMDP (assuming $b=(s,\phi)$ is a belief state in that POMDP). The value function is represented by a set of \emph{$\alpha$-functions} over the continuous space of transition functions (see Eqs.~\ref{eq:pomdp-bellman}-\ref{eq:pomdp-v*max}). In the case of BEETLE, it is possible to maintain a separate set of $\alpha$-functions for each MDP state, denoted $\Gamma^s$. Each $\alpha$-function is constructed as a linear combination of basis functions
$\alpha(b_t)=\int_\theta \alpha(\theta)b(\theta) d\theta$. In practice, the sampled hyper-states can serve to select the set of basis functions $\theta$.
The set of  $\alpha$-functions can then be constructed incrementally by applying Bellman updates at the sampled hyper-states using standard point-based POMDP methods~\citep{Spaan05Perseus,Pineau03PBVI}.

Thus, the constructed $\alpha$-functions can be shown to be multi-variate polynomials. The main computational challenge is that the number of terms in the polynomials increases exponentially with the planning horizon.  This can be mitigated by projecting each $\alpha$-function into a polynomial with a smaller number of terms (using basis functions as mentioned above).  The method has been tested experimentally in some small simulation domains.  The key to applying it in larger domains is to leverage knowledge about the structure of the domain to limit the parameter inference to a few key parameters, or by using parameter tying (whereby a subset of parameters are constrained to have the same posterior).



\section{Online near-myopic value approximation}
\label{ss:online-alg}

We recall from \S\ref{ss:belief-mdp} that for a planning horizon of $t$ steps, an offline BAMDP planner will consider optimal planning at $|\Scal|^t$ states.  In practice, there may be many fewer states, in particular because some trajectories will not be observed.  Online planning approaches interleave planning and execution on a step-by-step basis, so that planning resources are focused on those states that have been observed during actual trajectories.  We now review a number of online algorithms developed for the BAMDP framework.


\subsection*{Bayesian dynamic programming}
\label{sss:bayesianDP}
A simple approach, closely related to Thompson sampling (\S\ref{ss:thompson}), was proposed by Sterns~\citet{strens00}.  The idea is to sample a model from the posterior distribution over parameters, solve this model using dynamic programming techniques, and use the solved model to select actions.  Models are re-sampled periodically (e.g.,~at the end of an episode or after a fixed number of steps). The approach is  simple to implement and does not rely on any heuristics.  Goal-directed exploration is achieved via sampling of the models.  Convergence to the optimal policy is achievable because the method samples models from the full posterior over parameter uncertainty~\citep{strens00}. Of course this can be very slow, but it is useful to remember that the dynamic programming steps can be computed via simulation over the sampled model, and do not require explicit samples from the system.  Convergence of the dynamic programming inner-loop is improved by keeping maximum likelihood estimates of the value function for each state-action pair.  In the bandit case (single-step planning horizon), this method is in fact equivalent to Thompson sampling.   Recent work has provided a theoretical characterization of this approach, offering the first Bayesian regret bound for this setting~\citep{osband13}.


\subsection*{Value of information heuristic}

The Bayesian dynamic programming approach does not explicitly take into account the posterior uncertainty when selecting actions, and thus, cannot explicitly select actions which only provide information about the model. In contrast, Dearden et al.~\citet{dearden99model} proposed to select actions by considering their expected value of information (in addition to their expected reward). Instead of solving the BAMDP directly via Eq.~\ref{eq:ValFuncBAMDP}, the Dirichlet distributions are used to compute a distribution over the state-action values $Q^*(s,a)$, in order to select the action that has the highest expected return and \emph{value of information}. The distribution over Q-values is estimated by sampling MDPs from the posterior Dirichlet distribution, and then solving each sampled MDP (as detailed in \S\ref{ss:mdp}) to obtain different sampled Q-values.
	
The value of information is used to estimate the expected improvement in policy following an exploration action.  Unlike the full Bayes-optimal approach, this is defined myopically, over a 1-step horizon.

\begin{definition}[Value of perfect information~\citep{dearden99model}]
Let $VPI(s,a)$ denote the expected value of perfect information for taking action $a$ in state $s$.  This can be estimated by
$$VPI(s,a)=\int_{-\infty}^\infty Gain_{s,a}(x) \Pr(q_{s,a}=x) dx,$$
where
$$Gain_{s,a}(Q^*_{s,a})=
\left\{
\begin{array}{l l}
\exptE[q_{s,a_2}]-q^*_{s,a} & \text{if}\;\; a=a_1\;\; \text{and}\;\; q^*_{s,a}<\exptE[q_{s,a_2}],\\
q^*_{s,a}-E[q_{s,a_1}] & \text{if}\;\; a \neq a_1\;\; \text{and}\;\; q^*_{s,a}>\exptE[q_{s,a_2}],\\
0 & \text{otherwise},\\
\end{array}
\right.$$
assuming $a_1$ and $a_2$ denote the actions with the best and second-best expected values respectively, and $q^*_{s,a}$ denotes a random variable representing a possible value of $Q^*(s,a)$ in some realizable MDP.
\end{definition}

The value of perfect information gives an upper bound on the 1-step expected value of exploring with action $a$.  To balance exploration and exploitation, it is necessary to also consider the reward of action $a$. Thus, under this approach, the goal is to select actions that maximize $\exptE[q_{s,a}]+VPI(s,a)$.

From a practical perspective, this method is attractive because it can be scaled easily by varying the number of samples. Re-sampling and importance sampling techniques can be used to update the estimated Q-value distribution as the Dirichlet posteriors are updated.  The main limitation is that the myopic value of information may provide only a very limited view of the potential information gain of certain actions.


\section{Online Tree Search Approximation}
\label{ss:online-alg}

To select actions using a more complete characterization of the model uncertainty, an alternative is to perform a forward search in the space of hyper-states.


\subsection*{Forward search}

\begin{figure}[th!]
\centering
\includegraphics[scale=0.5]{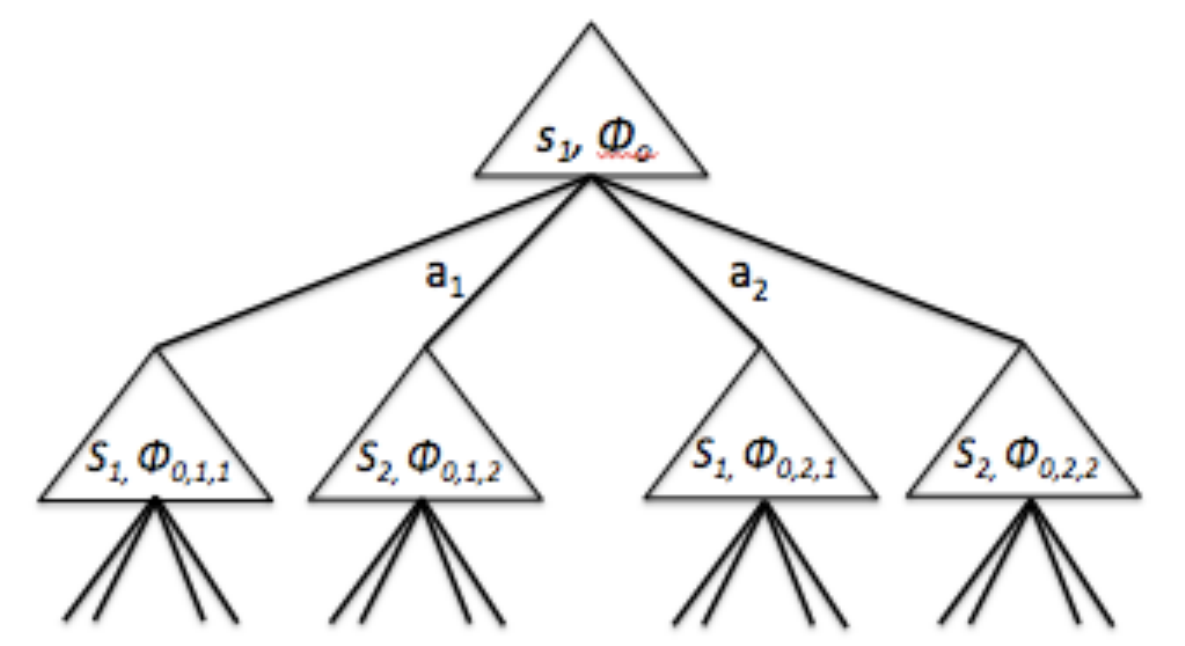}
\caption{An AND-OR tree constructed by the forward search process for the Chain problem. The top node contains the initial state $1$ and the prior over the model $\phi_0$. After the first action, the agent can end up in either state $1$ or state $2$ of the Chain and update its posterior accordingly. Note that depending on what action was chosen ($1$ or $2$) and the effect ($a$ or $b$), different parameters of $\phi$ are updated as per Algorithm~\ref{alOnline}}.
\label{fig:search-tree}
\end{figure}

An approach of this type was used in the case of a partially observable extension to the BAMDP~\citep{Ross08BAPOMDP}.  The idea here is to consider the current hyper-state and build a fixed-depth forward search tree containing all hyper-states reachable within some fixed finite planning horizon, denoted $d$.  Assuming some default value for the leaves of this tree, denoted $V_0(s,\phi)$, dynamic programming backups can be applied to estimate the expected return of the possible actions at the root hyper-state. The action with highest return over that finite horizon is executed and then the forward search is conducted again on the next hyper-state.  This approach is detailed in Algorithm \ref{alOnline} and illustrated in Figure~\ref{fig:search-tree}.   The main limitation of this approach is the fact that for most domains, a full forward search (i.e.,~without pruning of the search tree) can only be achieved over a very short decision horizon, since the number of nodes explored is $\mathcal{O}(|\Scal|^d)$, where $d$ is the search depth.
Another limitation is the need for a default value function to be used at the leaf nodes; this can either be a naive estimate, such as the immediate reward, $\max_a R(s,a)$, or a value computed from repeated Bellman backups, such as the one used for the Bayesian dynamic programming approach.  The next algorithm we review proposes some solutions to these problems.

We take this opportunity to draw the reader's attention to the survey paper on online POMDP planning algorithms~\citep{RossJAIR08}, which provides a comprehensive review and empirical evaluation of a range of search-based POMDP solving algorithms, including options for combining offline and online methods in the context of forward search trees.  Some of these methods may be much more efficient than those presented above and could be applied to plan in the BAMDP model.

\begin{algorithm}[h!]
 {\small
  \begin{algorithmic}[1]
	\STATE \textbf{function} $\textsc{ForwardSearch-BAMDP}(s,\phi,d)$%
	\IF{$d = 0$}
		\STATE \textbf{return} $V_0(s,\phi)$
	\ENDIF
    	\STATE $maxQ \leftarrow -\infty$
	\FORALL{$a \in \A$}
		\STATE $q \leftarrow  R(s,a)$
		\STATE $\phi'_{s,a} \leftarrow \phi_{s,a}$
		\FORALL{$s' \in \Scal$}
			\STATE $\phi'_{s,a,s'} \leftarrow \phi_{s,a,s'}+1$
			\STATE $q \leftarrow q + \gamma \frac{\phi_{s,a,s'}}{\sum_{s'' \in S}\phi_{s,a,s''}} \textsc{ForwardSearch-BAMDP}(s',\phi',d-1)$
			\STATE $\phi'_{s,a,s'} \leftarrow \phi_{s,a,s'}$
		\ENDFOR
		\IF{$q > maxQ$}
			\STATE $maxQ \leftarrow q$
		\ENDIF
	\ENDFOR
	\STATE \textbf{return} $maxQ$
  \end{algorithmic}}
\caption{Forward Search Planning in the BAMDP.}
\label{alOnline}
\end{algorithm}


\subsection*{Bayesian Sparse Sampling}

Wang et al.~\citet{wang05sparse} present an online planning algorithm that estimates the optimal value function of a BAMDP (Equation \ref{eq:ValFuncBAMDP}) using Monte-Carlo sampling. This algorithm is essentially an adaptation of the Sparse Sampling algorithm~\citep{kearns99sparse} to BAMDPs. The original Sparse Sampling approach is very simple:  a forward-search tree is grown from the current state to a fixed depth $d$.  At each internal node, a fixed number $C$ of next states are sampled from a simulator for each action in $\A$, to create $C|\A|$ children.  This is in contrast to the Forward Search approach of Algorithm~\ref{alOnline}, which considers all possible next states. Values at the leaves are estimated to be zero and the values at the internal nodes are estimated using the Bellman equation based on their children's values.  The main feature of Sparse Sampling is that it can be shown to achieve low error with high probability in a number of samples independent of the number of states~\citep{kearns99sparse}. A practical limitation of this approach is the exponential dependence on planning horizon.

To extend this to BAMDP, Bayesian Sparse Sampling introduces the following modifications. First, instead of growing the tree evenly by looking at all actions at each level of the tree, the tree is grown stochastically.  Actions are sampled according to their likelihood of being optimal, according to their Q-value distributions (as defined by the Dirichlet posteriors); next states are sampled according to the Dirichlet posterior on the model. This is related to Thompson sampling (\S\ref{ss:thompson}), in that actions are sampled according to their current posterior.  The difference is that the Bayesian Sampling explicitly considers the long-term return (as estimated from Monte-Carlo sampling of the posterior over model parameters), whereas Thompson sampling considers just the posterior over immediate rewards.  The specific algorithm reported in~\citet{wang05sparse} proposes a few improvements on this basic framework: such as, in some cases, sampling from the mean model (rather than the posterior) to reduce the variance of the Q-value estimates in the tree, and using myopic Thompson sampling (rather than the sequential estimate) to decide which branch to sample when growing the tree.  This approach requires repeatedly sampling from the posterior to find which action has the highest Q-value at each state node in the tree. This can be very time consuming, and thus, so far the approach has only been applied to small MDPs.


\subsection*{HMDP: A linear program approach for the hyper-state MDP}

Castro and Precup~\citet{Castro07LPBAMDP} present a similar approach to \citet{wang05sparse}. However, their approach differs on two main points. First, instead of maintaining only the posterior over models, they also maintain Q-value estimates at each state-action pair (for the original MDP states, not the hyper-states) using standard Q-learning~\citep{Sutton98IR}. Second, values of the hyper-states in the stochastic tree are estimated using \emph{linear programming}~\citep{Puterman94MD} instead of dynamic programming. The advantage of incorporating Q-learning is that it provides estimates for the fringe nodes which can be used as constraints in the LP solution.  There is currently no theoretical analysis to accompany this approach. However, the empirical tests on small simulation domains show somewhat better performance than the method of~\citet{wang05sparse}, but with similar scalability limitations.


\subsection*{Branch and bound search}

Branch and bound is a common method for pruning a search tree. The main idea is to maintain a lower bound on the value of each node in the tree (e.g.,~using partial expansion of the node to some fixed depth with a lower-bound at the leaf) and then prune nodes whenever they cannot improve the lower-bound (assuming we are maximizing values). In the context of BAMDPs, this can be used to prune hyper-state nodes in the forward search tree~\citep{Paquet05AAMAS}.  The challenge in this case is to find good bounds; this is especially difficult given the uncertainty over the underlying model. The method has been used in the context of partially observable BAMDP~\citep{png11,png2011thesis} using a  naive heuristic, $\sum_{d=0}^D \gamma^d R_{\max}$, where $D$ is the search depth and $R_{\max}$ is the maximum reward. The method was applied successfully to solve simulated dialogue management problems; computational scalability was achieved via a number of structural constraints, including the parameter tying method proposed by~\citet{poupart06beetle}.


\subsection*{BOP: Bayesian Optimistic Planning}

The BOP method~\citep{fonteneau13} is a variant on Branch and bound search in which nodes are expanded based on an upper-bound on their value.  The upper-bound at a root node, $s_0$, is calculated by building an optimistic subtree, $\mathcal{T}^+$, where leaves are expanded using: $x_t = \argmax_{x \in \mathcal{L}(\mathcal{T}^+)} P(x)\gamma^d$, where $\mathcal{L}(\mathcal{T}^+)$ denotes the fringe of the forward search tree, $P(x)$ denotes the probability of reaching node $x$ and $d$ denotes the depth of node $x$.

Theoretical analysis shows that the regret of BOP decreases exponentially with the planning budget.  Empirical results on the classic 5-state Chain domain (Figure~\ref{fig:chain}) confirm that the performance improves with the number of nodes explored in the forward search tree.

\subsection*{BAMCP: Bayes-adaptive Monte-Carlo planning}

In recent years, Monte-Carlo tree search methods have been applied with significant success to planning problems in games and other domains~\citep{gelly12}. The BAMCP approach exploits insights from this family of approaches to tackle the problem of joint learning and planning in the BAMDP~\citep{guez12}.

BAMCP provides a Bayesian adaptation of an algorithm, called \emph{Upper Confidence bounds applied to Trees} (UCT)~\citep{kocsis06}.  This approach is interesting in that it incorporates two different policies (UCT and rollout) to traverse and grow the forward search tree.  Starting at the root node, for any node that has been previously visited, it uses the UCT criteria
\begin{equation*}
a^*=\argmax_a Q(s,h,a) + c \sqrt{\frac{\log\big(n(s,h)\big)}{n(s,h,a)}}
\end{equation*}
to select actions. Along the way, it updates the statistics
$n(s,h)$ (the number of times the node corresponding to state $s$ and history $h$ has been visited in the search tree) and $n(s,h,a)$ (the number of times action $a$ was chosen in this node); these are initialized at $0$ when the tree is created. Once it reaches a leaf node, instead of using a default value function (or bound), it first selects any untried action (updating its count to $1$) and then continues to search forward using a \emph{rollout} policy until it reaches a given depth (or terminal node).  The nodes visited by the rollout policy are not added to the tree (i.e.,~no $n(\cdot)$ statistics are preserved).

To apply this to the Bayesian context, BAMCP must select actions according to the posterior of the model parameters. Rather than sampling multiple models from the posterior, BAMCP samples a single model $P_t$ at the root of the tree and uses this same model (without posterior updating) throughout the search tree to sample next states, after both UCT and rollout actions. In practice, to reduce computation in large domains, the model $P_t$ is not sampled in its entirety at the beginning of the tree building process, rather, it is generated lazily as samples are required.

In addition, to further improve efficiency, BAMCP uses learning within the forward rollouts to direct resources to important areas of the search space.  Rather than using a random policy for the rollouts, a model-free estimate of the value function is maintained
\begin{equation*}
\hat{Q}(s_t,a_t) \leftarrow \hat{Q}(s_t,a_t) + \alpha \big(r_t + \gamma \max_a \hat{Q}(s_{t+1},a) - \hat{Q}(s_t,a_t)\big),
\end{equation*}
and actions during rollouts are selected according to the $\epsilon$-greedy policy defined by this estimated $\hat{Q}$ function.

BAMCP is shown to converge (in probability) to the optimal \emph{Bayesian} policy (denoted $V^*(s,h)$ in general, or $V^*(s,\phi)$ when the posterior is constrained to a Dirichlet distribution).  The main complexity result for BAMCP is based on the UCT analysis and shows that the error in estimating $V(s_t,h_t)$ decreases as $\O \Big(\log\big(n(s_t,h_t)\big)/n(s_t,h_t) \Big)$.

Empirical evaluation of BAMCP with a number of simulation domains has shown that it outperforms Bayesian Dynamic Programming, the Value of Information heuristic, BFS3~\citep{asmuth11}, as well as BEB~\citep{kolter09} and SmartSampler~\citep{castro10}, both of which will be described in the next section.  A good part of this success could be attributed to the fact that unlike many forward search sparse sampling algorithm (e.g.,~BFS3), BAMCP can take advantage of its learning during rollouts to effectively bias the search tree towards good solutions.


\section{Methods with Exploration Bonus to Achieve PAC Guarantees}
\label{ss:sampling}

We now review a class of algorithms for the BAMDP model that are guaranteed to select actions such as to incur only a small loss compared to the optimal Bayesian policy.  Algorithmically, these approaches are similar to those examined in \S\ref{ss:online-alg} and typically require forward sampling of the model and decision space.  Analytically, these approaches have been shown to achieve bounded error in a polynomial number of steps using analysis techniques from the \emph{Probably Approximately Correct} (PAC) literature.  These methods are rooted in earlier papers showing that reinforcement learning in finite MDPs can be achieved in a polynomial number of steps~\citep{kearns98,brafman03,littman05}. These earlier papers did not assume a Bayesian learning framework; the extension to Bayesian learning was first established in the BOSS (Best of Sampled Sets) approach.

The main idea behind many of the methods presented in this section is the notion of \emph{Optimism in the Face of Uncertainty}, which suggests that, when in doubt, an agent should act according to an optimistic model of the MDP; in the case where the optimistic model is correct, the agent will gather reward, and if not, the agent will receive valuable information from which it can infer a better model.


\subsection*{BFS3: Bayesian Forward Search Sparse Sampling}

The BFS3 approach~\citep{asmuth11} is an extension to the Bayesian context of the Forward Search Sparse Sampling (FSSS) approach~\citep{walsh10}.  The Forward Search Sparse Sampling itself extends the Sparse Sampling approach~\citep{kearns99sparse} described above in a few directions.  In particular, it maintains both lower and upper bounds on the value of each state-action pair, and uses this information to direct forward rollouts in the search tree. Consider a node $s$ in the tree, then the next action is chosen greedily with respect to the upper-bound $U(s,a)$. The next state $s'$ is selected to be the one with the largest difference between its lower and upper bound (weighted by the number of times it was visited, denoted by $n(s,a,s')$), i.e.,~$s' \leftarrow \argmax_{s' \sim P(\cdot | s,a)}\big(U(s')-L(s')\big)n(s,a,s')$.  This is repeated until the search tree reaches the desired depth.

The BFS3 approach takes this one step further by building the sparse search tree over hyper-states $(s,\phi)$, instead of over simple states $s$. As with most of the other approaches presented in this section, the framework can handle model uncertainty over either the transition parameters and/or the reward parameters.

Under certain conditions, one can show that BFS3 chooses at most a polynomial number of sub-optimal actions compared to an policy.

\begin{theorem}~\citep{asmuth13}\label{thm:bfs3}\footnote{We have slightly modified the description of this theorem, and others below, to improve legibility of the paper. Refer to the original publication for full details.}
With probability at least $1-\delta$, the expected number of sub-$\epsilon$-Bayes-optimal actions taken by BFS3 is at most $BSA(S+1)d/\delta_t$ under assumptions on the accuracy of the prior and optimism of the underlying FSSS procedure.
\end{theorem}

Empirical results show that the method can be used to solve small domains (e.g.,~5x5 grid) somewhat more effectively than a non-Bayesian method such as RMAX~\citep{brafman03}. Results also show that BFS3 can take advantage of structural assumptions in the prior (e.g.,~parameter tying) to tackle much larger domains, up to $10^{16}$ states.


\subsection*{BOSS: Best of Sample Sets}

As per other model-based Bayesian RL approaches presented above, BOSS~\citep{asmuth09} maintains uncertainty over the model parameters using a parametric posterior distribution and incorporates new information by updating the posterior. In discrete domains, this posterior is also typically represented using Dirichlet distributions.

When an action is required, BOSS draws a set of sample models $\M_i, i=1,\ldots,K$, from the posterior $\phi_t$. It then creates a hyper-model $\M\#$, which has the same state space $\Scal$ as the original MDP, but has $K|\A|$ actions, where $a_{ij}$ corresponds to taking action $a_j \in \A$ in model $\M_i$. The transition function for action $a_{i \cdot}$ is constructed directly by taking the sampled transitions $P_i$ from $\M_i$.  It then solves the hyper-model $\M\#$ (e.g.,~using dynamic programming methods) and selects the best action according to this hyper-model.  This sampling procedure is repeated a number of times (as determined by $B$, the \emph{knownness} parameter) for each state-action pair, after which the policy of the last $\M\#$ is retained.

While the algorithm is simple, the more interesting contribution is in the analysis. The main theoretical result shows that assuming a certain knownness parameter $B$, the value at state $s$ when visited by algorithm $\mathfrak{A}$ at time $t$ (which we denote $V^{\mathfrak{A}_t}(s_t)$), is very close to the optimal value of the correct model (denoted $V^*$) with high probability in all but a small number of steps. For the specific case of Dirichlet priors on the model, it can be shown that the number of necessary samples, $f$, is a polynomial function of the relevant parameters.
	
\begin{theorem}~\citep{asmuth09}\label{thm:boss}
When the knownness parameter $B=\max_{s,a} f\big(s,a,\epsilon(1-\gamma)^2,\frac{\delta}{|\Scal||\A|},\frac{\delta}{|\Scal|^2|\A|^2K}\big)$, then with probability at least $1-4\delta$, $V^{\mathfrak{A}_t}(s_t) \geq V^*(s_t)-4\epsilon$ in all but $\xi(\epsilon,\delta)=\O\left(\frac{|\Scal||\A| B}{\epsilon(1-\gamma)^2}\ln \frac{1}{\delta}\ln\frac{1}{\epsilon(1-\gamma)}\right)$ steps.
\end{theorem}

Experimental results show that empirically, BOSS performs similarly to BEETLE and Bayesian dynamic programming in simple problems, such as the Chain problem (Figure~\ref{fig:chain}).  BOSS can also be extended to take advantage of structural constraints on the state relations to tackle larger problems, up to a $6\times 6$ maze.

Castro and Precup~\citet{castro10} proposed an improvement on BOSS, called {\sc SmartSampler}, which addresses the problem of how many models to sample.  The main insight is that the number of sampled models should depend on the variance of the posterior distribution.  When the variance is larger, more models are necessary to achieve good value function estimation.  When the variance is reduced, it is sufficient to sample a smaller number of models. Empirical results show that this leads to reduced computation time and increased accumulated reward, compared to the original BOSS.


\subsection*{BEB: Bayesian Exploration Bonus}

Bayesian Exploration Bonus (BEB) is another simple approach for achieving Bayesian reinforcement learning with guaranteed bounded error within a small number of samples~\citep{kolter09}.  The algorithm simply chooses actions greedily with respect to the following value function:
$$\tilde{V}^*_t(s,\phi) = \max_{a \in \A} \left\{  R(s,a) \!+\! \frac{\beta}{1+\sum_{s' \in \Scal} \phi_{s,a,s'}} \!+\! \sum_{s' \in \Scal}  P(s' | s,\phi,a) \tilde{V}^*_{t-t}(s',\phi) \right\} $$

Here the middle term on the right-hand side acts as exploration bonus, whose magnitude is controlled by parameter $\beta$.  It is worth noting that the posterior (denoted by $\phi$) is not updated in this equation, and thus, the value function can be estimated via standard dynamic programming over the state space (similar to BOSS and Bayesian dynamic programming).

The exploration bonus in BEB decays with $1/n$ (consider $n \sim \sum_{s' \in \Scal} \phi_{s,a,s'}$), which is significantly faster than similar exploration bonuses in the PAC-MDP literature, for example the MBIE-EB algorithm~\citep{strehl08}, which typically declines with $1/\sqrt{n}$.  It is possible to use this faster decay because BEB aims to match the performance of the optimal \emph{Bayesian} solution (denoted $V^*(s,\phi)$), as defined in Theorem~\ref{thm:beb-1}. We call this property PAC-BAMDP. This is in contrast to BOSS where the analysis is with respect to the optimal solution of the correct model (denoted $V^*(s)$).

\begin{theorem}~\citep{kolter09}\label{thm:beb-1}
Let $\mathcal{A}_t$ denote the policy followed by the BEB algorithm (with $\beta=2H^2$) at time $t$, and let $s_t$ and $\phi_t$ be the corresponding state and posterior belief. Also suppose we stop updating the posterior for a state-action pair when $\sum_{s' \in \Scal} \phi_{s,a,s'} \geq 4H^3/\epsilon$. Then with probability at least $1-\delta$, for a planning horizon of $H$, we have
$$V^{\mathcal{A}_t}(s_t,\phi_t) \geq V^*(s_t,\phi_t) - \epsilon$$
for all but $m$ time steps, where
$$m=\O \left( \frac{|\Scal||\A|H^6}{\epsilon^2}\log\frac{|\Scal||\A|}{\delta} \right).$$
\end{theorem}

It is worth emphasizing that because BEB considers optimality with respect to the value of the Bayesian optimal policy, it is possible to obtain theoretical results that are tighter (i.e.,~fewer number of steps) with respect to the optimal value function.  But this comes at a cost, and in particular, there are some MDPs for which the BEB algorithm, though it performs closely to the Bayesian optimum, $V^*(s,\phi)$, may never find the actual optimal policy, $V^*(s)$. This is illustrated by an example in the dissertation of Li~\citet{li09} (see Example 9, p.80), and formalized in Theorem~\ref{thm:beb-2} (here $n(s,a)$ denotes the number of times the state-action pair $s,a$ has been observed).

\begin{theorem} \citep{kolter09}\label{thm:beb-2}
Let $\mathcal{A}_t$ denote the policy followed by an algorithm using any (arbitrary complex) exploration bonus that is upper bounded by $\frac{\beta}{n(s,a)^p}$, for some constants $\beta$ and $p > 1/2$. Then there exists some MDP, $\M$, and parameter, $\epsilon_0(\beta,p)$, such that with probability greater than $\delta_0=0.15$,
$$V^{\mathcal{A}_t}(s_t) < V^*(s_t) - \epsilon_0,$$
will hold for an unbounded number of time steps.
\end{theorem}

Empirical evaluation of BEB showed that in the Chain problem (Figure~\ref{fig:chain}), it could outperform a (non-Bayesian) PAC-MDP algorithm in terms of sample efficiency and find the correct optimal policy.


\subsection*{VBRB:  Variance-Based Reward Bonus}

The \emph{Variance-based reward} approach also tackles the problem of Bayesian RL by applying an exploration bonus. However in this case, the exploration bonus depends on the variance of the model parameters with respect to the posterior~\citep{sorg10}, i.e.,

$$\tilde{V}^*_t(s,\phi) = \max_{a \in \A} \left\{  R(s,\phi,a) + \hat{R}_{s,\phi,a}+ \sum_{s' \in \Scal}  P(s' | s,\phi,a) \tilde{V}^*_{t-t}(s',\phi) \right\},$$
where
the reward bonus $\hat{R}_{s,\phi,a}$ is defined as
$$\beta_R\;\sigma_{R(s,\phi,a)} + \beta_P\;\sqrt{\sum_{s' \in \Scal} \sigma^2_{P(s'|s,\phi,a)}}\;,$$
with
\begin{eqnarray}
\sigma^2_{R(s,\phi,a)} &= & \int_\theta R(s,\theta,a)^2b(\theta)d\theta - R(s,\phi,a)^2,\\
\sigma^2_{P(s'|s,\phi,a)} &= & \int_\theta P(s'|s,\theta,a)^2b(\theta)d\theta - P(s'|s,\phi,a)^2,
\label{eqn:sorg-bonus}
\end{eqnarray}
and $\beta_R$ and $\beta_P$ are constants controlling the magnitude of the exploration bonus.\footnote{The approach is presented here in the context of Dirichlet distributions. There are ways to generalize this use of variance of the reward as an exploration bonus for other Bayesian priors~\citep{sorg10}.}

The main motivation for considering a variance-based bonus is that the error of the algorithm can then be analyzed by drawing on Chebyshev's inequality (which states that with high probability, the deviation of a random variable from its mean is bounded by a multiple of its variance).  The main theoretical result concerning the variance-based reward approach bounds the sample complexity with respect to the optimal policy of the true underlying MDP, like BOSS (and unlike BEB).

\begin{theorem} \citep{sorg10}\label{thm:sorg}
Let the sample complexity of state $s$ and action $a$ be $C(s,a)=f\big(b_0,s,a,\frac{1}{4}\epsilon(1-\gamma)^2,\frac{\delta}{|\Scal||\A|},\frac{\delta}{2|\Scal|^2|\A|^2}\big)$. Let the internal reward be defined as $\hat{R}(s,\phi,a)=\frac{1}{\sqrt{\rho}}\left(\sigma_{R(s,\phi,a)}+\frac{\gamma|\Scal|}{1-\gamma}\sqrt{\sum_{s'}\sigma^2_{P(s'|s,\phi,a)}} \right)$ with $\rho=\frac{\delta}{2|\Scal|^2|\A|^2}$. Let $\theta^*$ be the random true model parameters distributed according to the prior belief $\phi_0$. The variance-based reward algorithm will follow a $4\epsilon$-optimal policy from its current state, with respect to the MDP $\theta^*$, on all but $\O\left( \frac{\sum_{s,a}C(s,a)}{\epsilon(1-\gamma)^2}\ln\frac{1}{\delta}\ln\frac{1}{\epsilon(1-\gamma)}\right)$ time steps with probability at least $1-4\delta$.
\end{theorem}

For the case where uncertainty over the transition model is modelled using independent Dirichlet priors (as we have considered throughout this section), the sample complexity of this approach decreases as a function of $\O(1/\sqrt{n})$.  This is consistent with other PAC-MDP results, which also bound the sample complexity to achieve small error with respect to the optimal policy of the true underlying MDP. However, it is not as fast as the BEB approach, which bounds error with respect to the best Bayesian policy.

Empirical results for the variance-based approach show that it is competitive with BEETLE, BOSS and BEB on versions of the Chain problem that use parameter tying to reduce the space of model uncertainty, and shows better performance for the variant of the domain where uncertainty over all state-action pairs is modelled with an independent Dirichlet distribution.  Results also show superior performance on a $4\times 4$ grid-world inspired by the Wumpus domain of~\citet{russell02}.



\subsection*{BOLT: Bayesian Optimistic Local Transitions}

Both BEB and the variance-based reward approach encourage exploration by putting an exploration bonus on the reward, and solving this modified MDP. The main insight in BOLT is to put a similar exploration bonus on the transition probabilities~\citep{araya12}.  The algorithm is simple and modelled on BOSS. An extended action space is considered: $\A'=\A \times S$, where each action in $\zeta=(a,\sigma) \in \A'$ has transition probability $\hat{P}(s'|s,\zeta)=\exptE\big[P(s'|s,a)|s,\phi,\lambda^\eta_{s,a,\sigma}\big]$, with $\phi$ being the posterior on the model parameters and $\lambda^\eta$ being a set of $\eta$ \emph{imagined} transitions $\lambda^\eta_{s,a,\sigma}=\{(s,a,\sigma), \ldots, (s,a,\sigma)\}$. The extended MDP is solved using standard dynamic programming methods over a horizon $H$, where $H$ is a parameter of BOLT, not the horizon of the original problem.  The effect of this exploration bonus is to consider an action $\zeta$ that takes the agent to state $\sigma \in \Scal$ with greater probability than has been observed in the data, and by optimizing a policy over all such possible actions, we consider an optimistic evaluation over the possible set of models.  The parameter $\eta$ controls the extent of the optimistic evaluation, as well as its computational cost (larger $\eta$ means more actions to consider).   When the posterior is represented using standard Dirichlet posteriors over model parameters, as considered throughout this section, it can be shown that BOLT is always optimistic with respect to the optimal Bayesian policy~\citep{araya12}.

\begin{theorem} \citep{araya12}\label{thm:bolt}
Let $\mathcal{A}_t$ denote the policy followed by BOLT at time $t$ with $\eta=H$. Let also $s_t$ and $\phi_t$ be the corresponding state and posterior at the time. Then, with probability at least $1-\delta$, BOLT is $\epsilon$-close to the optimal Bayesian policy $$V^{\mathcal{A}_t}(s_t,\phi_t) < V^*(s_t,\phi_t) - \epsilon,$$
for all but $\widetilde{\O}(\frac{|\Scal||\A|\eta^2}{\epsilon^2(1-\gamma)^2})=\widetilde{\O}(\frac{|\Scal||\A|H^2}{\epsilon^2(1-\gamma)^2})$ time steps.
\end{theorem}
To simplify, the parameter $H$ can be shown to depend on the desired correctness ($\epsilon,\delta$).

Empirical evaluation of BOLT on the Chain problem shows that while it may be outperformed by BEB with well-tuned parameters, it is more robust to the choice of parameter ($H$ for BOLT, $\beta$ for BEB).  The authors suggest that BOLT is more stable with respect to the choice of parameter because optimism in the transitions is bounded by laws of probability (i.e.,~even with a large exploration bonus, $\eta$, the effect of extended actions $\zeta$ will simply saturate), which is not the case when the exploration bonus is placed on the rewards.

BOLT was further extended, in the form of an algorithm called POT:  Probably Optimistic Transition~\citep{kawaguchi13}, where the parameter controlling the extended bonus, $\eta$, is no longer constant.  Instead, the transition bonus is estimated online for each state-action pair using a UCB-like criteria that incorporates the notion of the current estimate of the transition and the variance of that estimate (controlled by a new parameter $\beta$).  One advantage of the method is that empirical results are improved (compared to BOLT, BEB, and VBRB) for domains where the prior is good, though empirical performance can suffer  when the prior is misspecified.  POT is also shown to have tighter bounds on the sample complexity. However, the analysis is done with respect to a modified optimality criteria, called ``Probably Upper Bounded Belief-based Bayesian Planning", which is weaker than the standard Bayesian optimality, as defined in \S\ref{ss:explore-exploit}.

Table~\ref{table:onlineBRL} summarizes the key features of the online Bayesian RL methods presented in this section.  This can be used as a quick reference to visualize similarities and differences between the large number of related approaches.  It could also be used to devise new approaches, by exploring novel combinations of the existing components.

\vspace{0.15in}
\begin{table}
\centering
\begin{footnotesize}
\rotatebox{90}{
\begin{tabular}{|l|c|c|c|c|c|c|}
\hline
Algorithm & Depth & Actions selected & Next-states & Posterior & Solution & Theoretical\\
& & in search & considered & sampling & method & properties\\
\hline
&&&&&&\\
Bayesian DP & N/A & $\forall a \in \A$ & $\forall s' \in S$ & Sample 1 &  $Q()$ solved by DP & MDP opt. $\rightarrow \infty$ \\
&&&&per step && Bounded\\
&&&&&& expected \\
&&&&&& regret\\
&&&&&& \\
Value of & N/A & Information gain &  $\forall s' \in S$ & Sample $K$& $Q()$ solved by DP & Not known\\
information &&&& per step &&\\
&&&&&&\\
&&&&&&\\
Forward& fixed $d$ & $\forall a \in \A$ & $\forall s' \in S$ & Re-sample& Backups in tree, & Bayes-opt. $\rightarrow \infty$\\
search &&&& at node & heuristic at leave&\\
&&&&&&\\
Bayesian& variable $d$ &  $a \sim E[\hat{R}(s,a)]$ & $\!\!s'\!\! \sim \hat{Pr}(s'|s,a,\phi)$  & Re-sample  & Backups in tree, & Not known\\
Sparse&&&& at node & heuristic at leaves&\\
Sampling&&&&&&\\
&&&&&&\\
HMDP & fixed $d$ & $a \sim rand()$ & $\!\!s'\!\! \sim \hat{Pr}(s'|s,a,\phi)$ & Re-sample & Backups in tree, & Not known\\
&&&& at node & $Q()$ by LP at leaves &\\
&&&&&&\\
Branch and & variable $d$ & $\forall a \in \A$, except if & $\forall s' \in S$ & Re-sample & Backups in tree, & Bayes-opt. $\rightarrow \infty$\\
Bound & & $\exists a' U(s,\phi,a)<$ & & at node & heuristic at leaves & \\
&&$L(s,\phi,a')$&&&&\\
&&&&&&\\
BOP & variable $d$ & $\argmax_a B(x,a)$ & $\forall s' \in S$ & Re-sample & Backups in tree, & Bayes-opt. $\rightarrow \infty$\\
& &  & &  at node & $\frac{1}{1-\gamma}$ at leaves & \\
&&&&&&\\
BAMCP & fixed $d$ & UCT criteria & $\!\!s'\!\! \sim \hat{Pr}(s'|s,a,\phi)$ & Sample 1 & Rollouts with & Bayes-opt. $\rightarrow \infty$\\
&&&& per step & fixed policy $\hat{\mu}^*$  &\\
&&&&&&\\
\hline
\end{tabular}
}
\caption{Online Model-based Bayesian RL methods (DP=Dynamic programming, LP = Linear programming, U=Upper-bound, L=Lower-bound)}\label{table:onlineBRL}
\end{footnotesize}
\end{table}
\addtocounter{table}{-1}
\begin{table}
\centering
\begin{footnotesize}
\rotatebox{90}{
\begin{tabular}{|l|c|c|c|c|c|c|}
\hline
Algorithm & Depth & Actions selected & Next-states & Posterior & Solution & Theoretical\\
& & in search & considered & sampling & method & properties\\
\hline
&&&&&&\\
BFS3 & variable $d$ & $\argmax_a$ & $\argmax_{s'}$ & Re-sample& Backups in tree, & PAC guarantee\\
&&$U(s,\phi,a)$&$(U(s')-L(s'))$ & at node & heuristic at leaves &\\
&&&&&&\\
BOSS & N/A & $\A \times K$ & $\forall s' \in S$ & Sample $K$ & Q() solved by DP & PAC-MDP\\
&&&&per step &&\\
&&&&&&\\
Smart- & N/A & $\A \times K$ & $\forall s' \in S$ & Sample $K$& Q() solved by DP & PAC-MDP\\
Sampler &&&&  per step, &&\\
&&&&$K=\max_{s' }$ &&\\
&&&&$\frac{var(P(s,a,s'))}{\epsilon}$ &&\\
&&&&&&\\
BEB & N/A & $\forall a \in \A$ & $\forall s' \in S$ & N/A & $Q()$ + reward bonus & PAC-BAMDP\\
& & & & & solved by DP & \\
&&&&&&\\
VBRB & N/A & $\forall a \in \A$ & $\forall s' \in S$ & N/A & $Q()$ + variance bonus & PAC-MDP\\
rewards& & & & & solved by DP & \\
&&&&&&\\
BOLT & N/A & $\A \times \Scal$ & $\forall s' \in S$ & Create $\eta$ & $Q()$ solved by DP & PAC-BAMDP\\
&&&&per step &&\\
\hline
\end{tabular}
}
\caption{(cont'd) Online Model-based Bayesian RL methods (DP=Dynamic programming, LP = Linear programming, U=Upper-bound, L=Lower-bound)}\label{table:onlineBRL}
\end{footnotesize}
\end{table}


\section{Extensions to Unknown Rewards}
\label{ss:ext-rewards}

Most work on model-based Bayesian RL focuses on unknown transition functions and assumes the reward function is known. This is seen as the more interesting case because there are many domains in which dynamics are difficult to model \emph{a priori}, whereas the reward function is set by domain knowledge.  Nonetheless a number of papers explicitly consider the case where the reward function is uncertain~\citep{dearden99model,strens00,wang05sparse,Castro07LPBAMDP,sorg10}.  The BAMDP model extends readily to this case, by extending the set of hyper-states to full set of unknown parameters: $\Scal' \in \Scal \times \Theta$, with $\theta=(\phi,\vartheta)$, with $\vartheta$ representing the prior over the unknown reward function.  The next challenge is to choose an appropriate distribution for this prior.

The simplest model is to assume that each state-action pair generates a binary reward, $\vartheta=\{0,1\}$, according to a Bernoulli distribution~\citep{wang05sparse,Castro07LPBAMDP}.  In this case, a conjugate prior over the reward can be captured with a Beta distribution (see \S\ref{sss:Dir}).  This is by far the most common approach when dealing with bandit problems.  In the cases where the reward function is drawn from a set of discrete values, $\vartheta=\{r_1, r_2,\dots, r_k\}$, it can be modeled by a Dirichlet prior over those values.  However, many MDP domains are characterized by more complex (e.g.,~real-valued) reward functions.

In those cases, an alternative is to assume that the reward is Gaussian distributed, i.e.,~$\vartheta=\{\mu,\sigma\}$.  In this case the choice of prior on the standard deviation $\sigma$ is of particular importance; a uniform prior could lead the posterior to converge to $\sigma \rightarrow 0$. Following Sterns~\citet{strens00}, we can define the precision $\psi=1/\sigma$ with prior density $f(\psi) \propto \psi\exp(-\psi^2\sigma_0^2/2)$; this will have a maximum at $\sigma=\sigma_0$.  We can also define the prior on the mean to be $f(\mu) \propto \N(\mu_0,\sigma^2)$. The parameters $(\mu_0,\sigma_0)$ capture any prior knowledge about the mean and standard deviation of the reward function.  After $n$ observations of the reward with sample mean $\hat{\mu}$ and sample variance $\hat{\sigma}$, the posterior distribution on $\psi$ is defined by: $f(\psi) \propto \psi^{n-1}\exp\big(-\psi^2(n\hat{\sigma}+\sigma_0^2)/2\big)$, and the posterior on $\mu$ is defined by $f(\mu) \propto \N(\hat{\mu},\sigma^2/n)$, where $\sigma=1/\psi$. Note that the posterior on the variance is updated first and used to calculate the posterior on the mean.

Most of the online model-based BRL algorithms presented in Table~\ref{table:onlineBRL} extend readily to the case of unknown rewards, with the added requirement of sampling the posterior over rewards (along with the posterior over transition functions).   For an algorithm such as BEB, that uses an exploration bonus in the value function, it is also necessary to incorporate the uncertainty over rewards within that exploration bonus. Sorg et al,~\citet{sorg10} deals with the issue by expressing the bonus over the variance of the unknown parameters (including the unknown rewards), as in Eq.~\ref{eqn:sorg-bonus}.


\section{Extensions to Continuous MDPs}
\label{ss:ext-cmdp}

Most of the work reviewed in this section focuses on discrete domains. In contrast, many of the model-free BRL methods concern the case of continuous domains, where the Bayesian approach is used to infer a posterior distribution over value functions, conditioned on the state-action-reward trajectory observed in the past.

The problem of optimal control under uncertain model parameters for continuous systems was originally introduced by~\cite{feldbaum61}, as the theory of dual control (also referred to as adaptive control or adaptive dual control).  Several authors studied the problem for different classes of time-varying systems~\citep{filatov00,wittenmark95}: linear time invariant systems under partial observability~\citep{rusnak95}, linear time varying Gaussian models with partial observability~\citep{ravikanth92}, nonlinear systems with full observability~\citep{zane92}, and nonlinear systems with partial observability~\citep{greenfield03,ross08icra}.

This last case is closest mathematically to the Bayes-Adaptive MDP model considered throughout this paper.  The dynamical system is described by a Gaussian transition model (not necessarily linear):
$$s_{t} = g_T(s_{t-1},a_{t-1},V_t),$$
where $g_T$ is a specified function, and $V_t \sim \N(\mu_v,\Sigma_v)$ is an unknown $k$-variate normal distribution with mean vector $\mu_v$ and covariance matrix $\Sigma_v$.  Here the prior distribution on $V_t$ is represented using a Normal-Wishart distribution~\citep{ross08icra}.
Particle filters using Monte-Carlo sampling methods are used to track the posterior over the parameters of this distribution.  Planning is achieved using a forward search algorithm similar to Algorithm~\ref{alOnline}.

The Bayesian DP strategy (\S\ref{sss:bayesianDP}) and its analysis were also extended to the continuous state and action space and average-cost setting, under the assumption of smoothly parameterizable dynamics~\citep{abbasi15}.   The main algorithmic modification in this extension is to incorporate a specific schedule for updating the policy, that is based on a measure of uncertainty.

Another approach proposed by~\cite{dallaire09} allows flexibility over the choice of transition function.  Here the transition and reward functions are defined by:
\begin{eqnarray}
s_t=g_T(s_{t-1},a_{t-1})+\epsilon_T,\\
r_t=g_R(s_t,a_t) + \epsilon_R,
\end{eqnarray}
where $g_T$ and $g_R$ are modelled by (independent) Gaussian Processes (as defined in \S\ref{sss:GP}) and $\epsilon_T$ and $\epsilon_R$ are zero-mean Gaussian noise terms.  Belief tracking is done by updating the Gaussian process.  Planning is achieved by a forward search tree similar to Algorithm~\ref{alOnline}.

It is also worth pointing out that the Bayesian Sparse Sampling approach~\citep{wang05sparse} has also been extended to the case where the reward function is expressed over a continuous action space and represented by a Gaussian process.  In this particular case, the authors only considered Bayesian inference of the reward function for a single state and a single-step horizon problem (i.e.,~a bandit with continuous actions).  Under these conditions, inferring the reward function is the same as inferring the value function, so no planning is required.


\section{Extensions to Partially Observable MDPs}
\label{ss:ext-pomdp}

We can extend the BAMDP model to capture uncertainty over the parameters of a POMDP (as defined in \S\ref{ss:pomdp}) by introducing the Bayes-Adaptive POMDP (BAPOMDP) model~\citep{Ross08BAPOMDP,ross11}.  Given a POMDP model, $\langle\Scal,\A,\O,P,\Omega,P_0,R\rangle$, we define a corresponding BAPOMDP as follows:

\begin{model}[(Bayes-Adaptive POMDP)]
Define a BAPOMDP $\M$ to be a tuple $\langle\Scal',\A,P',P_0',R'\rangle$ where\\
$\bullet$ $\Scal'$ is the set of hyper-states, $\Scal \times \Phi \times \Psi$,\\
$\bullet$ $\A$ is the set of actions,\\
$\bullet$ $P'(\cdot|s,\phi,\psi,a)$ is the transition function between hyper-states, conditioned on action $a$ being taken in hyper-state $(s,\phi,\psi)$,\\
$\bullet$ $\Omega'(\cdot|s,\phi,\psi,a)$ is the observation function, conditioned on action $a$ being taken in hyper-state $(s,\phi,\psi)$,\\
$\bullet$ $P'_0\in\P(\Scal \times \Phi \times \Psi)$ combines the initial distribution over physical states, with the prior over transition functions $\phi_0$ and the prior over observation functions $\psi_0$,\\
$\bullet$ $R'(s,\phi,\psi,a)=R(s,a)$ represents the reward obtained when action $a$ is taken in state $s$.
\end{model}

Here $\Phi$ and $\Psi$ capture the space of Dirichlet parameters for the conjugate prior over the transition and observation functions, respectively.  The Bayesian approach to learning the transition $P$ and observation $\Omega$ involves starting with a prior distribution, which we denote $\phi_0$ and $\psi_0$, and maintaining the posterior distribution over $\phi$ and $\psi$ after observing the history of action-observation-rewards.  The belief can be tracked using a Bayesian filter, with appropriate handing of the observations.  Using standard laws of probability and independence assumptions, we have
\begin{equation}
\begin{split}
&\P(s',\phi',\psi',z|s,\phi,\psi,a) = \\ &\P(s'|s,a,\phi)\P(o|a,s',\psi)\P(\phi'|\phi,s,a,s')\P(\psi'|\psi,a,s',o).
\end{split}
\end{equation}

As in the BAMDP case (Eq.~\ref{eq:bamdp-belief}), $\P(s'|s,a,\phi)$ corresponds to the expected transition model under the Dirichlet posterior defined by $\phi$, and $\P(\phi'|\phi,s,a,s')$ is either 0 or 1, depending on whether $\phi'$ corresponds to the posterior after observing transition $(s,a,s')$ from prior $\phi$.  Hence $\P(s'|s,a,\phi)=\frac{\phi_{s,a,s'}}{\sum_{s'' \in \Scal}\phi_{s,a,s''}}$ and $\P(\phi'|\phi,s,a,s')=\mathbb{I}(\phi'_{s,a,s'}=\phi_{s,a,s'}+1)$.  Similarly, $\P(o|a,x',\psi)=\frac{\psi_{s',a,o}}{\sum_{o' \in \O}\psi_{s',a,o}}$ and $\P(\psi'|\psi,s',a,o)=I(\psi'_{s',a,o}=\psi_{s',a,o}+1)$.

Also as in the BAMDP, the number of possible hyper-states, $(s,\phi,\psi)$, grows exponentially (by a factor of $|\Scal|$) with the prediction horizon.  What is particular to the BAPOMDP is that the number of possible beliefs for a \emph{given trajectory}, $(a_0, o_1,\ldots, a_t, o_{t+1})$, also grows exponentially.  In contrast, given an observed trajectory, the belief in the BAMDP corresponds to a single hyper-state.  In the BAPOMDP, value-based planning requires estimating the Bellman equation  over all possible hyper-states for every belief:

\vspace{-0.125in}
\begin{equation} \label{eq:ValFuncBAPOMDP}
\begin{split}
V^*\big(b_t(s,\phi,\psi)\big) =& \max_a\Big\{ \sum_{s,\phi,\psi} R'(s,\phi,\psi,a)b_t(s,\phi,\psi) \\
&+ \gamma \sum_{o \in \O} \P(o|b_{t-1},a) V^*\big(\tau(b_t,a,o)\big) \Big\}.
\end{split}
\end{equation}
\vspace{-0.1in}

This is intractable for most problems, due to the large number of possible beliefs.  It has been shown that Eq.~\ref{eq:ValFuncBAPOMDP} can be approximated with an $\epsilon$-optimal value function defined over a smaller finite-dimensional space~\citep{ross11}. In particular, there exists a point where if we simply stop incrementing the counts ($\phi, \psi$), the value function of that approximate BAPOMDP (where the counts are bounded) approximates the BAPOMDP within some $\epsilon > 0$. In practice, that space is still very large.

The Minerva approach~\citep{jaulmes05} overcomes this problem by assuming that there is an oracle who is willing to provide full state identification at any time step. This oracle can be used to deterministically update the counts $\psi$ and $\phi$ (rather than keeping a probability distribution over them, as required when the state is not fully observable).  It is then possible to sample a fixed number of models from the Dirichlet posterior, solve these models using standard POMDP planning techniques, and sample an action from the solved models according to the posterior weight over the corresponding model.

The assumption of a state oracle in Minerva is unrealistic for many domains.  The approach of~\cite{atrash09} weakens the assumption to an action-based oracle, which can be called on to reveal the optimal POMDP action at any time step for the current belief over states.  This oracle does not consider the uncertainty over model parameters and computes its policy based on the correct parameters.  There are some heuristics to decide when to query the oracle, but this is largely an unexplored question.

An alternative approach for the BAPOMDP is to use the Bayes risk as a criterion for choosing actions~\citep{doshi08,doshi12}.  This framework considers an extended action set, which augments the initial action set with a set of query actions corresponding to a consultation with an oracle.  Unlike the oracular queries used in~\cite{atrash09}, which explicitly ask for the optimal action (given the current belief), the oracular queries in~\cite{doshi08} request confirmation of an optimal action choice:  \emph{I think $a_i$ is the best action. Should I do $a_i$?}  If the oracle answers to the negative, there is a follow-up query: \emph{Then I think $a_j$ is best. Is that correct?}, and so on until a positive answer is received.  In this framework, the goal is to select actions with smallest expected loss, defined here as the Bayes risk (cf.~the regret definition in Chapter~\ref{s:MAB}):
$$BR(a)=\sum_{s,\phi,\psi} Q(b_t(s,\phi,\psi,a)) b_t(s,\phi,\psi) - \sum_{s,\phi,\psi} Q(b_t(s,\phi,\psi,a^*)) b_t(s,\phi,\psi), $$
where $Q(\cdot)$ is just Eq.~\ref{eq:ValFuncBAPOMDP} without the maximization over actions and $a^*$ is the optimal action at $b_t(s,\phi,\psi)$. Because the second term is independent of the action choice, to minimize the Bayes risk, we simply consider maximizing the first term over actions.  The analysis of this algorithm has yielded a bound on the number of samples needed to ensure $\epsilon$-error. The bound is quite loose in practice, but at least provides some upper-bound on the number of queries needed to achieve good performance.  Note that this analysis is based on the Bayes risk criterion that provides a myopic view of uncertainty (i.e.,~it assumes that the next action will resolve the uncertainty over models).

The planning approach suggested by~\cite{ross11} aims to approximate the optimal BAPOMDP strategy by employing a forward search similar to that outlined in Algorithm~\ref{alOnline}.  In related work,~\cite{png11} use a branch-and-bound algorithm to approximate the BAPOMDP solution. Many of the other techniques outlined in Table~\ref{table:onlineBRL} could also be extended to the BAPOMDP model.

Finally, it is worth mentioning that the method of~\cite{dallaire09}, described in \S\ref{ss:ext-cmdp}, is also able to handle continuous partially observable domains by using an additional Gaussian process for the observation function.

\vspace{-5pt}
\section{Extensions to Other Priors and Structured MDPs}
\label{ss:ext-fmdp}

The work and methods presented above focus on the case where the model representation consists of a  discrete (flat) set of states.  For many larger domains, it is common to assume that the states are arranged in a more sophisticated structure, whether a simple clustering or more complicated hierarchy.  It is therefore interesting to consider how Bayesian reinforcement learning methods can be used in those cases.

The simplest case, called \emph{parameter tying} in previous sections, corresponds to the case where states are grouped according to a pre-defined clustering assignment.  In this case, it is common to aggregate learned parameters according to this assignment~\citep{poupart06beetle,sorg10,png11}.  The advantage is that there are fewer parameters to estimate, and thus, learning can be achieved with fewer samples.  The main downside is that this requires a hand-coded assignment.  In practice, it may also be preferable to use a coarser clustering than is strictly correct in order to improve variance (at the expense of more bias).  This is a standard model selection problem.

More sophisticated approaches have been proposed to automate the process of clustering.  In cases where the (unknown) model parameters can be assumed to be sparse, it is possible to incorporate a sparseness assumption in the Dirichlet estimation through use of a hierarchical prior~\citep{friedman99}. An alternative is to maintain a posterior over the state clustering.   Non-parametric models of state clustering have been considered using a Chinese Restaurant Process~\citep{asmuth09,asmuth11}. These could be extended to many of the other model-based BRL methods described above.  A related approach was proposed for the case of partially observable MDPs, where the posterior is expressed over the set of latent (unobserved) states and is represented using a hierarchical Dirichlet process~\citep{doshi09}.

A sensibly different approach proposes to express the prior over the space of policies, rather than over the space of parametric models~\citep{doshi10}.  The goal here is to leverage trajectories acquired from an expert planner, which can be used to define this prior over policies.  It is assumed that the expert knew something about the domain when computing its policy. An interesting insight is to use the infinite POMDP model~\citep{doshi09} to specify the policy prior, by simply reversing the role of actions and observations; a preference for simpler policies can be expressed by appropriately setting the hyper-parameters of the hierarchical Dirichlet process.

A few works have considered the problem of model-based BRL in cases where the underlying MDP has specific structure.  In the case of factored MDPs,~\cite{ross08uai} show that it is possible to simultaneously maintain a posterior over the structure and the parameters of the domain.  The structure in this case captures the bipartite dynamic Bayes network that describes the state-to-state transitions.  The prior over structures is maintained using an MCMC algorithm with appropriate graph to graph transition probabilities.  The prior over model parameters is conditioned on the graph structure. Empirical results show that in some domains, it is more efficient to simultaneously estimate the structure and the parameters, rather than estimate only the parameters given a known structure.

When the model is parametrized,  \cite{GopalanMannor15} use an information theoretic approach to quickly find the set of ``probable'' parameters in a pseudo-Bayesian setting. They show that
the regret can be exponentially lower than models where the model is flat. Furthermore, the analysis can be done in a frequentist fashion leading eventually to a logarithmic regret.

In multi-task reinforcement learning, the goal is learn a good policy over a distribution of MDPs. A naive approach would be to assume that all observations are coming from a single model, and apply Bayesian RL to estimate this mean model.  However by considering a hierarchical infinite mixture model over MDPs (represented by a hierarchical Dirichlet process), the agent is able to learn a distribution over different classes of MDPs, including estimating the number of classes and the parameters of each class~\citep{Wilson07MR}.



\chapter{Model-free Bayesian Reinforcement Learning}
\label{s:model-free-brl}

As discussed in \S\ref{ss:rl}, model-free RL methods are those that do not explicitly learn a model of the system and only use sample trajectories obtained by direct interaction with the system. In this section, we present a family of value function Bayesian RL (BRL) methods, called Gaussian process temporal difference (GPTD) learning, and two families of policy search BRL techniques: a class of Bayesian policy gradient (BPG) algorithms and a class of Bayesian actor-critic (BAC) algorithms.


\section{Value Function Algorithms}
\label{ss:value-alg}

As mentioned in \S\ref{ss:rl}, value function RL methods search in the space of value functions, functions from the state (state-action) space to real numbers, to find the optimal value (action-value) function, and then use it to extract an optimal policy. In this section, we focus on the policy iteration (PI) approach and start by tackling the policy evaluation problem, i.e., the process of estimating the value (action-value) function $V^\mu$ ($Q^\mu$) of a given policy $\mu$ (see \S\ref{ss:rl}). In policy evaluation, the quantity of interest is the value function of a given policy, which is unfortunately hidden. However, a closely related random variable, the reward signal, is observable. Moreover, these hidden and observable quantities are related through the Bellman equation (Eq.~\ref{eq:Bellman-eq}). Thus, it is possible to extract information about the value function from the noisy samples of the reward signals. A Bayesian approach to this problem employs the Bayesian methodology to infer a posterior distribution over value functions, conditioned on the state-reward trajectory observed while running a MDP. Apart from the value estimates given by the posterior mean, a Bayesian solution also provides the variance of values around this mean, supplying the practitioner with an accuracy measure of the value estimates.

In the rest of this section, we study a Bayesian framework that uses a Gaussian process (GP) approach to this problem, called Gaussian process temporal difference (GPTD) learning~\citep{Engel03BM,Engel05RL,Engel05AR}. We then show how this Bayesian policy evaluation method (GPTD) can be used for control (to improve the policy and to eventually find a good or an optimal policy) and present a Bayesian value function RL method, called GPSARSA.


\subsection{Gaussian Process Temporal Difference Learning}
\label{sss:GPTD}

Gaussian process temporal difference (GPTD) learning~\citep{Engel03BM,Engel05RL,Engel05AR} is a GP-based framework that uses linear statistical models (see \S\ref{sss:GP}) to relate, in a probabilistic way, the underlying hidden value function with the observed rewards, for a MDP controlled by some fixed policy $\mu$. The GPTD framework may be used with both parametric and non-parametric representations of the value function, and applied to general MDPs with infinite state and action spaces. Various flavors of the basic model yields several different online algorithms, including those designed for learning action-values, providing the basis for model-free policy improvement, and thus, full PI algorithms.

Since the focus of this section is on policy evaluation, to simplify the notation, we remove the dependency to the policy $\mu$ and use $D$, $P$, $R$, and $V$ instead of $D^\mu$, $P^\mu$, $R^\mu$, and $V^\mu$. As shown in \S\ref{ss:mdp}, the value $V$ is the result of taking the expectation of the discounted return $D$ with respect to the randomness in the trajectory and in the rewards collected therein (Eq.~\ref{eq:val-func}). In the classic frequentist approach, $V$ is not random, since it is the true, albeit unknown, value function of policy $\mu$. On the other hand, in the Bayesian approach, $V$ is viewed as a random entity by assigning it additional randomness due to our subjective uncertainty regarding the MDP's transition kernel $P$ and reward function $q$ ({\em intrinsic} uncertainty). We do not know what the true functions $P$ and $q$ are, which means that we are also uncertain about the true value function. We model this additional {\em extrinsic} uncertainty by defining $V$ as a random process indexed by the state variable $x$. In order to separate the two sources of uncertainty inherent in the discounted return process $D$, we decompose it as
\begin{equation}\label{eq:D-decompose}
D(s)=\exptE\big[D(s)\big]+D(s)-\exptE\big[D(s)\big]=V(s)+\Delta V(s),
\end{equation}
where
\begin{equation}\label{eq:residual}
\Delta V(s)\stackrel{\text{def}}{=}D(s)-V(s)
\end{equation}
is a zero-mean residual. When the MDP's model is known, $V$ becomes deterministic and the randomness in $D$ is fully attributed to the intrinsic randomness in the state-reward trajectory, modelled by $\Delta V$. On the other hand, in a MDP in which both the transitions and rewards are deterministic but unknown, $\Delta V$ becomes zero (deterministic), and the randomness in $D$ is due solely to the extrinsic uncertainty, modelled by $V$.

We write the following Bellman-like equation for $D$ using its definition (Eq.~\ref{eq:D-def})
\begin{equation}\label{eq:D-Bellman}
D(s)=R(s)+\gamma D(s'),\hspace{0.2in}s'\sim P(\cdot|s).
\end{equation}
Substituting Eq.~\ref{eq:D-decompose} into Eq.~\ref{eq:D-Bellman} and rearranging we obtain\footnote{Note that in Eq.~\ref{eq:GPTD-model1}, we removed the dependency of $N$ to policy $\mu$ and replaced $N^\mu$ with $N$.}
\begin{align}\label{eq:GPTD-model1}
R(s)&=V(s)-\gamma V(s')+N(s,s'),\hspace{0.2in}s'\sim P(\cdot|s),\hspace{0.2in}\text{where} \nonumber \\
N(s,s')&\stackrel{\text{def}}{=}\Delta V(s)-\gamma\Delta V(s').
\end{align}
When we are provided with a system trajectory $\xi$ of size $T$, we write the model-equation~\eqref{eq:GPTD-model1} for $\xi$, resulting in the following set of $T$ equations
\begin{equation}\label{eq:GPTD-model2}
R(s_t)=V(s_t)-\gamma V(s_{t+1})+N(s_t,s_{t+1}),\hspace{0.2in}t=0,\ldots,T-1.
\end{equation}
By defining $\vecR_T=\big(R(s_0),\ldots,R(s_{T-1})\big)^\top$, $\vecV_{T+1}=\big(V(s_0),\ldots,V(s_T)\big)^\top$, and $\vecN_T=\big(N(s_0,s_1),\ldots,N(s_{T-1},s_T)\big)^\top$, we write the above set of $T$ equations as
\begin{equation}\label{eq:GPTD-model3}
\vecR_T=\matH\vecV_{T+1}+\vecN_T\;,
\end{equation}
where $\matH$ is the following $T\times (T+1)$ matrix:
\begin{equation}\label{eq:H}
\matH = \left[ \begin{array}{cccccc}
1 & -\gamma & 0 & \ldots & 0 & 0\\
0 & 1 & -\gamma & \ldots & 0 & 0\\
\vdots & \vdots & \vdots & & \vdots & \vdots\\
0 & 0 & 0 &\ldots & 1 & -\gamma
\end{array} \right].
\end{equation}
Note that in episodic problems, if a goal-state is reached at time-step $T$, the final equation in~\eqref{eq:GPTD-model2} is
\begin{equation}\label{eq:GPTD-model4-1}
R(s_{T-1})=V(s_{T-1})+N(s_{T-1},s_T),
\end{equation}
and thus, Eq.~\ref{eq:GPTD-model3} becomes
\begin{equation}\label{eq:GPTD-model4}
\vecR_T=\matH\vecV_T+\vecN_T\;,
\end{equation}
where $\matH$ is the following $T\times T$ square invertible matrix with determinant equal to $1$,

\vspace{-0.1in}
\begin{small}
\begin{equation}\label{eq:H-episodic}
\matH \!=\! \left[ \begin{array}{cccccc}
1 & -\gamma & 0 & \ldots & 0 & 0\\
0 & 1 & -\gamma & \ldots & 0 & 0\\
\vdots & \vdots & \vdots & & \vdots & \vdots\\
0 & 0 & 0 &\ldots & 1 & -\gamma\\
0 & 0 & 0 &\ldots & 0 & 1
\end{array} \right]\;\text{and}\;\;
\matH^{-1} \!=\! \left[ \begin{array}{ccccc}
1 & \gamma & \gamma^2 & \ldots & \gamma^{T-1}\\
0 & 1 & \gamma & \ldots & \gamma^{T-2}\\
\vdots & \vdots & \vdots & & \vdots\\
0 & 0 & 0 &\ldots & 1
\end{array} \right].
\end{equation}
\end{small}

\begin{figure}[t]
\centering
\includegraphics[width=\textwidth]{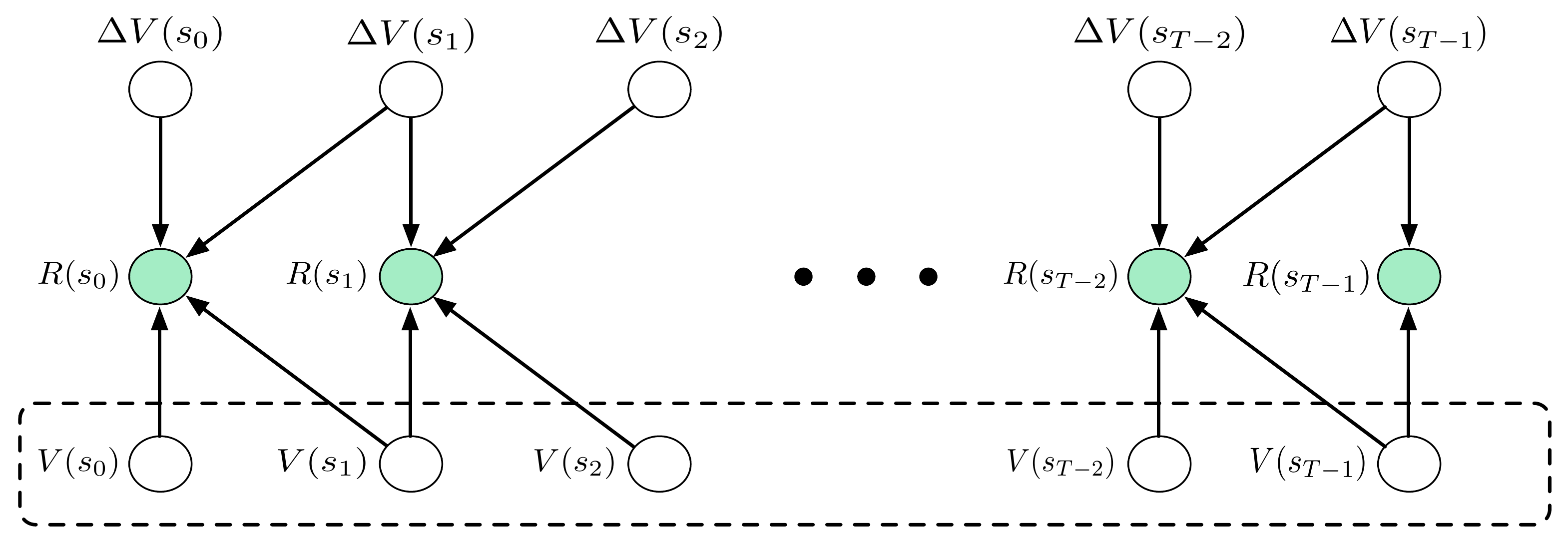}
\caption{A graph illustrating the conditional independencies between the latent $V(s_t)$ value variables (bottom row), the noise variables $\Delta V(s_t)$ (top row), and the observable $R(s_t)$ reward variables (middle row), in the GPTD model. As in the case of GP regression, all of the $V(s_t)$ variables should be connected by arrows, due to the dependencies introduced by the prior. To avoid cluttering the diagram, this was marked by the dashed frame surrounding them.}
\label{fig:GPTD-graphical}
\end{figure}

\vspace{-0.1in}
\noindent
Figure~\ref{fig:GPTD-graphical} illustrates the conditional dependency relations between the latent value variables $V(s_t)$, the noise variables $\Delta V(s_t)$, and the observable rewards $R(s_t)$. Unlike the GP regression diagram of Figure~\ref{fig:GP-reg}, there are vertices connecting variables from different time steps, making the ordering of samples important. Since the diagram in Figure~\ref{fig:GPTD-graphical} is for the episodic setting, also note that for the last state in each episode ($s_{T-1}$ in this figure), $R(s_{T-1})$ depends only on $V(s_{T-1})$ and $\Delta V(s_{T-1})$ (as in Eqs.~\ref{eq:GPTD-model4-1} and~\ref{eq:GPTD-model4}).

In order to fully define a probabilistic generative model, we also need to specify the distribution of the noise process $\vecN_T$. In order to derive the noise distribution, we make the following two assumptions (see Appendix~\ref{gptd-assumptions} for a discussion about these assumptions):

\begin{ass}\label{ass:residual1}
The residuals $\Delta \vecV_{T+1}=\big(\Delta V(s_0),\ldots,\Delta V(s_T)\big)^\top$ can be modeled as a Gaussian process.
\end{ass}


\begin{ass}\label{ass:residual2}
Each of the residuals $\Delta V(s_t)$ is generated independently of all the others, i.e., $\exptE\big[\Delta V(s_i)\Delta V(s_j)\big]=0$, for $i\neq j$.
\end{ass}

By definition (Eq.~\ref{eq:residual}), $\exptE\big[\Delta V(s)\big]=0$ for all $s$. Using Assumption~\ref{ass:residual2}, it is easy to show that $\exptE\big[\Delta V(x_t)^2\big]=\Var\big[D(x_t)\big]$. Thus, denoting $\Var\big[D(x_t)\big]=\sigma^2_t$, Assumption~\ref{ass:residual1} may be written as $\Delta \vecV_{T+1}\sim\N\big(\vec0,\diag(\vecsigma_{T+1})\big)$. Since $\vecN_T=\matH\Delta\vecV_{T+1}$, we have $\vecN_T\sim\N(\vec0,\matSigma)$ with

\begin{small}
\begin{align}\label{eq:noise-cov}
&\matSigma = \matH\diag(\vecsigma_{T+1})\matH^\top \\
&= \left[ \begin{array}{cccccc}
\sigma_0^2+\gamma^2\sigma_1^2 & -\gamma\sigma_1^2 & 0 & \ldots & 0 & 0\\
-\gamma\sigma_1^2 & \sigma_1^2+\gamma^2\sigma_2^2 & -\gamma\sigma_2^2 & \ldots & 0 & 0\\
\vdots & \vdots & \vdots & & \vdots & \vdots\\
0 & 0 & 0 &\ldots & \sigma_{T-2}^2+\gamma^2\sigma_{T-1}^2 & -\gamma\sigma_{T-1}^2\\
0 & 0 & 0 &\ldots & -\gamma\sigma_{T-1}^2 & \sigma_{T-1}^2+\gamma^2\sigma_T^2
\end{array} \right]. \nonumber
\end{align}
\end{small}

\vspace{-0.1in}
\noindent
Eq.~\ref{eq:noise-cov} indicates that the Gaussian noise process $\vecN_T$ is colored with a tri-diagonal covariance matrix. If we assume that for all $t=0,\ldots,T$, $\sigma_t=\sigma$, then $\diag(\vecsigma_{T+1})=\sigma^2\matI$ and Eq.~\ref{eq:noise-cov} may be simplified and written as
\begin{equation*}
\matSigma = \sigma^2\matH\matH^\top=
\sigma^2\left[ \begin{array}{cccccc}
1+\gamma^2 & -\gamma & 0 & \ldots & 0 & 0\\
-\gamma & 1+\gamma^2 & -\gamma & \ldots & 0 & 0\\
\vdots & \vdots & \vdots & & \vdots & \vdots\\
0 & 0 & 0 &\ldots & 1+\gamma^2 & -\gamma\\
0 & 0 & 0 &\ldots & -\gamma & 1+\gamma^2
\end{array} \right].
\end{equation*}

Eq.~\ref{eq:GPTD-model3} (or in case of episodic problems Eq.~\ref{eq:GPTD-model4}), along with the measurement noise distribution of Eq.~\ref{eq:noise-cov}, and a prior distribution for $V$ (defined either parametrically or non-parametrically, see \S\ref{sss:GP}), completely specify a statistical generative model relating the value and reward random processes. In order to infer value estimates from a sequence of observed rewards, Bayes' rule can be applied to this generative model to derive a posterior distribution over the value function conditioned on the observed rewards.

\begin{figure}[ht]
\centering
\includegraphics[scale=0.33]{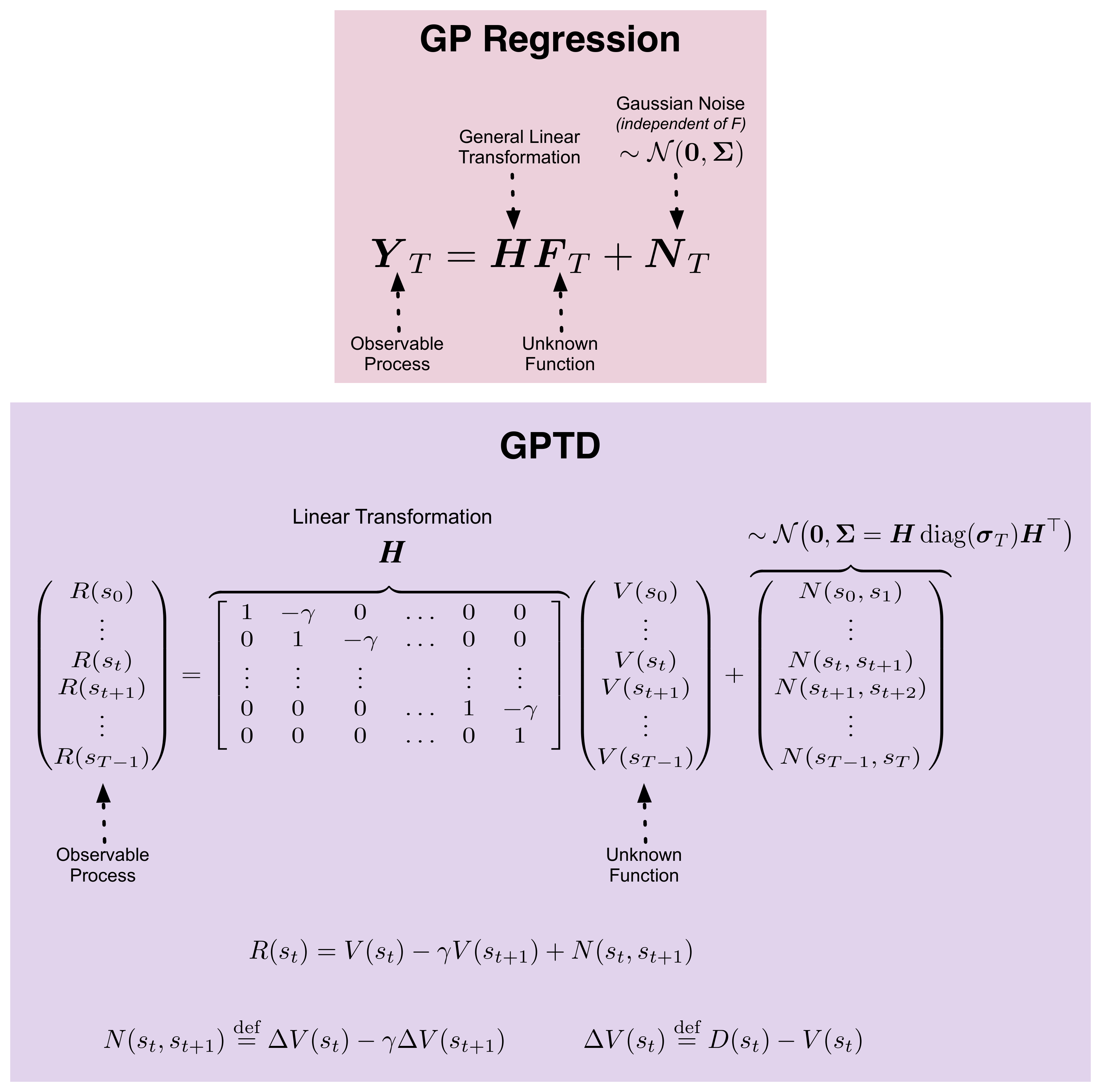}
\caption{The connection between the GPTD model described in \S\ref{sss:GPTD} and the family of linear statistical models and GP regression discussed in \S\ref{sss:GP}.}
\label{fig:GP-GPTD}
\end{figure}

In the case in which we place a Gaussian prior over $\vecV_T$, both $\vecV_T$ and $\vecN_T$ are normally distributed, and thus, the generative model of Eq.~\ref{eq:GPTD-model3} (Eq.~\ref{eq:GPTD-model4}) will belong to the family of linear statistical models discussed in \S\ref{sss:GP}. Consequently, both parametric and non-parametric treatments of this model (see \S\ref{sss:GP}) may be applied in full to the generative model of Eq.~\ref{eq:GPTD-model3} (Eq.~\ref{eq:GPTD-model4}), with $\matH$ given by Eq.~\ref{eq:H} (Eq.~\ref{eq:H-episodic}). Figure~\ref{fig:GP-GPTD} demonstrates how the GPTD model described in this section is related to the family of linear statistical models and GP regression discussed in \S\ref{sss:GP}.

We are now in a position to write a closed form expression for the posterior moments of $V$, conditioned on an observed sequence of rewards $\vecr_T=\big(r(s_0),\ldots,r(s_{T-1})\big)^\top$, and derive GPTD-based algorithms for value function estimation. We refer to this family of algorithms as Monte-Carlo GPTD (MC-GPTD) algorithms. \\

\begin{paragraph}
{\bf Parametric Monte-Carlo GPTD Learning:} In the parametric setting, the value process is parameterized as $V(s)=\vecphi(s)^\top\vecW$, and therefore, $\vecV_{T+1}=\matPhi^\top\vecW$ with $\matPhi=\big[\vecphi(s_0),\ldots,\vecphi(s_T)\big]$. As in \S\ref{sss:GP}, if we use the prior distribution $\vecW\sim\N(\vec0,\matI)$, the posterior moments of $\vecW$ are given by
\begin{align}\label{eq:post-param1}
\exptE\big[\vecW|\vecR_T=\vecr_T\big]&=\Delta\matPhi(\Delta\matPhi^\top\Delta\matPhi+\matSigma)^{-1}\vecr_T, \nonumber \\
\Cov\big[\vecW|\vecR_T=\vecr_T\big]&=\matI-\Delta\matPhi(\Delta\matPhi^\top\Delta\matPhi+\matSigma)^{-1}\Delta\matPhi^\top,
\end{align}
where $\Delta\matPhi=\matPhi\matH^\top$. To have a smaller matrix inversion, Eq.~\ref{eq:post-param1} may be written as
\begin{align}\label{eq:post-param2}
\exptE\big[\vecW|\vecR_T=\vecr_T\big]&=(\Delta\matPhi\matSigma^{-1}\Delta\matPhi^\top+\matI)^{-1}\Delta\matPhi\matSigma^{-1}\vecr_T, \nonumber \\
\Cov\big[\vecW|\vecR_T=\vecr_T\big]&=(\Delta\matPhi\matSigma^{-1}\Delta\matPhi^\top+\matI)^{-1}.
\end{align}
Note that in episodic problems $\matH$ is invertible, and thus, assuming a constant noise variance, i.e., $\diag(\vecsigma_T)=\sigma^2\matI$, Eq.~\ref{eq:post-param2} becomes
\begin{align}\label{eq:post-param3}
\exptE\big[\vecW|\vecR_T=\vecr_T\big]&=(\matPhi\matPhi^\top+\sigma^2\matI)^{-1}\matPhi\matH^{-1}\vecr_T, \nonumber \\
\Cov\big[\vecW|\vecR_T=\vecr_T\big]&=\sigma^2(\matPhi\matPhi^\top+\sigma^2\matI)^{-1}.
\end{align}
Eq.~\ref{eq:post-param3} is equivalent to Eq.~\ref{eq:param-GPreg2} with $\vecy_T=\matH^{-1}\vecr_T$ (see the discussion of Assumption~\ref{ass:residual2}).

Using Eq.~\ref{eq:post-param2} (or Eq.~\ref{eq:post-param3}), it is possible to derive both a batch algorithm~\citep[Algorithm~17]{Engel05AR} and a recursive online algorithm~\citep[Algorithm~18]{Engel05AR} to compute the posterior moments of the weight vector $\vecW$, and thus, the value function $V(\cdot)=\vecphi(\cdot)^\top\vecW$. \\
\end{paragraph}

\begin{paragraph}
{\bf Non-parametric Monte-Carlo GPTD Learning:} In the non-parametric case, we bypass the parameterization of the value process by placing a prior directly over the space of value functions. From \S\ref{sss:GP} with the GP prior $\vecV_{T+1}\in\N(\vec0,\matK)$, we obtain
\begin{align}\label{eq:post-nonparam1}
\exptE\big[V(s)|\vecR_T=\vecr_T\big]&=\veck(s)^\top\vecalpha,\nonumber \\
\Cov\big[V(s),V(s')|\vecR_T=\vecr_T\big]&=k(s,s')-\veck(s)^\top\matC\veck(s'),
\end{align}
where
\begin{equation}\label{eq:post-nonparam2}
\vecalpha=\matH^\top(\matH\matK\matH^\top+\matSigma)^{-1}\vecr_T\hspace{0.2in}\text{and}\hspace{0.2in}\matC=\matH^\top(\matH\matK\matH^\top+\matSigma)^{-1}\matH.
\end{equation}
Similar to the parametric case, assuming a constant noise variance, Eq.~\ref{eq:post-nonparam2} may be written for episodic problems as
\begin{equation}\label{eq:post-nonparam3}
\vecalpha=(\matK+\sigma^2\matI)^{-1}\matH^{-1}\vecr_T\hspace{0.4in}\text{and}\hspace{0.4in}\matC=(\matK+\sigma^2\matI)^{-1}.
\end{equation}
Eq.~\ref{eq:post-nonparam3} is equivalent to Eq.~\ref{eq:nonparam-GPreg2} with $\vecy_T=\matH^{-1}\vecr_T$ (see the discussion of Assumption~\ref{ass:residual2}).

As in the parametric case, it is possible to derive recursive updates for $\vecalpha$ and $\matC$, and an online algorithm~\citep[Algorithm~19]{Engel05AR} to compute the posterior moments of the value function $V$. \\
\end{paragraph}

\begin{paragraph}
{\bf Sparse Non-parametric Monte-Carlo GPTD Learning:} In the parametric case, the computation of the posterior may be performed online in $O(n^2)$ time per sample and $O(n^2)$ memory, where $n$ is the number of basis functions used to approximate $V$. In the non-parametric case, we have a new basis function for each new sample we observe, making the cost of adding the $t$'th sample $O(t^2)$ in both time and memory. This would seem to make the non-parametric form of GPTD computationally infeasible except in small and simple problems. However, the computational cost of non-parametric GPTD can be reduced by using an online sparsification method (\citealt{Engel02SO},~\citealt[Chapter~2]{Engel05AR}), to a level where it can be efficiently implemented online. %
In many cases, this results in significant computational savings, both in terms of memory and time, when compared to the non-sparse solution. For the resulting algorithm, we refer the readers to the sparse non-parametric GPTD algorithm in~\citet[Table~1]{Engel05RL} or~\citet[Algorithm~20]{Engel05AR}. \\

\end{paragraph}

\subsection{Connections with Other TD Methods}
\label{sss:GPTD-LSTD}

In \S\ref{sss:GPTD}, we showed that the stochastic GPTD model is equivalent to GP regression on MC samples of the discounted return (see the discussion of Assumption~\ref{ass:residual2}). Engel~\citet{Engel05AR} showed that by suitably selecting the noise covariance matrix $\matSigma$ in the stochastic GPTD model, it is possible to obtain GP-based variants of LSTD($\lambda$) algorithm~\citep{Bradtke96LL,Boyan99LT}. The main idea is to obtain the value of the weight vector $\vecW$ in the parametric GPTD model by carrying out maximum likelihood (ML) inference ($\vecW$ is the value for which the observed data is most likely to be generated by the stochastic GPTD model). This allows us to derive LSTD($\lambda$) for each value of $\lambda\leq1$, as an ML solution arising from some GPTD generative model with a specific noise covariance matrix $\matSigma$.

\subsection{Policy Improvement with GPSARSA}
\label{sss:GPSARSA}

SARSA~\citep{Rummery94OQ} is a  straightforward extension of the TD algorithm~\citep{Sutton88LP} to control, in which action-values are estimated. This allows policy improvement steps to be performed without requiring any additional knowledge of the MDP model. The idea is that under the policy $\mu$, we may define a process with state space $\Z=\Scal\times\A$ (the space of state-action pairs) and the reward model of the MDP. This process is Markovian with transition density $P^\mu(z'|z)=P^\mu(s'|s)\mu(a'|s')$, where $z=(s,a)$ and $z'=(s',a')$ (see \S\ref{ss:mdp} for more details on this process). SARSA is simply the TD algorithm applied to this process. The same reasoning may be applied to derive a GPSARSA algorithm from a GPTD algorithm. This is equivalent to rewriting the model-equations of \S\ref{sss:GPTD} for the action-value function $Q^\mu$, which is a function of $z$, instead of the value function $V^\mu$, a function of $s$.

In the non-parametric setting, we simple need to define a covariance kernel function over state-action pairs, i.e., $k:\Z\times\Z\rightarrow\Re$. Since states and actions are different entities, it makes sense to decompose $k$ into a state-kernel $k_s$ and an action-kernel $k_a$: $k(z,z')=k_s(s,s')k_a(a,a')$. If both $k_s$ and $k_a$ are kernels, we know that $k$ is also a kernel~\citep{Scholkopf02LK,ShaweTaylor04KM}. The state and action kernels encode our prior beliefs on value correlations between different states and actions, respectively. All that remains is to run the GPTD algorithms (sparse, batch, online) on the state-action-reward sequence using the new state-action kernel. Action selection may be performed by the $\epsilon$-greedy exploration strategy. The main difficulty that may arise here is in finding the greedy action from a large or infinite number of possible actions (when $|\A|$ is large or infinite). This may be solved by sampling the value estimates for a few randomly chosen actions and find the greedy action among only them. However, an ideal approach would be to design the action-kernel in such a way as to provide a closed-form expression for the greedy action. Similar to the non-parametric setting, in the parametric case, deriving GPSARSA algorithms from their GPTD counterparts is still straightforward; we just need to define a set of basis functions over $\Z$ and use them to approximate action-value functions.


\subsection{Related Work}
\label{sss:rel-work}

Another approach to employ GPs in RL was proposed by Rasmussen and Kuss \citet{Rasmussen04GP}. The approach taken in this work is notably different from the generative approach of the GPTD framework. Two GPs are used in~\citet{Rasmussen04GP}, one to learn the MDP's transition model and one to estimate the value function. This leads to an inherently offline algorithm. There are several other limitations to this framework. First, the state dynamics is assumed to be factored, in the sense that each state coordinate evolves in time independently of all others. This is a rather strong assumption that is not likely to be satisfied in most real problems. Second, it is assumed that the reward function is completely known in advance, and is of a very special form -- either polynomial or Gaussian. Third, the covariance kernels used are also restricted to be either polynomial or Gaussian or a mixture of the two, due to the need to integrate over products of GPs. Finally, the value function is only modelled at a predefined set of support states, and is solved only for them. No method is proposed to ensure that this set of states is representative in any way.

Similar to most kernel-based methods, the choice of prior distribution significantly affects the empirical performance of the learning algorithm. Reisinger et al.~\citet{Reisinger08OK} proposed an online model (prior covariance function) selection method for GPTD using sequential Monte-Carlo methods and showed that their method yields better asymptotic performance than standard GPTD for many different kernel families.


\section{Bayesian Policy Gradient}
\label{ss:BPG}

Policy gradient (PG) methods are RL algorithms that maintain a class of smoothly parameterized stochastic policies $\big\{\mu(\cdot|s;\vectheta), s \in \Scal, \vectheta \in \Theta\big\}$, and update the policy parameters $\vectheta$ by adjusting them in the direction of an estimate of the gradient of a performance measure (e.g.,~\citealt{Williams92SS,Marbach98SM,Baxter01IP}). As an illustration of such a parameterized policy, consider the following example:
\begin{example}[Online shop -- parameterized policy]
Recall the online shop domain of Example \ref{exm:online_shop}. Assume that for each customer state $s\in \X$, and advertisement $a \in \A$, we are given a set of $n$ numeric values $\phi(s,a)\in \Re^n$ that represent some features of the state and action. For example, these features could be the number of items in the cart, the average price of items in the cart, the age of the customer, etc. A popular parametric policy representation is the softmax policy, defined as
\begin{equation*}
    \mu(a|s;\vectheta) = \frac{\exp\left(\vectheta^\top \phi(s,a)\right)}{\sum_{a'\in \A}\exp\left(\vectheta^\top \phi(s,a')\right)}.
\end{equation*}
The intuition behind this policy, is that if $\vectheta$ is chosen such that $\vectheta^\top \phi(s,a)$ is an approximation of the state-action value function $Q(s,a)$, then the softmax policy is an approximation of the greedy policy with respect to $Q$ -- which is the optimal policy.
\end{example}
The performance of a policy $\mu$ is often measured by its {\em expected return}, $\eta(\mu)$, defined by Eq.~\ref{eq:pol-expt-ret1}. Since in this setting a policy $\mu$ is represented by its parameters $\vectheta$, policy dependent functions such as $\eta(\mu)$ and $\Pr\big(\xi;\mu)$ may be written as $\eta(\vectheta)$ and $\Pr(\xi;\vectheta)$. 

The {\em score function} or {\em likelihood ratio} method has become the most prominent technique for gradient estimation from simulation (e.g.,~\cite{Glynn90LR,Williams92SS}). This method estimates the gradient of the expected return with respect to the policy parameters $\vectheta$, defined by Eq.~\ref{eq:pol-expt-ret1}, using the following equation:\footnote{We use the notation $\nabla$ to denote $\nabla_{\vectheta}$ -- the gradient with respect to the policy parameters.}
\begin{equation}
\label{eq:grad}
\nabla\eta(\vectheta)=\int\bar{\rho}(\xi)\frac{\nabla \Pr(\xi;\vectheta)}{\Pr(\xi;\vectheta)}\Pr(\xi;\vectheta)d\xi.
\end{equation}
In Eq.~\ref{eq:grad}, the quantity
\begin{align}
\label{eq:score}
\vecu(\xi;\vectheta)&=\frac{\nabla \Pr(\xi;\vectheta)}{\Pr(\xi;\vectheta)}=\nabla\log \Pr(\xi;\vectheta) \\
&=\sum_{t=0}^{T-1}\frac{\nabla\mu(a_t|s_t;\vectheta)}{\mu(a_t|s_t;\vectheta)} = \sum_{t=0}^{T-1}\nabla\log\mu(a_t|s_t;\vectheta) \nonumber
\end{align}
is called the (Fisher) score function or likelihood ratio of trajectory $\xi$ under policy $\vectheta$. Most of the work on PG has used classical Monte-Carlo (MC) to estimate the (integral) gradient in Eq.~\ref{eq:grad}. These methods (in their simplest form) generate $M$ i.i.d.~sample paths $\xi_1,\ldots,\xi_M$ according to $\Pr(\xi;\vectheta)$, and estimate the gradient $\nabla\eta(\vectheta)$ using the {\em unbiased} MC estimator

\vspace{-0.1in}
\begin{small}
\begin{equation}
\label{eq:grad_MC}
\widehat{\nabla\eta}(\vectheta)=\frac{1}{M}\sum_{i=1}^M\rho(\xi_i)\nabla\log \Pr(\xi_i;\vectheta)=\frac{1}{M}\sum_{i=1}^M\rho(\xi_i)\sum_{t=0}^{T_i-1}\nabla\log\mu(a_{t,i}|s_{t,i};\vectheta).
\end{equation}
\end{small}

\vspace{-0.1in}
\noindent
Both theoretical results and empirical evaluations have highlighted a major shortcoming of these algorithms: namely, the high variance of the gradient estimates, which in turn results in sample-inefficiency (e.g.,~\cite{Marbach98SM,Baxter01IP}). Many solutions have been proposed for this problem, each with its own pros and cons.
%
%
In this section, we describe Bayesian policy gradient (BPG) algorithms that tackle this problem by using a Bayesian alternative to the MC estimation of Eq.~\ref{eq:grad_MC}~\citep{Ghavamzadeh06BP}. These BPG algorithms are based on {\em Bayesian quadrature}, i.e.,~a Bayesian approach to integral evaluation, proposed by~O'Hagan~\citet{Ohagan91BQ}.\footnote{O'Hagan~\citet{Ohagan91BQ} mentions that this approach may be traced even as far back as the work by Poincar{\'e}~\citet{Poincare1896CP}.} The algorithms use Gaussian processes (GPs) to define a prior distribution over the gradient of the expected return, and compute its posterior conditioned on the observed data. This reduces the number of samples needed to obtain accurate (integral) gradient estimates. Moreover, estimates of the natural gradient as well as a measure of the uncertainty in the gradient estimates, namely the gradient covariance, are provided at little extra cost. In the next two sections, we first briefly describe Bayesian quadrature and then present the BPG models and algorithms.


\subsection{Bayesian Quadrature}
\label{subsubsec:BQ}

Bayesian quadrature (BQ)~\citep{Ohagan91BQ}, as its name suggests, is a Bayesian method for evaluating an integral using samples of its integrand. Let us consider the problem of evaluating the following integral\footnote{Similar to~\citet{Ohagan91BQ}, here for simplicity we consider the case where the integral to be estimated is a scalar-valued integral. However, the results of this section can be extended to vector-valued integrals, such as the gradient of the expected return with respect to the policy parameters that we shall study in \S\ref{subsubsec:BPG-algo} (see~\cite{Ghavamzadeh12BP-TECH}).}
\begin{equation}
\label{eq:integral}
\zeta=\int f(x)g(x)dx .
\end{equation}
BQ is based on the following reasoning: In the Bayesian approach, $f(\cdot)$ is random simply because it is unknown. We are therefore uncertain about the value of $f(x)$ until we actually evaluate it. In fact, even then, our uncertainty is not always completely removed, since measured samples of $f(x)$ may be corrupted by noise. Modeling $f$ as a GP means that our uncertainty is completely accounted for by specifying a Normal prior distribution over the functions, $f(\cdot) \sim \N\big(\bar{f}(\cdot),k(\cdot,\cdot)\big)$, i.e.,
\begin{equation*}
\exptE\big[f(x)\big] = \bar{f}(x) \quad \text{and} \quad \Cov\big[f(x),f(x')\big]=k(x,x'), \quad \forall x,x'\in\X.
\end{equation*}
The choice of kernel function $k$ allows us to incorporate prior knowledge on the smoothness properties of the integrand into the estimation procedure. When we are provided with a set of samples $\D_M = \big\{\big(x_i,y(x_i)\big)\big\}_{i=1}^M$, where $y(x_i)$ is a (possibly noisy) sample of $f(x_i)$, we apply Bayes' rule to condition the prior on these sampled values. If the measurement noise is normally distributed, the result is a Normal posterior distribution of $f|\D_M$. The expressions for the posterior mean and covariance are standard (see Eqs.~\ref{eq:lin-stat-mod4}--\ref{eq:lin-stat-mod6}):
\begin{align}
\label{eq:f_post}
\exptE\big[f(x)|\D_M\big] &= \bar{f}(x) + \veck(x)^\top\matC (\vecy - \bvecf), \nonumber\\
\Cov\big[f(x),f(x')|\D_M\big] &= k(x,x') - \veck(x)^\top\matC \veck(x'),
\end{align}
where $\matK$ is the kernel (or Gram) matrix, and $[\matSigma]_{i,j}$ is the measurement noise covariance between the $i$th and $j$th samples. It is typically assumed that the measurement noise is i.i.d.,~in which case $\matSigma = \sigma^2 \matI$, where $\sigma^2$ is the noise variance and $\matI$ is the ($M \times M$) identity matrix.
\begin{align*}
\bvecf&=\big(\bar{f}(x_1),\ldots,\bar{f}(x_M)\big)^\top, \quad \veck(x)=\big(k(x_1,x),\ldots,k(x_M,x)\big)^\top, \\
\vecy&=\big( y(x_1),\ldots,y(x_M) \big)^\top, \quad [\matK]_{i,j}=k(x_i,x_j), \quad \matC=\left(\matK+\matSigma \right)^{-1}.
\end{align*}

Since integration is a linear operation, the posterior distribution of the integral in Eq.~\ref{eq:integral} is also Gaussian, and the posterior moments are given by~\citep{Ohagan91BQ}
\begin{align}
\label{eq:rho_post}
\exptE[\zeta|\D_M] &= \int \exptE\big[f(x)|\D_M\big] g(x)dx , \nonumber\\
\Var[\zeta|\D_M] &= \iint \Cov\big[f(x),f(x')|\D_M\big] g(x)g(x')dxdx'.
\end{align}
Substituting Eq.~\ref{eq:f_post} into Eq.~\ref{eq:rho_post}, we obtain
\begin{equation*}
\exptE[\zeta|\D_M] = \zeta_0 + \vecb^\top\matC (\vecy - \bvecf)\hspace{0.25in}\text{and}\hspace{0.25in} \Var[\zeta|\D_M] = b_0 - \vecb^\top\matC \vecb,
\end{equation*}
where we made use of the definitions

\vspace{-0.15in}
\begin{small}
\begin{equation}
\label{eq:q}
\zeta_0 = \!\int \bar{f}(x) g(x)dx\;,\; \vecb = \!\int \veck(x) g(x)dx\;,\; b_0 = \!\iint k(x,x') g(x)g(x')dxdx'.
\end{equation}
\end{small}

\vspace{-0.15in}
\noindent
Note that $\zeta_0$ and $b_0$ are the prior mean and variance of $\zeta$, respectively. It is important to note that in order to prevent the problem from ``degenerating into infinite regress'', as phrased by O'Hagan~\citet{Ohagan91BQ}, we should decompose the integrand into parts: $f$ (the GP part) and $g$, and select the GP prior, i.e.,~prior mean $\bar{f}$ and covariance $k$, so as to allow us to solve the integrals in Eq.~\ref{eq:q} analytically. Otherwise, we begin with evaluating one integral (Eq.~\ref{eq:integral}) and end up with evaluating three integrals (Eq.~\ref{eq:q}).


\subsection{Bayesian Policy Gradient Algorithms}
\label{subsubsec:BPG-algo}

In this section, we describe the Bayesian policy gradient (BPG) algorithms~\citep{Ghavamzadeh06BP}. These algorithms use Bayesian quadrature (BQ) to estimate the gradient of the expected return with respect to the policy parameters  (Eq.~\ref{eq:grad}). In the Bayesian approach, the expected return of the policy characterized by the parameters $\vectheta$
\begin{equation}
\eta_B(\vectheta) = \int\bar{\rho}(\xi) \Pr(\xi;\vectheta) d\xi
\label{eq:eta_bayes}
\end{equation}
is a random variable because of our subjective Bayesian uncertainty concerning the process generating the return. Under the quadratic loss, the optimal Bayesian performance measure is the posterior expected value of $\eta_B(\vectheta)$, $\exptE\big[\eta_B(\vectheta)|\D_M\big]$. However, since we are interested in optimizing the performance rather than in evaluating it, we would rather evaluate the posterior mean of the gradient of $\eta_B(\vectheta)$ with respect to the policy parameters $\vectheta$, i.e.,
\begin{align}
\label{eq:grad:Bayes}
\nabla\exptE\big[\eta_B(\vectheta)|\D_M\big]&=\exptE\big[\nabla\eta_B(\vectheta)|\D_M\big] \\
 &=\exptE\left[\int \bar{\rho}(\xi)\frac{\nabla \Pr(\xi;\vectheta)}{\Pr(\xi;\vectheta)}\Pr(\xi;\vectheta)d\xi \Big| \D_M\right].
 \nonumber
\end{align}

\begin{paragraph}{Gradient Estimation:} In BPG, we cast the problem of estimating the gradient of the expected return (Eq.~\ref{eq:grad:Bayes}) in the form of Eq.~\ref{eq:integral} and use the BQ approach described in
\S\ref{subsubsec:BQ}. We partition the integrand into two parts, $f(\xi;\vectheta)$ and $g(\xi;\vectheta)$, model $f$ as a GP, and assume that $g$ is a function known to us. We then proceed by calculating the posterior moments of the gradient  $\nabla\eta_B(\vectheta)$ conditioned on the observed data. Ghavamzadeh and Engel~\citet{Ghavamzadeh06BP} proposed two different ways of partitioning the integrand in Eq.~\ref{eq:grad:Bayes}, resulting in the two distinct Bayesian models summarized in Table~\ref{tab:models} (see~\cite{Ghavamzadeh12BP-TECH}, for more details). Figure~\ref{fig:BQ-BPG-BAC} shows how the BQ approach has been used in each of the two BPG models of~\citet{Ghavamzadeh06BP}, summarized in Table~\ref{tab:models}, as well as in the Bayesian actor-critic (BAC) formulation of \S\ref{ss:ac-alg}.

\begin{figure}[!h]
\centering
\includegraphics[scale=0.3]{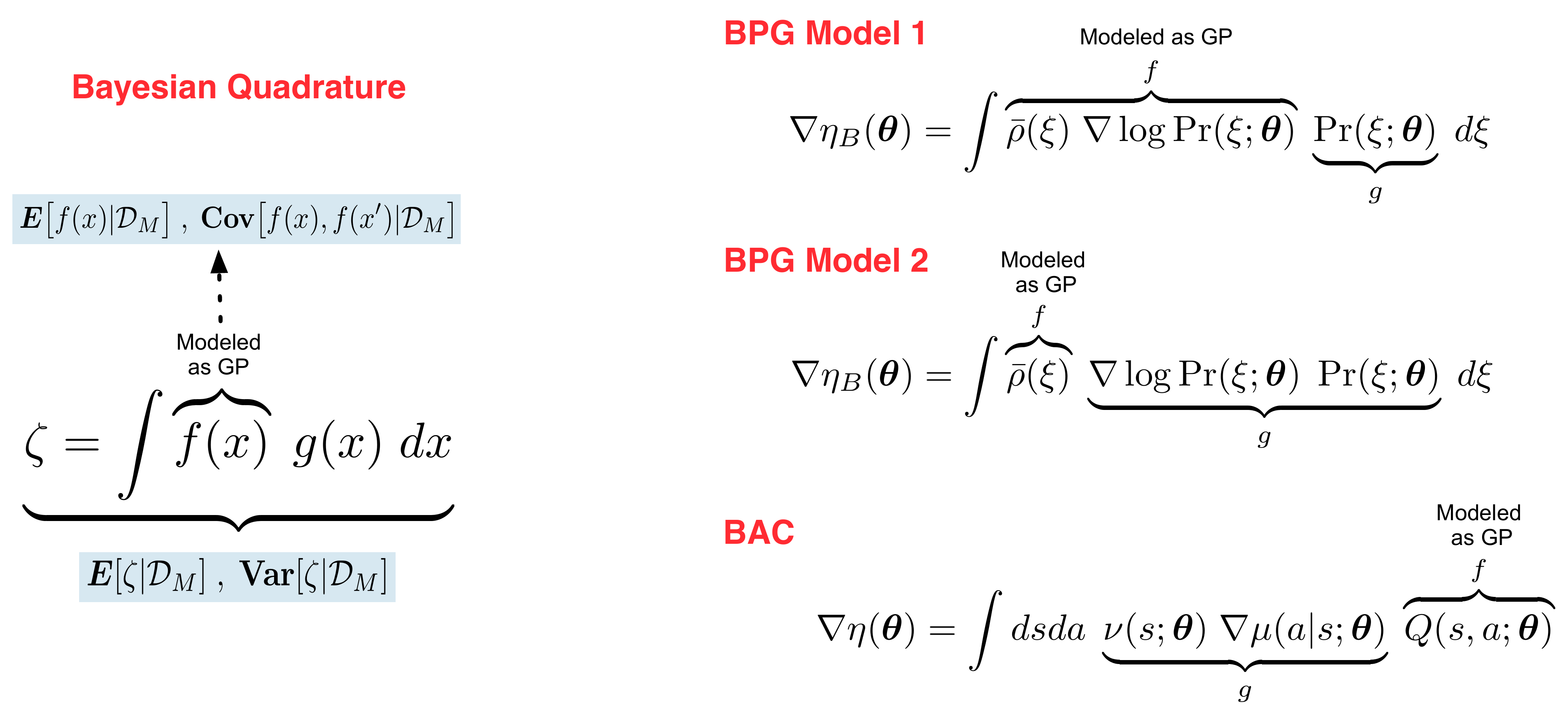}
\caption{The connection between BQ and the two BPG models of~\citet{Ghavamzadeh06BP}, and the Bayesian actor-critic (BAC) formulation of \S\ref{ss:ac-alg}.}
\label{fig:BQ-BPG-BAC}
\end{figure}

It is important to note that in both models $\bar{\rho}(\xi)$ belongs to the GP part, i.e.,~$f(\xi;\vectheta)$. This is because in general, $\bar{\rho}(\xi)$ cannot be known exactly, even for a given $\xi$ (due to the stochasticity of the rewards), but can be estimated for a sample trajectory $\xi$. The more important and rather critical point is that $\Pr(\xi;\vectheta)$ cannot belong to either the $f(\xi;\vectheta)$ or $g(\xi;\vectheta)$ part of the model, since it is not known (in model-free setting) and cannot be estimated for a sample trajectory $\xi$. However, Ghavamzadeh and Engel~\citet{Ghavamzadeh06BP} showed that interestingly, it is sufficient to assign $\Pr(\xi;\vectheta)$ to the $g(\xi;\vectheta)$ part of the model and use an appropriate Fisher kernel (see Table~\ref{tab:models} for the kernels), rather than having exact knowledge of $\Pr(\xi;\vectheta)$ itself (see~\cite{Ghavamzadeh12BP-TECH} for more details).

\begin{table}[!t]
\begin{center}
\begin{tabular}{|l|l|l|}
\hline
& {\bf Model 1} & {\bf Model 2} \\
\hline
Deter. factor ($g$)
& $g(\xi;\vectheta)=\Pr(\xi;\vectheta)$
& $g(\xi;\vectheta)=$ \\
&
& $\quad\quad\quad\nabla\Pr(\xi;\vectheta)$\\
GP factor ($f$)
& $f(\xi;\vectheta)=\bar{\rho}(\xi)\nabla\log\Pr(\xi;\vectheta)$
& $f(\xi)=\bar{\rho}(\xi)$ \\
Measurement ($y$)
& $y(\xi;\vectheta) = \rho(\xi)\nabla\log\Pr(\xi;\vectheta)$
& $y(\xi) = \rho(\xi)$\\
Prior mean of $f$
& $\exptE\big[f_j(\xi;\vectheta)\big] = 0$
& $\exptE\big[f(\xi)\big] = 0$ \\
Prior Cov. of $f$
&
$\Cov\big[f_j(\xi;\vectheta),f_\ell(\xi';\vectheta)\big] $
& $\Cov\big[f(\xi),f(\xi')\big] $\\
& $ \qquad\quad\quad\quad\quad =  \delta_{j,\ell} \, k(\xi,\xi')$
& $\quad\quad\quad\;= k(\xi,\xi') $\\
Kernel function
& $k(\xi,\xi') = $
& $k(\xi,\xi') = $\\
& $\quad\quad\big( 1+\vecu(\xi)^\top \matG^{-1} \vecu(\xi') \big)^2$
& $\quad\vecu(\xi)^\top \matG^{-1} \vecu(\xi')$\\
$\exptE\big[\nabla\eta_B(\vectheta)|\D_M\big]$
& $\matY\matC\vecb$
& $\matB\matC\vecy$\\
$\Cov\big[\nabla\eta_B(\vectheta)|\D_M\big]$
& $(b_0-\vecb^\top \matC \vecb) \matI$
& $\matB_0 - \matB \matC \matB^\top$\\
$\vecb$ or $\matB$
& $(\vecb)_i=1+\vecu(\xi_i)^\top \matG^{-1} \vecu(\xi_i)$
& $\matB=\matU$  \\
$b_0$ or $\matB_0$
& $b_0 = 1+n$
& $\matB_0 = \matG$\\
\hline
\end{tabular}
\end{center}
\caption{Summary of the Bayesian policy gradient Models 1 and 2.}
\label{tab:models}
\end{table}

In {\bf Model~1}, a vector-valued GP prior is placed over $f(\xi;\vectheta)$. This induces a GP prior over the corresponding noisy measurement $y(\xi;\vectheta)$. It is assumed that each component of $f(\xi;\vectheta)$ may be evaluated independently of all other components, and thus, the same kernel function $\matK$ and noise covariance $\matSigma$ are used for all components of $f(\xi;\vectheta)$. Hence for the $j$th component of $f$ and $y$ we have a priori
\begin{align*}
\vecf_j &= \big( f_j(\xi_1;\vectheta),\ldots,f_j(\xi_M;\vectheta) \big)^\top \sim \N(\vec0,\matK), \\
\vecy_j &= \big( y_j(\xi_1;\vectheta),\ldots,y_j(\xi_M;\vectheta) \big)^\top \sim \N(\vec0,\matK+\matSigma).
\end{align*}
This vector-valued GP model gives us the posterior mean and covariance of $\nabla\eta_B(\vectheta)$ reported in Table~\ref{tab:models}m in which $\matY=\big[\vecy_1^\top;\ldots;\vecy_n^\top\big]$, $\matC=(\matK+\matSigma)^{-1}$, $n$ is the number of policy parameters, and $\matG(\vectheta)$ is the Fisher information matrix of policy $\vectheta$ defined as\footnote{To simplify notation, we omit  $\matG$'s dependence on the policy parameters $\vectheta$, and denote $\matG(\vectheta)$ as $\matG$ in the rest of this section.}
\begin{equation}
\label{Fisher:Matrix}
\matG(\vectheta)=\exptE\big[\vecu(\xi)\vecu(\xi)^\top\big]=\int\vecu(\xi)\vecu(\xi)^\top\Pr(\xi;\vectheta)d\xi,
\end{equation}
where $\vecu(\xi)$ is the score function of trajectory $\xi$ defined by Eq.~\ref{eq:score}. Note that the choice of the kernel $k$ allows us to derive closed form expressions for $\vecb$ and $b_0$, and, as a result $k$ is the quadratic Fisher kernel for the posterior moments of the gradient (see Table~\ref{tab:models}).

In {\bf Model~2}, $g$ is a vector-valued function and $f$ is a scalar valued GP representing the expected return of the path given as its argument. The noisy measurement corresponding to $f(\xi_i)$ is $y(\xi_i)=\rho(\xi_i)$, namely, the actual return accrued while following the path $\xi_i$. This GP model gives us the posterior mean and covariance of $\nabla\eta_B(\vectheta)$ reported in Table~\ref{tab:models} in which $\vecy=\big(\rho(\xi_1),\ldots,\rho(\xi_M)\big)^\top$ and $\matU=\big[\vecu(\xi_1),\ldots,\vecu(\xi_M)\big]$. Here the choice of kernel $k$ allows us to derive closed-form expressions for $\matB$ and $\matB_0$, and as a result, $k$ is again the Fisher kernel for the posterior moments of the gradient (see Table~\ref{tab:models}).

Note that the choice of Fisher-type kernels is motivated by the notion that a good representation should depend on the process generating the data (see~\cite{Jaakkola98EG,ShaweTaylor04KM}, for a thorough discussion). The particular selection of linear and quadratic Fisher kernels is guided by the desideratum that the posterior moments of the gradient be analytically tractable as discussed at the end of \S\ref{subsubsec:BQ}.

The above two BPG models can define Bayesian algorithms for evaluating the gradient of the expected return with respect to the policy parameters (see~\citealt{Ghavamzadeh06BP} and~\citealt{Ghavamzadeh12BP-TECH} for the pseudo-code of the resulting algorithms). It is important to note that computing the quadratic and linear Fisher kernels used in Models~1 and~2 requires calculating the Fisher information matrix $\matG(\vectheta)$ (Eq.~\ref{Fisher:Matrix}). Consequently, every time the policy parameters are updated, $\matG$ needs to be recomputed.
Ghavamzadeh and Engel~\citet{Ghavamzadeh06BP} suggest two possible approaches for online estimation of the Fisher information matrix.

Similar to most non-parametric methods, the Bayesian gradient evaluation algorithms can be made more efficient, both in time and memory, by {\em sparsifying} the solution. Sparsification also helps to numerically stabilize the algorithms when the kernel matrix is singular, or nearly so. Ghavamzadeh et al.~\citet{Ghavamzadeh12BP-TECH} show how one can incrementally perform such sparsification in their Bayesian gradient evaluation algorithms, i.e.,~how to selectively add a new observed path to a set of {\em dictionary} paths used as a basis for representing or approximating the full solution.
\end{paragraph}


\begin{paragraph}{Policy Improvement:} Ghavamzadeh and Engel~\citet{Ghavamzadeh06BP} also show how their Bayesian algorithms for estimating the gradient can be used to define a Bayesian policy gradient algorithm. The algorithm starts with an initial set of policy parameters $\vectheta_0$, and at each iteration $j$, updates the parameters in the direction of the posterior mean of the gradient of the expected return $\exptE\big[\nabla\eta_B(\vectheta_j)|\D_M\big]$ estimated by their Bayesian gradient evaluation algorithms. This is repeated $N$ times, or alternatively, until the gradient estimate is sufficiently close to zero. Since the Fisher information matrix, $\matG$, and the posterior distribution (both mean and covariance) of the gradient of the expected return are estimated at each iteration of this algorithm, we may make the following modifications in the resulting Bayesian policy gradient algorithm at little extra cost:
\begin{itemize}
\item Update the policy parameters in the direction of the natural gradient, $\matG(\vectheta)^{-1}\exptE\big[\nabla\eta_B(\vectheta)|\D_M\big]$, instead of the regular gradient, $\exptE\big[\nabla\eta_B(\vectheta)|\D_M\big]$.
\item Use the posterior covariance of the gradient of the expected return as a measure of the uncertainty in the gradient
estimate, and thus, as a measure to tune the step-size parameter in the gradient update (the larger the posterior variance the smaller the step-size) (see the experiments in~\cite{Ghavamzadeh12BP-TECH} for more details). In a similar approach, Vien et al.~\citet{Vien11HM} used BQ to estimate the Hessian matrix distribution and then used its mean as learning rate schedule to improve the performance of BPG. They empirically showed that their method performs better than BPG and BPG with natural gradient in terms of convergence speed.
\end{itemize}

It is important to note that similar to the gradient estimated by the GPOMDP algorithm of Baxter and Bartlett~\citet{Baxter01IP}, the gradient estimated by these algorithms, $\exptE\big[\nabla\eta_B(\vectheta)|\D_M\big]$, can be used with the conjugate-gradient and line-search methods of Baxter and Bartlett~\citet{Baxter01EI} for improved use of gradient information. This allows us to exploit the information contained in the gradient estimate more aggressively than by simply adjusting the parameters by a small amount in the direction of $\exptE\big[\nabla\eta_B(\vectheta)|\D_M\big]$.

The experiments reported in~\citet{Ghavamzadeh12BP-TECH} indicate that the BPG algorithm tends to significantly reduce the number of samples needed to obtain accurate gradient estimates. Thus, given a fixed number of samples per iteration, finds a better policy than MC-based policy gradient methods. These results are in line with previous work on BQ, for example a work by Rasmussen and Ghahramani~\citet{Rasmussen03BM} that demonstrates how BQ, when applied to the evaluation of an expectation, can outperform MC estimation by orders of magnitude in terms of the mean-squared error.
\end{paragraph}

\begin{paragraph}{Extension to Partially Observable Markov Decision Processes:}
The above models and algorithms can be easily extended to partially observable problems without any changes using similar techniques as in Section~6 of~\citet{Baxter01IP}. This is due to the fact that the BPG framework considers complete system trajectories as its basic observable unit, and thus, does not require the dynamic within each trajectory to be of any special form. This generality has the downside that it cannot take advantage of the Markov property when the system is Markovian (see~\cite{Ghavamzadeh12BP-TECH} for more details). To address this issue, Ghavamzadeh and Engel~\citet{Ghavamzadeh07BA} then extended their BPG framework to actor-critic algorithms and present a new Bayesian take on the actor-critic architecture, which is the subject of the next section.
\end{paragraph}


\section{Bayesian Actor-Critic}
\label{ss:ac-alg}

Another approach to reduce the variance of the policy gradient estimates is to use an explicit representation for the value function of the policy. This class of PG algorithms are called {\em actor-critic} and they were among the earliest to be investigated in RL~\citep{Barto83NE,Sutton84TC}.
%
%
Unlike in \S\ref{ss:BPG} where we consider complete trajectories as the basic observable unit, in this section we assume that the system is Markovian, and thus, the basic observable unit is one step system transition (state, action, next state). It can be shown that under certain regularity conditions~\citep{Sutton00PG}, the expected return of policy $\mu$ may be written in terms of state-action pairs (instead of in terms of trajectories as in Eq.~\ref{eq:pol-expt-ret1}) as
\begin{equation}
\eta(\mu) = \int_{\Z} dz \pi^\mu(z) \bar{r}(z),
\label{eq:exp-ret2}
\end{equation}
where $z=(s,a)$ is a state-action pair and $\pi^\mu(z)=\sum_{t=0}^\infty \gamma^t {P_t^\mu}(z)$ is a discounted weighting of state-action pairs encountered while following policy $\mu$. In the definition of $\pi^\mu$, the term $P_t^\mu(z_t)$ is the $t$-step state-action occupancy density of policy $\mu$ given by
\begin{equation*}
P_t^\mu(z_t)=\int_{\Z^t} dz_0 \ldots dz_{t-1} P_0^\mu(z_0) \prod_{i=1}^{t}P^\mu(z_{i}|z_{i-1}).
\end{equation*}
Integrating $a$ out of $\pi^\mu(z)=\pi^\mu(s,a)$ results in the corresponding discounted weighting of states encountered by following policy $\mu$: $\nu^\mu(s)=\int_{\A}da\pi^\mu(s,a)$. Unlike $\nu^\mu$ and $\pi^\mu$, $(1-\gamma)\nu^\mu$ and $(1-\gamma)\pi^\mu$ are distributions and are analogous to the stationary distributions over states and state-action pairs of policy $\mu$ in the undiscounted setting, respectively.


The policy gradient theorem (\cite[Proposition~1]{Marbach98SM}; \cite[Theorem~1]{Sutton00PG}; \cite[Theorem~1]{Konda00AA}) states that the gradient of the expected return, defined by Eq.~\ref{eq:exp-ret2}, is given by 
\begin{equation}
\nabla\eta(\vectheta)=\int dsda\;\nu(s;\vectheta)\nabla\mu(a|s;\vectheta)Q(s,a;\vectheta).
\label{eq:policy_grad}
\end{equation}
%
%
%
%
%

In an AC algorithm, the actor updates the policy parameters $\vectheta$ along the direction of an estimate of the gradient of the performance measure (Eq.~\ref{eq:policy_grad}), while the critic helps the actor in this update by providing it with an estimate of the action-value function $Q(s,a;\vectheta)$. In most existing AC algorithms (both conventional and natural), the actor uses Monte-Carlo (MC) techniques to estimate the gradient of the performance measure and the critic approximates the action-value function using some form of temporal difference (TD) learning~\citep{Sutton88LP}.

The idea of Bayesian actor-critic (BAC)~\citep{Ghavamzadeh07BA} is to apply the Bayesian quadrature (BQ) machinery described in \S\ref{subsubsec:BQ} to the policy gradient expression given by Eq.~\ref{eq:policy_grad} in order to reduce the variance in the gradient estimation procedure (see Figure~\ref{fig:BQ-BPG-BAC} for the connection between the BQ machinery and BAC).

Similar to the BPG methods described in \S\ref{subsubsec:BPG-algo}, in BAC, we place a Gaussian process (GP) prior over action-value functions using a prior covariance kernel defined on state-action pairs: $k(z,z')=\Cov\big[Q(z),Q(z')\big]$. We then compute the GP posterior conditioned on the sequence of individual observed transitions. By an appropriate choice of a prior on action-value functions, we are able to derive closed-form expressions for the posterior moments of $\nabla\eta(\vectheta)$. The main questions here are: {\bf 1)} how to compute the GP posterior of the action-value function given a sequence of observed transitions? and {\bf 2)} how to choose a prior for the action-value function that allows us to derive closed-form expressions for the posterior moments of $\nabla\eta(\vectheta)$? The Gaussian process temporal difference (GPTD) method~\citep{Engel05RL} described in \S\ref{sss:GPTD} provides a machinery for computing the posterior moments of $Q(z)$. After $t$ time-steps, GPTD gives us the following posterior moments for $Q$ (see Eqs.~\ref{eq:post-nonparam1} and~\ref{eq:post-nonparam2} in \S\ref{sss:GPTD}):
\begin{align*}
\widehat{Q}_t(z) &= \exptE\left[ Q(z)|\D_t \right] = \veck_t(z)^\top \vecalpha_t ,\nonumber\\
\widehat{S}_t(z,z') &= \Cov \left[Q(z),Q(z')|\D_t \right]= k(z,z')-\veck_t(z)^\top \matC_t \veck_t(z'),
\end{align*}
where $\D_t$ denotes the observed data up to and including time step $t$, and
\begin{align}
\label{eq:alpha_C}
&\veck_t(z)=\big( k(z_0,z),\ldots, k(z_t,z)\big)^\top,\quad
\matK_t =\big[\veck_t(z_0), \veck_t(z_1), \ldots, \veck_t(z_t)\big],\nonumber\\
&\vecalpha_t=\matH_t^\top \!\left(\matH_t\matK_t\matH_t^\top\!+\matSigma_t \right)^{-1}\vecr_{t-1}\;,\quad
\matC_t =\matH_t^\top \!\left(\matH_t \matK_t\matH_t^\top\!+\matSigma_t\right)^{-1}\!\!\matH_t\;.
\end{align}
Using the above equations for the posterior moments of $Q$, making use of the linearity of Eq.~\ref{eq:policy_grad} in $Q$, and denoting $g(z;\vectheta)=\pi^\mu(z)\nabla\log\mu(a|s;\vectheta)$, the posterior moments of the policy gradient $\nabla\eta(\vectheta)$ may be written as~\citep{Ohagan91BQ}

\vspace{-0.15in}
\begin{small}
\begin{align*}
\exptE\big[\nabla\eta(\vectheta)|\D_t\big]&=\int_{\Z} dzg(z;\vectheta)\veck_t(z)^\top \vecalpha_t ,\nonumber \\
\Cov \big[\nabla\eta(\vectheta)|\D_t \big]&=\int_{\Z^2}dz dz'g(z;\vectheta)\left(k(z,z')-\veck_t(z)^\top\matC_t\veck_t(z')\right)g(z';\vectheta)^\top.
\end{align*}
\end{small}

\vspace{-0.15in}
\noindent
These equations provide us with the general form of the posterior policy gradient moments. We are now left with a computational issue, namely, how to compute the integrals appearing in these expressions? We need to be able to evaluate the following integrals:
\begin{equation} \label{eq:UV_def}
\matB_t=\int_{\Z} dzg(z;\vectheta)\veck_t(z)^\top,\hspace{0.4in}\matB_0=\int_{\Z^2} dz dz' g(z;\vectheta)k(z,z')g(z';\vectheta)^\top.
\end{equation}
Using the definitions of $\matB_t$ and $\matB_0$, the gradient posterior moments may be written as
\begin{equation} \label{eq:post_compact}
\exptE\big[\nabla\eta(\vectheta)|\D_t\big]=\matB_t\vecalpha_t\;,\hspace{0.25in}\Cov\big[\nabla\eta(\vectheta)|\D_t \big]=\matB_0-\matB_t\matC_t\matB_t^\top.
\end{equation}

In order to render these integrals analytically tractable, Ghavamzadeh and Engel~\citet{Ghavamzadeh07BA} chose the prior covariance kernel to be the sum of an arbitrary state-kernel $k_s$ and the Fisher kernel $k_F$ between state-action pairs, i.e.,
\begin{equation}
k_F(z,z')=\vecu(z;\vectheta)^\top \matG(\vectheta)^{-1}\vecu(z')\;,\hspace{0.3in}k(z,z')=k_s(s,s') + k_F(z,z')\;,
\label{eq:kernels}
\end{equation}
where $\vecu(z;\vectheta)$ and $\matG(\vectheta)$ are respectively the score function and the Fisher information matrix, defined as\footnote{Similar to $\vecu(\xi)$ and $\matG$ defined by Eqs.~\ref{eq:score} and~\ref{Fisher:Matrix}, to simplify the notation, we omit the dependence of $\vecu$ and $\matG$ to the policy parameters $\vectheta$, and replace $\vecu(z;\vectheta)$ and $\matG(\vectheta)$ with $\vecu(z)$ and $\matG$ in the sequel.}
\begin{align}
\vecu(z;\vectheta)&=\nabla\log\mu(a|s;\vectheta), \nonumber \\
\label{eq:Fisher-state-action}
\matG(\vectheta)&=\exptE_{s\sim \nu^\mu,a\sim\mu}\left[\nabla\log\mu(a|s;\vectheta)\nabla\log\mu(a|s;\vectheta)^\top\right]\\
&=\exptE_{z\sim\pi^\mu}\left[\vecu(z;\vectheta)\vecu(z;\vectheta)^\top\right].\nonumber
\end{align}
Using the prior covariance kernel of Eq.~\ref{eq:kernels}, Ghavamzadeh and Engel~\citet{Ghavamzadeh07BA} showed that the integrals in Eq.~\ref{eq:UV_def} can be computed as
\begin{equation} \label{eq:UV}
\matB_t=\matU_t\;,\hspace{1.0in}\matB_0=\matG,
\end{equation}
where $\matU_t=\big[\vecu(z_0),\vecu(z_1),\ldots,\vecu(z_t)\big]$. As a result, the integrals of the gradient posterior moments (Eq.~\ref{eq:post_compact}) are analytically tractable (see~\cite{Ghavamzadeh12BP-TECH} for more details). An immediate consequence of Eq.~\ref{eq:UV} is that, in order to compute the posterior moments of the policy gradient, we only need to be able to evaluate (or estimate) the score vectors $\vecu(z_i),\;i=0,\ldots,t$ and the Fisher information matrix $\matG$ of the policy.

Similar to the BPG method, Ghavamzadeh and Engel~\citet{Ghavamzadeh07BA} suggest methods for online estimation of the Fisher information matrix in Eq.~\ref{eq:Fisher-state-action} and for using online sparsification to make the BAC algorithm more efficient in both time and memory (see~\cite{Ghavamzadeh12BP-TECH} for more details). They also report experimental results~\citep{Ghavamzadeh07BA,Ghavamzadeh12BP-TECH} which indicate that the BAC algorithm tends to significantly reduce the number of samples needed to obtain accurate gradient estimates, and thus, given a fixed number of samples per iteration, finds a better policy than both MC-based policy gradient methods and the BPG algorithm, which do not take into account the Markovian property of the system.

\chapter{Risk-aware Bayesian Reinforcement Learning}
\label{s:risk}

The results presented so far have all been concerned with optimizing the \emph{expected return} of the policy. However, in many applications, the decision-maker is also interested in minimizing the \emph{risk} of a policy. By risk,\footnote{The term risk here should not be confused with the Bayes risk, defined in Chapter \ref{s:Bayesian-bandits}.} we mean performance criteria that take into account not only the expected return, but also some additional statistics of it, such as variance, Value-at-Risk (VaR), expected shortfall (also known as conditional-value-at-risk or CVaR), etc. The primary motivation for dealing with risk-sensitive performance criteria comes from finance, where risk plays a dominant role in decision-making. However, there are many other application domains where risk is important, such as process control, resource allocation, and clinical decision-making.

In general, there are two sources that contribute to the reward uncertainty in MDPs: {\em internal uncertainty} and {\em parametric uncertainty}. Internal uncertainty reflects the uncertainty of the return due to the stochastic transitions and rewards, for a single and \emph{known} MDP. Parametric uncertainty, on the other hand, reflects the uncertainty about the unknown MDP parameters -- the transition and reward \emph{distributions}. As a concrete example of this dichotomy consider the following two experiments. First, select a \emph{single} MDP and execute some fixed policy on it several times. In each execution, the return may be different due to the stochastic transitions and reward. This variability corresponds to the internal uncertainty. In the second experiment, consider drawing several MDPs from some distribution (typically, the posterior distribution in a Bayesian setting). For each drawn MDP, execute the same policy several times and compute the average return across the executions. The variability in the \emph{average} returns across the different MDPs corresponds to the parametric type of uncertainty. The Bayesian setting offers a framework for dealing with parameter uncertainty in a principled manner. Therefore, work on risk-sensitive RL in the Bayesian setting focuses on risk due to parametric uncertainty, as we shall now survey.


\subsection*{Bias and Variance Approximation in Value Function Estimates}

The first result we discuss concerns policy evaluation. Mannor et al.~\citet{Mannor07BV} derive approximations to the bias and variance of the value function of a fixed policy due to parametric uncertainty.

Consider an MDP $\M$ with \emph{unknown} transition probabilities $P$,\footnote{Following the framework of Chapter~\ref{s:model-based-brl}, we assume that the rewards are known and only the transitions are unknown. The following results may be extended to cases where the rewards are also unknown, as outlined in \S\ref{ss:ext-rewards}.} a Dirichlet prior on the transitions, and assume that we have observed $n(s,a,s')$ transitions from state $s$ to state $s'$ under action $a$. Recall that the posterior transition probabilities are also Dirichlet, and may be calculated as outlined in \S\ref{sss:Dir}.\footnote{Note that this is similar to the BAMDP formulation of \S\ref{ss:belief-mdp}, where the hyper-state encodes the posterior probabilities of the transitions given the observations.} Consider also a stationary Markov policy $\mu$, and let $P^\mu$ denote the unknown transition probabilities induced by $\mu$ in the unknown MDP $\M$ and $V^\mu$ denote the corresponding value function. Recall that $V^\mu$ is the solution to the Bellman equation~\eqref{eq:Bellman-eq} and may be written explicitly as $V^\mu = \left( I - \gamma P^\mu\right)^{-1} R^\mu$, where we recall that $R^\mu$ denotes the expected rewards induced by $\mu$. The Bayesian formalism allows the calculation of the \emph{posterior} mean and covariance of $V^\mu$, given the observations.

Let $\hat{P}\left( s' | s,a\right)=\exptE_{\text{post}}\big[ P\left( s' | s,a\right)\big]$ denote the expected transition probabilities with respect to their posterior distribution, and let $\hat{P}^\mu\left( s' | s\right)=\sum_a \mu(a|s)\hat{P}\left( s' | s,a\right)$ denote the expected transitions induced by $\mu$. We denote by
$
\hat{V}^\mu = ( I - \gamma \hat{P}^\mu)^{-1} R^\mu
$
the \emph{estimated} value function. Also, let $\mathbf{e}$ denote a vector of ones. The posterior mean and covariance of $V^\mu$ are given in the following two theorems:

\begin{theorem} \citep{Mannor07BV}\label{thm:biasVar-1}
The expectation (under the posterior) of $V^\mu$ satisfies
\begin{equation*}
\exptE_{\text{post}} \left[ V^\mu \right] = \hat{V}^\mu + \gamma^2 \hat{X} \hat{Q} \hat{V}^\mu + \gamma \hat{B} + L_{\text{bias}},
\end{equation*}
where $\hat{X} = \left( I - \gamma \hat{P}^\mu\right)^{-1}$, and the vector $\hat{B}$ and matrix $\hat{Q}$ are computed according to
\begin{equation*}
\hat{B}_i = \sum_a \mu(a|i)^2 R(i,a) \mathbf{e}^\top \hat{M}^{i,a} \hat{X}_{\cdot,i},
\end{equation*}
and
\begin{equation*}
\hat{Q}_{i,j} = \hat{\Cov}^{(i)}_{j,\cdot} \hat{X}_{\cdot,i} \quad\quad \text{in which} \quad\quad \hat{\Cov}^{(i)} = \sum_a \mu(a|i)^2 \hat{M}^{i,a},
\end{equation*}
where the matrix $\hat{M}^{i,a}$ is the posterior covariance matrix of $P\left( \cdot | s,a\right)$, and higher order terms
\begin{equation*}
L_{\text{bias}} = \sum_{k=3}^{\infty} \gamma^k \exptE \left[ f_k(\tilde{P})\right] R^\mu,
\end{equation*}
in which $\tilde{P} = P - \hat{P}$, and $f_k(\tilde{P}) = \hat{X}\left( \tilde{P} \hat{X}\right)^k$.
\end{theorem}

\begin{theorem} \citep{Mannor07BV}\label{thm:biasVar-2}
Using the same notations as Theorem~\ref{thm:biasVar-1}, the second moment (under the posterior) of $V^\mu$ satisfies

\begin{small}
\begin{equation*}
\begin{split}
\exptE_{\text{post}} \big[ V^\mu {V^\mu}^\top \big] &= \hat{V}^\mu \hat{V}^{\mu^\top} + \hat{X} \Big( \gamma^2 \big(\hat{Q} \hat{V}^\mu {R^\mu}^\top + R^\mu \hat{V}^{\mu^\top} \hat{Q}^\top \big) \\
&+ \gamma\big( \hat{B} {R^\mu}^\top + R^\mu\hat{B}^\top\big) + \hat{W}\Big) \hat{X}^\top + L_{\text{var}},
\end{split}
\end{equation*}
\end{small}
where $\hat{W}$ is a diagonal matrix with elements
\begin{equation*}
\hat{W}_{i,i} = \sum_a \mu(a|i)^2 \big(\gamma V^{\mu^\top} + R(i,a) \mathbf{e}\big) \hat{M}^{i,a} \big(\gamma V^\mu + R(i,a) \mathbf{e}^{\top}\big),
\end{equation*}
and higher order terms
\begin{equation*}
L_{\text{var}} = \sum_{k,l:k+l>2} \gamma^{k+l} \exptE \left[ f_k(\tilde{P}) R^\mu {R^\mu}^\top f_l(\tilde{P})^\top \right].
\end{equation*}
\end{theorem}

It is important to note that except for the higher order terms $L_{\text{bias}}$ and $L_{\text{var}}$, the terms in Theorems~\ref{thm:biasVar-1} and~\ref{thm:biasVar-2} do not depend on the unknown transitions, and thus, may be calculated \emph{from the data}. This results in a second-order approximation of the bias and variance of $V^\mu$, which may be used to derive confidence intervals around the estimated value function. This approximation was also used by Delage and Mannor~\citet{Delage10PO} for risk-sensitive policy optimization, as we describe next.


\subsection*{Percentile Criterion}

Consider again a setting where $n(s,a,s')$ transitions from state $s$ to state $s'$ under action $a$ from MDP $\M$ were observed, and let $P_{\text{post}}$ denote the posterior distribution of the transition probabilities in $\M$ given these observations. Delage and Mannor~\citet{Delage10PO} investigate  the \emph{percentile criterion}\footnote{This is similar to the popular financial risk measure Value-at-Risk (VaR). However, note that VaR is typically used in the context of \emph{internal} uncertainty.} for $\M$, defined as
\begin{equation}
\label{eq:perc_crit}
\begin{split}
\max_{y\in\Re, \mu} & \quad y \\
\text{s.t.} & \quad P_{\text{post}} \left( \exptE\left[\left.\sum_{t=0}^\infty\gamma^tR(Z_t)\;\right| Z_0\sim P_0, \mu \right] \geq y\right) \geq 1 - \epsilon,
\end{split}
\end{equation}
where $P_{\text{post}}$ denotes the probability of drawing a transition matrix $P'$ from the posterior distribution of the transitions, and the expectation is with respect to a concrete realization of that $P'$. Note that the value of the optimization problem in \eqref{eq:perc_crit} is a $(1-\epsilon)$-guarantee to the performance of the optimal policy with respect to the parametric uncertainty.

Unfortunately, solving~\eqref{eq:perc_crit} for general uncertainty in the parameters is NP-hard \citep{Delage10PO}. However, for the case of a Dirichlet prior, the $2$nd order approximation of Theorem~\ref{thm:biasVar-1} may be used to derive an approximately optimal solution.

Let $F(\mu)$ denote the $2$nd order approximation\footnote{Note that the $1$st order term is ignored, since it would cancel anyway in the optimization that follows.} of the expected return under policy $\mu$ and initial state distribution $P_0$ (cf.~Theorem~\ref{thm:biasVar-1})
\begin{equation*}
F(\mu) = P_0^\top \hat{V}^\mu + \gamma^2 P_0^\top \hat{X} \hat{Q} \hat{V}^\mu.
\end{equation*}
The next result of~\citet{Delage10PO} shows that given enough observations, optimizing $F(\mu)$ leads to an approximate solution of the percentile problem~\eqref{eq:perc_crit}.

\begin{theorem} \citep{Delage10PO}\label{thm:percentile}
Let $N^* = \min_{s,a} \sum_{s'} n(s,a,s')$ denote the minimum number of state transitions observed from any state using any action, and $\epsilon\in (0,0.5]$. The policy
\begin{equation*}
\hat{\mu} = \argmax_\mu F(\mu)
\end{equation*}
is $\O \big( 1/ \sqrt{\epsilon N^*}\big)$ optimal with respect to the percentile optimization criterion~\eqref{eq:perc_crit}.
\end{theorem}

Note that, as discussed earlier, $F(\mu)$ may be efficiently evaluated for every $\mu$. However, $F(\mu)$ is non-convex in $\mu$, but empirically, global optimization techniques for maximizing $F(\mu)$ lead to useful solutions~\citep{Delage10PO}.

Delage and Mannor~\citep{Delage10PO} also consider a case where the state transitions are known, but there is uncertainty in the reward distribution. For general reward distributions the corresponding percentile optimization problem is also NP-hard. However, for the case of a Gaussian prior, the resulting optimization problem is a second-order cone program, for which efficient solutions are known.


\subsection*{Max-Min Criterion}

Consider the percentile criterion~\eqref{eq:perc_crit} in the limit of $\epsilon \to 0$. In this case, we are interested in the performance under the worst realizable posterior transition probability, i.e.,
\begin{equation}
\label{eq:minmax_crit}
\max_{\mu} \min_{P\in \P_{\text{post}} } \exptE\left[\left.\sum_{t=0}^\infty\gamma^tR(Z_t)\;\right| Z_0\sim P_0, P, \mu \right],
\end{equation}
where $\P_{\text{post}}$ denotes the set of all realizable transition probabilities in the posterior. For the case of a Dirichlet prior, this criterion is useless, as the set $\P_{\text{post}}$ contains the entire simplex for each state. Bertuccelli et al.~\citep{Bertuccelli12RA} consider instead a \emph{finite} subset\footnote{This is also known as a set of \emph{scenarios}.} $\hat{\P}_{\text{post}} \in \P_{\text{post}}$, and minimize only over the set $\hat{\P}_{\text{post}}$, resulting in the following criterion:
\begin{equation}
\label{eq:scenario_crit}
\max_{\mu} \min_{P\in \hat{\P}_{\text{post}} } \exptE\left[\left.\sum_{t=0}^\infty\gamma^tR(Z_t)\;\right| Z_0\sim P_0, P, \mu \right].
\end{equation}

Given a set $\hat{\P}_{\text{post}}$, the optimization in \eqref{eq:scenario_crit} may be solved efficiently using min-max dynamic programming. Thus, the `real question' is how to construct $\hat{\P}_{\text{post}}$. Naturally, the construction of $\hat{\P}_{\text{post}}$ should reflect the posterior distribution of transition probabilities. One approach is to construct it from a finite number of models sampled from the posterior distribution, reminiscent of Thompson sampling. However, as claimed in~\citep{Bertuccelli12RA}, this approach requires a very large number of samples in order to adequately represent the posterior.

Alternatively, Bertuccelli et al.~\citep{Bertuccelli12RA} propose a deterministic sampling procedure for the Dirichlet distribution based on sigma-points. In simple terms, sigma-points are points placed around the posterior mean and spaced proportionally to the posterior variance. The iterative algorithm of~\citet{Bertuccelli12RA} consists of two phases. In the first phase, the already observed transitions are used to derive the posterior Dirichlet distribution, for which an optimal policy is derived by solving \eqref{eq:scenario_crit}. In the second phase, this policy is then acted upon in the system to generate additional observations. As more data become available, the posterior variance decreases and the sigma points become closer, leading to convergence of the algorithm.


\subsection*{Percentile Measures Criteria}

The NP-hardness of the percentile criterion for general uncertainty structures motivates a search for more tractable risk-aware performance criteria. Chen and Bowling~\citep{Chen2012TO} propose to replace the individual percentile with a \emph{measure} over percentiles. Formally, given a measure $\psi$ over the interval $[0,1]$, consider the following optimization problem:
\begin{equation}
\label{eq:measure_perc_crit}
\begin{split}
\max_{\mu, y\in \F } & \quad \int_x y(x) d\psi \\
\text{s.t.} & \quad P_{\text{post}} \left( \exptE\left[\left.\sum_{t=0}^T R(Z_t)\;\right| Z_0\sim P_0, \mu \right] \geq y(x)\right) \geq x \quad\quad \forall x \in [0,1],
\end{split}
\end{equation}
where $\F$ is the class of real-valued and bounded $\psi$-integrable functions on the interval $[0,1]$, $P_{\text{post}}$ denotes the probability of drawing a transition matrix $P'$ from the posterior distribution of the transitions (as before), and the expectation is with respect to a concrete realization of that $P'$. Note that here the horizon is $T$ and there is no discounting, as opposed to the infinite horizon discounted setting discussed earlier.

The percentile measure criterion~\eqref{eq:measure_perc_crit} may be seen as a generalization of the percentile criterion~\eqref{eq:perc_crit}, which is obtained by setting $\psi$ to a Dirac delta at $1-\epsilon$. In addition, when $\psi$ is uniform on $[0,1]$,~\eqref{eq:measure_perc_crit} is equivalent to the expected value of Theorem~\ref{thm:biasVar-1}, and when $\psi$ is a delta Dirac at $0$, we obtain the max-min criterion~\eqref{eq:minmax_crit}. Finally, when $\psi$ is a step function, an expected shortfall (CVaR) criterion is obtained.

Chen and Bowling~\citep{Chen2012TO} introduce the $k$-of-$N$ family of percentile measures, which admits an efficient solution under a general uncertainty structure. For a policy $\mu$, the $k$-of-$N$ measure is equivalent to the following sampling procedure: first draw $N$ MDPs according to the posterior distribution, then select a set of the $k$ MDPs with the worst expected performance under policy $\mu$, and finally choose a random MDP from this set (according to a uniform distribution). By selecting suitable $k$ and $N$, the $k$-of-$N$ measure may be tuned to closely approximate the CVaR or max-min criterion.

The main reason for using the $k$-of-$N$ measure is that the above sampling procedure may be seen as a two-player zero-sum extensive-form game with imperfect information, which may be solved efficiently using counterfactual regret minimization. This results in the following convergence guarantee:

\begin{theorem} \citep{Chen2012TO}\label{thm:k-of-n_guarantee}
For any $\epsilon > 0$ and $\delta \in (0,1]$, let
\begin{equation*}
\bar{T} = \left( 1 + \frac{2}{\sqrt{\delta}}\right)^2 \frac{16T^2 \Delta^2 |\I_1|^2 |\A|}{\delta^2 \epsilon^2},
\end{equation*}
where $T \Delta$ is the maximum difference in total reward over $T$ steps. With probability $1-\delta$, the current strategy at iteration $T^*$, chosen uniformly at random from the interval $[1,\bar{T}]$, is an $\epsilon$-approximation to a solution of~\eqref{eq:measure_perc_crit} when $\psi$ is a $k$-of-$N$ measure. The total time complexity is $\O\left((T \Delta / \epsilon)^2 \frac{|\I_1|^3 |\A|^3 N \log N}{\delta^3} \right)$, where $|\I_1| \in \O \left( |\Scal| T\right)$ for arbitrary reward uncertainty and $|\I_1| \in \O \left( |\Scal|^{T+1} \A^T\right)$ for arbitrary transition and reward uncertainty.
\end{theorem}

The exponential dependence on the horizon $T$ in Theorem~\ref{thm:k-of-n_guarantee} is due to the fact that an optimal policy for the risk-sensitive criterion~\eqref{eq:measure_perc_crit} is not necessarily Markov and may depend on the complete history. In comparison, the previous results avoided this complication by searching only in  the space of Markov policies. An interesting question is whether other choices of measure $\psi$ admit an efficient solution. Chen and Bowling~\citep{Chen2012TO} provide the following sufficient condition for tractability:
\begin{theorem} \citep{Chen2012TO}\label{thm:perc_measure_tract}
Let $\psi$ be an absolutely continuous measure with density function $g_\psi$, such that $g_\psi$ is non-increasing and piecewise Lipschitz continuous with $m$ pieces and Lipschitz constant $L$. A solution of \eqref{eq:measure_perc_crit} can be approximated with high probability in time polynomial in $\left\{|\A|, |\Scal|,\Delta,L, m, \frac{1}{\epsilon},\frac{1}{\delta}\right\}$ for
(i) arbitrary reward uncertainty with time also polynomial in the horizon or (ii) arbitrary transition
and reward uncertainty with a fixed horizon.
\end{theorem}

Note that in line with the previous hardness results, both the CVaR and max-min criteria may be represented using a non-increasing and piecewise Lipschitz continuous measure, while the percentile criterion may not.


\chapter{BRL Extensions}
\label{s:misc}
\vspace{-20pt}
In this section, we discuss extensions of the Bayesian reinforcement learning (BRL) framework to the following classes of problems: PAC-Bayes model selection, inverse RL, multi-agent RL, and multi-task RL.


\section{PAC-Bayes Model Selection}
\label{s:pac-bayes}

While Bayesian RL provides a rich framework for incorporating domain knowledge, one of the often mentioned limitations is the requirement to have \emph{correct} priors, meaning that the prior has to admit the true posterior. Of course this issue is pervasive across Bayesian learning methods, not just Bayesian RL.  Recent work on PAC-Bayesian analysis seeks to provide tools that are robust to poorly selected priors. PAC-Bayesian methods provide a way to simultaneously exploit prior knowledge when it is appropriate, while providing distribution-free guarantees based on properties such as VC dimension~\citep{mcallester99}.

PAC-Bayesian bounds for RL in finite state spaces were introduced by~\citep{fard10}, showing that it is possible to remove the assumption on the correctness of the prior, and instead, measure the consistency of the prior over the training data. Bounds are available for both model-based RL, where a prior distribution is given on the space of possible models, and model-free RL, where a prior is given on the space of value functions.  In both cases, the bound depends on an empirical estimate and a measure of distance between the stochastic policy and the one imposed by the prior distribution.  The primary use of these bounds is for model selection, where the bounds can be useful in choosing between following the empirical estimate and the Bayesian posterior, depending on whether the prior was informative or misleading. PAC-Bayesian bounds for the case of RL with function approximation are also available to handle problems with continuous state spaces~\citep{fard11}.

For the most part, PAC-Bayesian analysis to date has been primarily theoretical, with few empirical results. However, recent theoretical results on PAC-Bayesian analysis of the bandit case may provide useful tools for further development of this area~\citep{seldin11,seldin11b}.


\section{Bayesian Inverse Reinforcement Learning}
\label{ss:inverse-rl}

Inverse reinforcement learning (IRL) is the problem of learning the underlying model of the decision-making agent (expert) from its observed behavior and the dynamics of the system~\citep{Russell98LA}. IRL is motivated by situations in which the goal is only to learn the reward function (as in preference elicitation) and by problems in which the main objective is to learn good policies from the expert (apprenticeship learning). Both reward learning (direct) and apprenticeship learning (indirect) views of this problem have been studied in the last decade (e.g.,~\citealt{Ng00AI,Abbeel04AL,Ratliff06MM,Neu07AL,Ziebart08ME,Syed08GT}). What is important is that the IRL problem is inherently {\em ill-posed} since there might be an infinite number of reward functions for which the expert's policy is optimal. One of the main differences between the various works in this area is in how they formulate the reward preference in order to obtain a {\em unique} reward function for the expert.

The main idea of Bayesian IRL (BIRL) is to use a prior to encode the reward preference and to formulate the compatibility with the expert's policy as a likelihood in order to derive a probability distribution over the space of reward functions, from which the expert's reward function is somehow extracted. Ramachandran and
Amir~\citet{Ramachandran07BI} use this BIRL formulation and propose a Markov chain Monte Carlo (MCMC) algorithm to find the posterior mean of the reward function and return it as the reward of the expert. Michini and How~\citet{Michini12IE} improve the efficiency of the method in~\citet{Ramachandran07BI} by not including the entire state space in the BIRL inference. They use a kernel function that quantifies the similarity between states and scales down the BIRL inference by only including those states that are similar (the similarity is defined by the kernel function) to the ones encountered by the expert.
Choi and Kim \citet{Choi11MA} use the BIRL formulation of~\citet{Ramachandran07BI} and first show that using the posterior mean may not be a good idea since it may yield a reward function whose corresponding optimal policy is inconsistent with the expert's behaviour; the posterior mean integrates the error over the entire space of reward functions by including (possibly) infinitely many rewards that induce policies that are inconsistent with the expert's demonstration. Instead, they suggest that the maximum-a-posteriori (MAP) estimate could be a better solution for IRL. They formulate IRL as a posterior optimization problem and propose a gradient method to calculate the MAP estimate that is based on the (sub)differentiability of the posterior distribution. Finally, they show that most of the non-Bayesian IRL algorithms in the literature~\citep{Ng00AI,Ratliff06MM,Neu07AL,Ziebart08ME,Syed08GT} can be cast as searching for the MAP reward function in BIRL with different priors and different ways of encoding the compatibility with the expert's policy.

Using a single reward function to explain the expert's behavior might be problematic as its complexity grows with the complexity of the task being optimized by the expert. Searching for a complex reward function is often difficult. If many parameters are needed to model it, we are required to search over a large space of candidate functions. This problem becomes more severe when we take into account the testing of each candidate reward function, which requires solving an MDP for its optimal value function, whose computation usually scales poorly with the size of the state space.
Michini and How \citet{Michini12BN} suggest a potential solution to this problem in which they partition the observations (system trajectories generated by the expert) into sets of smaller sub-demonstrations, such that each sub-demonstration is attributed to a smaller and less-complex class of reward functions. These simple rewards can be intuitively interpreted as {\em sub-goals} of the expert. They propose a BIRL algorithm that uses a Bayesian non-parametric mixture model to automate this partitioning process. The proposed algorithm uses a Chinese restaurant process prior over partitions so that there is no need to specify the number of partitions a priori.

Most of the work in IRL assumes that the data is generated by a single expert optimizing a fixed reward function. However, there are many applications in which we observe multiple experts, each executing a policy that is optimal (or good) with respect to its own reward function. There have been a few works that consider this scenario. Dimitrakakis and Rothkopf~\citet{Dimitrakakis11BM} generalize BIRL to multi-task learning. They assume a common prior for the trajectories and estimate the reward functions individually for each trajectory. Other than the common prior, they do not make any additional efforts to group trajectories that are likely to be generated from the same or similar reward functions.
Babes et al.~\citet{Babes11AL} propose a method that combines expectation maximization (EM) clustering with IRL. They cluster the trajectories based on the inferred reward functions, where one reward function is defined per cluster. However, their proposed method is based on the assumption that the number of clusters (the number of the reward functions) is known. Finally,~ Choi and Kim \citet{Choi12NV} present a nonparametric Bayesian approach using the Dirichlet process mixture model in order to address the IRL problem with multiple reward functions. They develop an efficient Metropolis-Hastings sampler that utilizes the gradient of the reward function posterior to infer the reward functions from the behavior data. Moreover, after completing IRL on the behavior data, their method can efficiently estimate the reward function for a new trajectory by computing the mean of the reward function posterior, given the pre-learned results.


\section{Bayesian Multi-agent Reinforcement Learning}
\label{ss:multiagent}

There is a rich literature on the use of RL methods for multi-agent systems.
More recently, the extension of BRL methods to collaborative multi-agent systems has been proposed~\citep{chalkiadakis03}. The objective in this case is consistent with the standard BRL objective, namely to optimally balance the cost of exploring the world with the expected benefit of new information. However, when dealing with multi-agent systems, the complexity of the decision problem is increased in the following way:  while single-agent BRL requires maintaining a posterior over the MDP parameters (in the case of model-based methods) or over the value/policy (in the case of model-free methods), in multi-agent BRL, it is also necessary to keep a posterior over the policies of the other agents. This belief can be maintained in a tractable manner subject to certain structural assumptions on the domain, for example that the strategies of the agents are independent of each other. Alternatively, this framework can be used to control the formation of coalitions among agents~\citep{chalkiadakis10}. In this case, the Bayesian posterior can be limited to the uncertainty about the capabilities of the other agents, and the uncertainty about the effects of coalition actions among the agents. The solution to the Bayes-optimal strategy can be approximated using a number of techniques, including myopic or short ($1$-step) lookahead on the value function, or an extension of the value-of-information criteria proposed by~\citet{dearden98bayesian}.


\section{Bayesian Multi-Task Reinforcement Learning}
\label{ss:transfer}

Multi-task learning (MTL) is an important learning paradigm and active area of research in machine learning (e.g.,~\citealt{Caruana97ML,Baxter00MI}). A common setup in  MRL considers multiple related tasks for which we are interested in improving the performance of individual learning by sharing information across tasks. This transfer of information is particularly important when we are provided with only a limited number of data with which learn each task. Exploiting data from related problems provides more training samples for the learner and can improve the performance of the resulting solution. More formally, the main objective in MTL is to maximize the improvement over individual learning, averaged over multiple tasks. This should be distinguished from transfer learning in which the goal is to learn a suitable bias for a class of tasks in order to maximize the expected future performance.

Traditional RL algorithms typically do not directly take advantage of information coming from other similar tasks.  However recent work has shown that transfer and multi-task learning techniques can be employed in RL to reduce the number of samples needed to achieve nearly-optimal solutions. All
approaches to multi-task RL (MTRL) assume that the tasks share similarity in some components of the problem such as dynamics, reward structure, or value function. While some methods explicitly assume that the shared components are drawn from a common generative model~\citep{Wilson07MR,Mehta08TV,Lazaric10BM}, this assumption is more implicit in others. In~\citet{Mehta08TV}, tasks share the same dynamics and reward features, and only differ in the weights of the reward function. The proposed method initializes the value function for a new task using the previously learned value functions as a prior. Wilson et al.~\citet{Wilson07MR} and Lazaric and Ghavamzadeh \citet{Lazaric10BM} both assume that the distribution over some components of the tasks is drawn from a hierarchical Bayesian model (HBM). We describe these two methods in more detail below.

Lazaric and Ghavamzadeh~\citet{Lazaric10BM} study the MTRL scenario in which the learner is provided with a number of MDPs with common state and action spaces. For any given policy, only a small number of samples can be generated in each MDP, which may not be enough to accurately evaluate the policy. In such a MTRL problem, it is necessary to identify classes of tasks with similar structure and to learn them jointly. It is important to note that here a task is a pair of MDP and policy (i.e.,~a Markov chain) such that all the MDPs have the same state and action spaces. They consider a particular class of MTRL problems in which the tasks {\em share structure in their value functions}. To allow the value functions to share a common structure, it is assumed that they are all sampled from a common prior. They adopt the GPTD value function model (\citealt{Engel05RL}, also see \S\ref{sss:GPTD}) for each task, model the distribution over the value functions using an HBM, and develop solutions to the following problems: (i) joint learning of the value functions (multi-task learning), and (ii) efficient transfer of the information acquired in (i) to facilitate learning the value function of a newly observed task (transfer learning). They first present an HBM for the case in which all of the value functions belong to the same class, and derive an EM algorithm to find MAP estimates of the value functions and the model's hyper-parameters. However, if the functions do not belong to the same class, simply learning them together can be detrimental ({\em negative transfer}). It is therefore important to have models that will generally benefit from related tasks and will not hurt performance when the tasks are unrelated. This is particularly important in RL as changing the policy at each step of policy iteration (this is true even for fitted value iteration) can change the way in which tasks are clustered together. This means that even if we start with value functions that all belong to the same class, after one iteration the new value functions may be clustered into several classes. To address this issue, they introduce a Dirichlet process (DP) based HBM for the case where the value functions belong to an undefined number of classes, and derive inference algorithms for both the multi-task and transfer learning scenarios in this model.

The MTRL approach in~\citet{Wilson07MR} also uses a DP-based HBM to model the distribution over a common structure of the tasks. In this work, the tasks {\em share structure in their dynamics and reward functions}. The setting is incremental, i.e., the tasks are observed as a sequence, and there is no restriction on the number of samples generated by each task. The focus is not on joint learning with finite number of samples, but on using the information gained from the previous tasks to facilitate learning in a new one. In other words, the focus in this work is on transfer and not on multi-task learning.


\chapter{Outlook}
\label{s:outlook}

Bayesian reinforcement learning (BRL) offers a coherent probabilistic model for reinforcement learning.  It provides a principled framework to express the classic exploration-exploitation dilemma, by keeping an explicit representation of uncertainty, and selecting actions that are optimal with respect to a version of the problem that incorporates this uncertainty.  This framework is of course most useful for tackling problems where there is an explicit representation of prior information.  Throughout this survey, we have presented several frameworks for leveraging this information, either over the model parameters (model-based BRL) or over the solution (model-free BRL).

In spite of its elegance, BRL has not, to date, been widely applied. While there are a few successful applications of BRL for advertising~\citep{Graepel10WB,Tang13AA} and robotics~\citep{EngelSV05,ross08icra}, adoption of the Bayesian framework in applications is lagging behind the comprehensive theory that BRL has to offer. In the remainder of this section we analyze the perceived limitations of BRL and offer some research directions that might circumvent these limitations.

One perceived limitation of Bayesian RL is the need to provide a prior. While this is certainly the case for model-based BRL, for larger problems there is always a need for some sort of regularization. A prior serves as a mean to regularize the model selection problem. Thus, the problem of specifying a prior is addressed by {\em any} RL algorithm that is supposed to work for large-scale problems. We believe that a promising direction for future research concerns devising priors based on observed data, as per the empirical Bayes approach~\citep{efron10}. A related issue is model mis-specification and how to quantify the performance degradation that may arise from not knowing the model precisely.

There are also some algorithmic and theoretical challenges that we would like to point out here. First, scaling BRL is a major issue. Being able to solve large problems is still an elusive goal. We should mention that currently, a principled approach for scaling up RL in general is arguably missing. Approximate value function methods~\citep{Bertsekas96NP} have proved successful for solving certain large scale problems~\citep{powell2007approximate}, and policy search has been successfully applied to many robotics tasks~\citep{kober13}. However, there is currently no solution (neither frequentist nor Bayesian) to the exploration-exploitation dilemma in large-scale MDPs. We hope that scaling up BRL, in which exploration-exploitation is naturally handled, may help us overcome this barrier. Conceptually, BRL may be easier to scale since it allows us, in some sense, to embed domain knowledge into a problem. Second, the BRL framework we presented assumes that the model is specified correctly. Dealing with model misspecification and consequently with model selection is a thorny issue for all RL algorithms. When the state parameters are unknown or not observed, the dynamics may stop being Markov and many RL algorithms fail in theory and in practice. The BRL framework may offer a solution to this issue since the Bayesian approach can naturally handle model selection and misspecification by not committing to a single model and rather sustaining a posterior over the possible models.

Another perceived limitation of BRL is the complexity of \emph{implementing} the majority of BRL methods. Indeed, some frequentist RL algorithms are more elegant than their Bayesian counterparts, which may discourage some practitioners from using BRL. This is, by no means, a general rule. For example, Thompson sampling is  one of the most elegant approaches to the multi-armed bandit problem. Fortunately, the recent release of a software library for Bayesian RL algorithms promises to facilitate this~\citep{bbrl15,castronovo15}. We believe that Thompson sampling style algorithms (e.g., \citealt{russo2014learning,GopalanMannor15}) can pave the way to efficient algorithms in terms of both sample complexity and computational complexity. Such algorithms require, essentially, solving an MDP once in a while while taking nearly optimal exploration rate.

From the application perspective, we believe that BRL is still in its infancy. In spite of some early successes, especially for contextual bandit problems~\citep{Chapelle11EE}, most of the successes of large-scale real applications of RL are not in the Bayesian setting. We hope that this survey will help facilitate the research needed to make BRL a success in practice.
A considerable benefit of BRL is its ability to work well ``out-of-the-box": you only need to know relatively little about the uncertainty to perform well. Moreover, adding complicating factors such as long-term constraints or short-term safety requirements can be easily embedded in the framework.

From the modelling perspective, deep learning is becoming increasingly more popular as a building block in RL. It would be interesting to use probabilistic deep networks, such as restricted Boltzman machines as an essential ingredient in BRL. Probabilistic deep models can not only provide a powerful value function or policy approximator, but also allow using historical traces that come from a different problem (as in transfer learning). We believe that the combination of powerful models for value or policy approximation in combination with a BRL approach can further facilitate better exploration policies.

From the conceptual perspective, the main question that we see as open is how to  embed domain knowledge into the model? One approach is to modify the prior according to domain knowledge. Another is to consider parametrized or factored models. While these approaches may work well for particular domains, they require careful construction of the state space. However, in some cases, such a careful construction defeats the purpose of a fast and robust ``out-of-the-box" approach. We believe that approaches that use causality and non-linear dimensionality reduction may be the key to using BRL when a careful construction of a model is not feasible. In that way, data-driven algorithms can discover simple structures that can ultimately lead to fast BRL algorithms.

\chapter*{Acknowledgements}\addcontentsline{toc}{chapter}{Acknowledgements}

The authors extend their warmest thanks to Michael Littman, James Finlay, Melanie Lyman-Abramovitch and the anonymous reviewers for their insights and support throughout the preparation of this manuscript.
The authors wish to also thank several colleagues for helpful discussions on this topic:  Doina Precup, Amir-massoud Farahmand, Brahim Chaib-draa, and Pascal Poupart.
The review of the model-based Bayesian RL approaches benefited significantly from comprehensive reading lists posted online by John Asmuth and Chris Mansley.
Funding for Joelle Pineau was provided by the Natural Sciences and Engineering Research Council Canada (NSERC) through their Discovery grants program, by the Fonds de Qu\'{e}b\'{e}cois de Recherche Nature et Technologie (FQRNT) through their Projet de recherche en \'{e}quipe program.
Funding for Shie Mannor and Aviv Tamar were partially provided by
the European Community's Seventh Framework Programme
(FP7/2007-2013) under grant agreement
306638 (SUPREL). Aviv Tamar is also partially funded by the Viterbi Scholarship, Technion.


\appendix
\chapter{Index of Symbols}
\label{notation}

Here we present a list of the symbols used in this paper to provide a handy reference.\footnote{In this paper, we use upper-case and lower-case letters to refer to random variables and the values taken by random variables, respectively. We also use bold-face letters for vectors and matrices.}

\vspace{0.15in}
\begin{table}[ht!]
\centering
\begin{small}
\begin{tabular}{|c|c|} \hline
{\bf Notation} & {\bf Definition}\\ \hline
$\Re$ & set of real numbers \\
$\Nat$ & set of natural numbers \\
$\exptE$ & expected value \\
$\Var$ & variance \\
$\Cov$ & covariance \\
$\KL$ & Kullback-Leibler divergence \\
$\Ent$ & Shannon entropy \\
$\N(m,\sigma^2)$ & Gaussian (normal) distribution with mean $m$ and variance $\sigma^2$ \\
$\P$ & probability distribution \\
$\M$ & model \\
$\A$ & set of actions \\
$K$  & cardinality of $\A$ \\
$T$  & time horizon \\
$a\in\A$ & a nominal action \\
\hline
\end{tabular}
\end{small}
\end{table}

\vspace{0.15in}
\begin{table}[ht!]
\centering
\begin{small}
\begin{tabular}{|c|c|} \hline
{\bf Notation} & {\bf Definition}\\ \hline
$\Y$ & set of outcomes (in MAB) \\
$y \in \Y$ & a nominal outcome \\
$r(y)$ & reward for outcome $y$ \\
$P(y|a)\in \P(\Y)$ & probability of observing outcome $y$ after taking \\
& action $a$ \\
$a^*$ & optimal arm \\
$\Delta_a$ & difference in expected reward between arms $a$ and $a^*$\\
$\Scal$ & set of states (or contexts in a contextual MAB)\\
$s\in\Scal$ & a nominal state (context)\\
$P_{S}(s)\in\P(\Scal)$ & context probability (in contextual MAB)\\
$P(y|a,s)\in \P(\Y)$ & probability of observing outcome $y$ after taking \\
& action $a$ when the context is $s$ (in contextual MAB)\\
$\O$ & set of observations \\
$o\in\O$ & a nominal observation \\
$P\in\P(\Scal)$ & transition probability function \\
$P(s'|s,a)\in\P(\Scal)$ & probability of being in state $s'$ after taking action \\
& $a$ in state $s$ \\
$\Omega(o|s,a)\in\P(\O)$ & probability of seeing observation $o$ after taking \\
& action $a$ to reach state $s$ \\
$q\in\P(\Re)$ & probability distribution over reward \\
$R(s,a)\sim q(\cdot|s,a)$ & random variable of the reward of taking action \\
& $a$ in state $s$ \\
$r(s,a)$ & reward of taking action $a$ in state $s$ \\
$\bar{r}(s,a)$ & expected reward of taking action $a$ in state $s$ \\
$R_{\max}$ & maximum random immediate reward \\
$\bar{R}_{\max}$ & maximum expected immediate reward \\
$P_0\in\P(\Scal)$ & initial state distribution \\
$b_t\in\P(\Scal)$ & POMDP's state distribution at time $t$\\
$\tau(b_{t},a,o)$ & the information state (belief) update equation\\
$\mu$ & a (stationary and Markov) policy \\
$\mu(a|s)$ & probability that policy $\mu$ selects action $a$ in state $s$ \\
$\mu^*$ & an optimal policy \\
$\M^\mu$ & the Markov chain induced by policy $\mu$ \\
$P^\mu$ & probability distribution of the Markov chain induced \\
& by policy $\mu$ \\
$P^\mu(s'|s)$ & probability of being in state $s'$ after taking an action \\
& according to policy $\mu$ in state $s$ \\
$P_0^\mu$ & initial state distribution of the Markov chain induced \\
& by policy $\mu$ \\
$q^\mu$ & reward distribution of the Markov chain induced by \\
& policy $\mu$ \\
\hline
\end{tabular}
\end{small}
\end{table}

\vspace{0.2in}
\begin{table}[ht]
\centering
\begin{small}
\begin{tabular}{|c|c|} \hline
{\bf Notation} & {\bf Definition}\\ \hline
$q^\mu(\cdot|s)$ & reward distribution of the Markov \\
& chain induced by policy $\mu$ at state $s$ \\
$R^\mu(s)\sim q^\mu(\cdot|s)$ & reward random variable of the Markov \\
& chain induced by policy $\mu$ at state $s$ \\
$\pi^\mu\in\P(\Scal)$ & stationary distribution over states of \\
& the Markov chain induced by policy $\mu$ \\
$\Z=\Scal\times\A$ & set of state-action pairs \\
$z=(s,a)\in\Z$ & a nominal state-action pair \\
$\xi=\{z_0,z_1,\ldots,z_T\}$ & a system path or trajectory \\
$\Xi$ & set of all possible system trajectories \\
$\rho(\xi)$ & (discounted) return of path $\xi$ \\
$\bar{\rho}(\xi)$ & expected value of the (discounted) \\
& return of path $\xi$ \\
$\eta(\mu)$ & expected return of policy $\mu$ \\
$D^\mu(s)$ & random variable of the (discounted) \\
& return of state $s$ \\
$D^\mu(z)$ & random variable of the (discounted) \\
& return of state-action pair $z$ \\
$V^\mu$ & value function of policy $\mu$ \\
$Q^\mu$ & action-value function of policy $\mu$ \\
$V^*$ & optimal value function \\
$Q^*$ & optimal action-value function \\
$\gamma$ & discount factor \\
$\alpha$ & linear function over the belief simplex \\
& in POMDPs\\
$\Gamma$ & the set of $\alpha$-functions representing the \\
& POMDP value function\\
$k(\cdot,\cdot)$ & a kernel function \\
$\matK$ & kernel matrix -- covariance matrix -- \\
& $[K]_{i,j}=k(x_i,x_j)$ \\
$\veck(x)=\big(k(x_1,x),\ldots,k(x_T,x)\big)^\top$ & a kernel vector of size $T$ \\
$\matI$ & identity matrix \\
$\D_T$ & a set of training samples of size $T$ \\
$\varphi_i$ & a basis function (e.g., over states or \\
& state-action pairs) \\
$\vecphi(\cdot)=\big(\varphi_1(\cdot),\ldots,\varphi_n(\cdot)\big)^\top$ & a feature vector of size $n$ \\
$\matPhi=[\vecphi(x_1),\ldots,\vecphi(x_T)]$ & a $n\times T$ feature matrix \\
$\vectheta$ & vector of unknown parameters \\
$\Theta$ & a parameter space that the unknown \\
& parameters belong to ($\vectheta\in\Theta$) \\
\hline
\end{tabular}
\end{small}
\end{table}


\chapter{Discussion on GPTD Assumptions on the Noise Process}
\label{gptd-assumptions}

{\bf Assumption~\ref{ass:residual1}} $\;$ {\em The residuals $\Delta \vecV_{T+1}=\big(\Delta V(s_0),\ldots,\Delta V(s_T)\big)^\top$ can be modeled as a Gaussian process.} \\
\vspace{-0.1in}

This may not seem like a correct assumption in general, however, in the absence of any prior information concerning the distribution of the residuals, it is the simplest assumption that can be made, because the Gaussian distribution has the highest entropy among all distributions with the same covariance. \\

\noindent
{\bf Assumption~\ref{ass:residual2}} $\;$ {\em Each of the residuals $\Delta V(s_t)$ is generated independently of all the others, i.e., $\exptE\big[\Delta V(s_i)\Delta V(s_j)\big]=0$, for $i\neq j$.} \\
\vspace{-0.1in}

This assumption is related to the well-known Monte-Carlo (MC) method for value function estimation~\citep{Bertsekas96NP,Sutton98IR}. MC approach to policy evaluation reduces it into a supervised regression problem, in which the target values for the regression are samples of the discounted return. Suppose that the last non-terminal state in the current episode is $s_{T-1}$, then the MC training set is $\Big\{\big(s_t,\sum_{i=t}^{T-1}\gamma^{i-t}r(s_i)\big)\Big\}_{t=0}^{T-1}$. We may whiten the noise in the episodic GPTD model of Eqs.~\ref{eq:GPTD-model4} and~\ref{eq:H-episodic} by performing a whitening transformation with the whitening matrix $\matH^{-1}$. The transformed model is $\matH^{-1}\vecR_T=\vecV_T+\vecN'_T$ with white Gaussian noise $\vecN'_T=\matH^{-1}\vecN_T\sim\N\big(\vec0,\diag(\vecsigma_T)\big)$, where $\vecsigma_T=(\sigma^2_0,\ldots,\sigma^2_{T-1})^\top$. The $t$th equation of this transformed model is
\begin{equation*}
R(s_t)+\gamma R(s_{t+1})+\ldots +\gamma^{T-1-t}R(s_{T-1})=V(s_t)+N'(s_t),
\end{equation*}
where $N'(s_t)\sim\N(0,\sigma^2_t)$. This is exactly the generative model we would use if we wanted to learn the value function by performing GP regression using MC samples of the discounted return as our target (see \S\ref{sss:GP}). Assuming a constant noise variance $\sigma^2$, in the parametric case, the posterior moments are given by Eq.~\ref{eq:param-GPreg2}, and in the non-parametric setting, $\vecalpha$ and $\matC$, defining the posterior moments, are given by Eq.~\ref{eq:nonparam-GPreg2}, with $\vecy_T=(y_0,\ldots,y_{T-1})^\top;\;y_t=\sum_{i=t}^{T-1}\gamma^{i-t}r(s_i)$. Note that here $\vecy_T=\matH^{-1}\vecr_T$, where $\vecr_T$ is a realization of the random vector $\vecR_T$.

This equivalence uncovers the implicit assumption underlying MC value estimation that the samples of the discounted return used for regression are statistically independent. In a standard online RL scenario, this assumption is clearly incorrect as the samples of the discounted return are based on trajectories that partially overlap (e.g., for two consecutive states $s_t$ and $s_{t+1}$, the respective trajectories only differ by a single state $s_t$). This may help explain the frequently observed advantage of TD methods using $\lambda<1$ over the corresponding MC ($\lambda=1$) methods. The major advantage of using the GPTD formulation is that it immediately allows us to derive exact updates for the parameters of the posterior moments of the value function, rather than waiting until the end of the episode. This point becomes more clear, when later in this section, we use the GPTD formulation and derive online algorithms for value function estimation.


\backmatter
\bibliographystyle{plainnat}
\bibliography{BRLS}


\end{document}